%% file: main.tex
\documentclass[twoside,11pt]{article}

\usepackage{graphicx} % Required for inserting images
\usepackage{xcolor}
\usepackage{amsmath}
\usepackage{amssymb}
\usepackage{mathtools}
\usepackage{amsthm}
\usepackage{algorithm}
\usepackage{algorithmic}
\usepackage{subcaption}
\usepackage{caption}
\usepackage{multirow}

\newtheorem{lem}{Lemma}[section]

\newcommand{\calS}{\mathcal{S}}
\newcommand{\calA}{\mathcal{A}}
\newcommand{\calT}{\mathcal{T}}
\newcommand{\calP}{\mathcal{P}}
\newcommand{\cvar}{\operatorname{CVaR}}
\newtheorem{assumption}[subsection]{Assumption}

\usepackage{jmlr2e}

\usepackage{lastpage}
\ShortHeadings{Online Bayesian Risk-Averse RL}{Wang and Zhou}
\firstpageno{1}

\begin{document}

\title{Online Bayesian Risk-Averse Reinforcement Learning}

\author{\name Yuhao Wang \email yuhaowang@gatech.edu \\
       \addr School of Industrial and Systems Engineering\\	
       Georgia Institute of Technology\\ 
       Atlanta, GA 30332, USA
       \AND
       \name Enlu Zhou \email enlu.zhou@isye.gatech.edu \\
       \addr School of Industrial and Systems Engineering\\	
       Georgia Institute of Technology\\ 
       Atlanta, GA 30332, USA}
\maketitle

\begin{abstract}%   <- trailing '%' for backward compatibility of .sty file
In this paper, we study the Bayesian risk-averse formulation in reinforcement learning (RL). To address the epistemic uncertainty due to a lack of data, we adopt the Bayesian Risk Markov Decision Process (BRMDP) to account for the parameter uncertainty of the unknown underlying model. We derive the asymptotic normality that characterizes the difference between the Bayesian risk value function and the original value function under the true unknown distribution.
The results indicate that the Bayesian risk-averse approach tends to pessimistically underestimate the original value function. This discrepancy increases with stronger risk aversion and decreases as more data become available.
We then utilize this adaptive property in the setting of online RL as well as online contextual multi-arm bandits (CMAB), a special case of online RL. We provide two procedures using posterior sampling for both the general RL problem and the CMAB problem. We establish a sub-linear regret bound, with the regret defined as the conventional regret for both the RL and CMAB settings. Additionally, we establish a sub-linear regret bound for the CMAB setting with the regret defined as the Bayesian risk regret. Finally, we conduct numerical experiments to demonstrate the effectiveness of the proposed algorithm in addressing epistemic uncertainty and verifying the theoretical properties.

\end{abstract}


\section{Introduction}
% \begin{itemize}
%     \item Online and offline RL
%     \item Robust reinforcement learning due to Model misspecification caused by lack of data (parameter uncertainty)
%     \item (discussion on inherent/parameter uncertainty (safe learning), constrained MDP)
%     \item Bayesian reinforcement learning
% \end{itemize}

Reinforcement Learning (RL) has emerged as a powerful tool across various domains, including autonomous driving and robotics, in recent years. Within the RL framework, an agent learns by continually interacting with the environment, selecting actions, observing rewards, and transitioning to the next state. 
However, the high cost of real physical interactions motivates learning in simulated environments, which are often constructed using previously collected datasets without further interaction with the environment. This presents a safer and more efficient learning paradigm for scenarios where exploration is costly or dangerous \citep{levine2020offline}. 
The first approach is often referred to as online RL, while the second approach is referred to as offline RL.

Although learning via simulations avoids expensive interactions with the real environment, deploying a policy learned from the simulated training environment into the real environment can be risky due to model mismatch between the two environments, primarily caused by a lack of data. To account for such model mismatch, one stream of work has taken a robust RL approach (e.g., recent works \cite{liu2022distributionally,wang2023finite,blanchet2024double,zhou2024natural}). In the robust RL approach, the unknown underlying distribution is assumed to belong to some ambiguity set, and the goal is to find an optimal robust policy that maximizes the worst-case performance among all distributions in the ambiguity set (i.e., under the adversarial environment).

Despite the fact that the robust approach works well to account for model mismatch, the single worst-case performance metric is inflexible, especially when the adversarial environment is unlikely to occur. Given this, \cite{zhou2015simulation} and \cite{wu2018bayesian} proposed a new way of hedging against model uncertainty called Bayesian risk optimization for the single-stage stochastic optimization problem, where they adopted a Bayesian posterior distribution to estimate the unknown model parameters and imposed a general risk measure on the objective taken with respect to the posterior distribution. For instance, if the risk measure is chosen as the Conditional Value at Risk (CVaR) with risk level $\alpha$, then the desired optimal policy maximizes the average performance over the most adversarial scenarios that occur with posterior probability $1-\alpha$. The choice of risk measures reflects the risk attitude of the decision-maker and is flexible. Moreover, the Bayesian risk formulation exhibits an adaptive property, where the level of risk aversion dynamically adjusts to model uncertainty as the Bayesian posterior is updated. The Bayesian risk formulation was also adapted to the setting of the Markov Decision Process (MDP) (\cite{lin2022bayesian,lin2023approximate}) and offline RL (\cite{wang2023bayesian}). 
% {\color{red} Yuhao: Should we mention that this is a conference paper? There is no overlap between the conference paper and this work. Also, NeurIPS and IJDS belong to different institutions.}
% and possibly Xiaoshuang's paper.
In this work, we also adopt the Bayesian risk formulation.

Notably, most works concerning robust RL have considered the offline setting, where the historical dataset remains fixed throughout the entire learning process. This static data pool fails to bridge the discrepancy between simulated and real-world environments due to a lack of data. When deploying a policy in a practical setting, additional data gathered can significantly reduce model uncertainty, thus paving the way for online RL. There are only a few studies on online robust RL (\cite{wang2021online,badrinath2021robust,dong2022online} with a robust formulation and \cite{wang2023bayesian} with a Bayesian risk formulation). All referenced studies (\cite{wang2021online,badrinath2021robust,wang2023bayesian}) employ off-policy learning, where data are collected using policies external to the agent's control. These approaches also require an exploratory behavior policy, such as the $\varepsilon$-greedy policy, to ensure policy convergence. Nevertheless, reliance on such exploration policies can lead to linear regret. The sole exception that considers on-policy learning is \cite{dong2022online}, which integrates an upper confidence bound (UCB) bonus term into the value function to promote exploration, albeit at the cost of compromising the value function's robust structure. In this work, we study online and on-policy Bayesian risk-averse RL. We choose the behavior policy as the optimal policy solved from a Bayesian risk MDP (BRMDP) with sample approximation. The sample size increases as the risk level increases, effectively reflecting the risk attitude and balancing the trade-off between exploration and exploitation.

On a related note, a particularly intriguing instance of the RL problem is the contextual multi-arm bandit (CMAB), which can be seen as a special case of the general Markov Decision Process (MDP)-based RL problem. This simplification arises either from a single decision horizon or state-independent transitions (refer to Section \ref{sec: CMAB}). CMAB is relevant in various application domains where contexts critically influence outcomes, including personalized treatment, online advertising, and recommendation systems. In this paper, we also study the Bayesian risk formulation of CMAB, treating it separately from the BRMDP setting due to its unique structural characteristics.

% On a related note, a special interest of the RL problem is the contextual multi-arm bandit (CMAB), which can be regarded as a special case of general MDP-based RL problem with the number of horizons to be one or with state-independent transition probability (see Section \ref{sec: CMAB}). CMAB has 
% various application domains where contexts can significantly affect the outcomes, such as personalized treatment, online advertising, recommendation system, etc. In this paper, we will also study the Bayesian risk formulation of CMAB and treat it separately from the BRMDP setting due to its special structure.

We summarize the main contributions of this paper as follows:
\begin{enumerate}
    \item 
    We derive the asymptotic normality of the Bayesian risk value function (the value function of BRMDP) when employing the CVaR risk measure and direct parameterization for the unknown transition kernel. This analysis shows that the Bayesian risk value function asymptotically follows a normal distribution with a mean lower than the original value function. This mean decreases with stronger risk aversion but increases as more data are collected, at a rate of \(O\left(\frac{1}{\sqrt{N}}\right)\), where \(N\) is the total number of data points used to estimate the posterior distribution. This result highlights how BRMDP naturally adjusts risk aversion based on model uncertainty. Compared to the concentration result in \cite{wang2023bayesian}, our findings are more precise and interpretable.

    \item We propose two procedures utilizing posterior sampling for online Bayesian risk-averse reinforcement learning (BRRL), employing CVaR as the risk measure. These methodologies are applied in two different settings: BRMDP and Bayesian risk contextual multi-arm bandit (BRCMAB) with linear payoff. Specifically, for the BRCMAB setting, we show that our proposed algorithm not only achieves sub-linear Bayesian risk regret (BR-Regret), which is consistent with the Bayesian risk formulation, but also attains sub-linear conventional regret as the Bayesian risk value function converges to the original value function over time. Additionally, our numerical studies highlight the effectiveness of both algorithms, particularly demonstrating how the risk level of CVaR effectively balances the exploration-exploitation trade-off, thereby enhancing decision-making in uncertain environments.

\end{enumerate}
% The performance of the optimal policy solved from offline RL depends heavily on the offline dataset, which further depends on the policy that one deployed in the real environment to collect the data. Consequently,

\subsection*{Related Work}
One stream of work that is closely related to the Bayesian risk formulation is the distributionally robust formulation, beginning with a large body of work on the robust Markov Decision Process (RMDP) that laid the foundation for robust RL, including but not limited to \cite{gonzalez2002minimax,el2005robust,iyengar2005robust,xu2010distributionally,wiesemann2013robust,mannor2016robust}. For robust RL, several works first tackled the setting of CMAB, e.g., \cite{si2023distributionally,levine2020offline,mo2021learning,kallus2022doubly,shen2023wasserstein}, and later extended their study to the general MDP setting (e.g., \cite{wang2021online,badrinath2021robust,zhou2021finite,dong2022online,liu2022distributionally,panaganti2022sample,panaganti2022robust,wang2022policy,wang2023finite,blanchet2024double,zhou2024natural}).

The Bayesian risk formulation was first proposed in \cite{zhou2015simulation}, namely Bayesian Risk Optimization (BRO) for stochastic optimization, to provide a flexible way to address parameter uncertainty, whose statistical properties were later established in \cite{wu2018bayesian}. More recent works \cite{lin2022bayesian,lin2023approximate} proposed the formulation of BRMDP with a finite horizon \cite{lin2022bayesian} and an infinite horizon \cite{wang2023bayesian,lin2023approximate}. Among these previous works, \cite{wang2023bayesian} is the most relevant to this paper. The authors demonstrated the validity of the BRMDP formulation with an infinite horizon and proposed a Q-learning algorithm to solve the BRMDP with varying posterior distributions. This work differs from \cite{wang2023bayesian} in four ways: (a) We derive the asymptotic normality for BRMDP with an infinite horizon, which characterizes the convergence of BRMDP with accurate and interpretable expressions, whereas \cite{wang2023bayesian} primarily focuses on the solvability of BRMDP. (b) We consider the online RL problem, where agents choose actions to interact with the real environment, and the actions incur regret, which we aim to minimize. In contrast, in \cite{wang2023bayesian}, the agent learns in a simulated environment with additional access to real-world data from a given behavior policy, where regret is not a concern. (c) Besides the (tabular) MDP setting, we also study the BRCMAB with a linear payoff. (d) We propose two algorithms for the MDP and BRCMAB settings with provable regret bounds defined in two ways: Bayesian risk regret and conventional regret.

\section{Bayesian Risk-Averse Formulation}
Consider a discounted RL environment $\mathcal{M} = (\calS,\calA,\calP,r,\gamma) $, where $\calS$ is the finite state space, $\calA$ is the finite action space, $\calP = \left( \calP_{s,a} \right)_{(s,a)}$ is the transition model, $r(s,a) = \mathbb{E}[R(s,a)] $, $R(s,a)$ and $r(s,a)$ are the random and expected reward function when taking action $a$ in state $s$, respectively, and $\gamma\in (0,1)$ is the discount factor. 

The classic discounted RL aims to learn a (deterministic) policy $\pi^*$ that maximizes the value function

\begin{equation} \label{eq: value function classic}
    V^{\pi} (s) = \mathbb{E}\left[ \sum_{t=0}^\infty \gamma^t R(s_t,\pi(s_t) )| s_0 = s\right], ~~ \forall s\in \calS,
\end{equation}
where $\pi$ is some deterministic policy and state transitions according to $s_{t+1} \sim \calP_{s_t,\pi(s_t)}$.  It is known that $V^\pi$ is the (unique) solution to the following Bellman equation 
\begin{equation} \label{eq: Bellman classic}
    V^{\pi} (s) = r(s,\pi(s))+ \gamma \mathbb{E}_{s'}[V^{\pi}(s')|s,\pi(s)], ~~ \forall s\in \calS.
\end{equation}
Let $\pi^* = \arg\max_\pi V^\pi$ be the optimal policy and  $V^* = V^{\pi^*}$ be the optimal value function. 

In practice, the model parameters such as the transition or reward are often unknown. In this paper, we adopt the Bayesian approach to estimate these unknown parameters.  
For the computational convenience of Bayesian updating, we make the following assumption.
\begin{assumption} \label{assump: parametric assumption}
The underlying MDP model belongs to some parametric family. That is,
$$\mathcal{M} \in \left\{ \mathcal{M}^\theta = (\calS,\calA,\calP^\theta,r^\theta,\gamma): \theta \in \Theta\right\},$$ 
where $\theta$ is the parameter that parametrizes the MDP model $\mathcal{M}^\theta$ and $\Theta$ is some finite-dimensional parametric space. $\calP^\theta = \left(\calP^\theta_{s,a}\right)_{(s,a)\in\calS\times\calA}$ is the parametrized transition kernel and $r^\theta = \left(r^\theta(s,a)\right)_{(s,a)\in\calS\times\calA}$ is the parametrized reward. 
\end{assumption}
Let \(\theta^c\) denote the ``correct" parameter representing the true underlying model \(\mathcal{M} = \mathcal{M}^{\theta^c}\). Assumption \ref{assump: parametric assumption} allows us to estimate \(\mathcal{M}\) by estimating \(\theta^c\). Notably, the parametrization of \(P^\theta\) (or \(r^\theta\)) does not require an explicit functional form in terms of \(\theta\); rather, it ensures that given \(\theta\), we can generate state-reward sample trajectories for the underlying MDP \(\mathcal{M}^\theta\). For transition model parameterization, a direct parameterization is to vectorize all the elements in (upper triangle of) the transition matrix $\calP(\cdot|s,a)$. Beyond direct parametrization, in many problems, the state evolves according to known system dynamics with unknown randomness, where the latter can be modeled using parametric distributions. For example, in inventory control applications, the demand distribution is often estimated from data and typically represented using a parametric distribution.

Taking the Bayesian perspective, the unknown parameter $\theta^c$ is viewed as a random vector $\theta$. Let $\psi$ denote a posterior distribution on $\theta$ computed from data and some prior. With this posterior distribution accounting for model uncertainty, we further impose a risk measure on the future reward, taken with respect to the posterior distribution. Given a coherent risk measure $\rho$ (see the definition of coherent risk measures in Appendix, Section \ref{sec: coherent risk measure}), the Bayesian risk-averse counterpart of the Bellman equation \eqref{eq: Bellman classic} is defined as 
\begin{equation} \label{eq: BR Bellman policy}
    V^{\psi,\pi} (s) =   \rho^\psi_\theta \left( r^{\theta}(s,\pi(s)) + \gamma \mathbb{E}_{s'}^\theta [V^{\psi,\pi}(s')|s,\pi(s)] \right)
\end{equation}
Compared with the Bellman equation \eqref{eq: Bellman classic} for the true problem, here we have an additional layer $\rho_\theta^\psi$, taken with respect to the posterior distribution $\psi$ of the model parameter $\theta$. The Bayesian risk value function under policy $\pi$ and posterior $\psi$ is defined as the unique solution to \eqref{eq: BR Bellman policy} (see \cite{lin2022bayesian,wang2023bayesian}).

The value function satisfying the Bellman equation \eqref{eq: BR Bellman policy} can also be expressed in the following nested form:
\begin{equation} \label{eq:value function}
    \begin{aligned}
    V^{\psi,\pi}(s_0) = \rho^\psi_{\theta_1} \left( r^{\theta_1}(s_0,a_0) +  \gamma\mathbb{E}_{s_1}^{\theta_1}\middle [
    \rho^\psi_{\theta_2} \middle(r^{\theta_2}(s_1,a_1)+ \gamma \mathbb{E}_{s_2}^{\theta_2}  \middle[ 
    \rho^\psi_{\theta_3} \middle( r^{\theta_3}(s_2,a_2)+  \cdots\right.
\end{aligned},
\end{equation}
where $a_t = \pi(s_t), t\ge 1$, $r^\theta$ represents the expected reward parametrized by $\theta$, $\mathbb{E}^{\theta_t}_{s_t}$ denotes the expectation  taken with respect to $s_t$, which follows the distribution $P^{\theta_t}_{s_{t-1},a_{t-1}}$, $\rho^\psi_\theta$ means the risk measure $\rho$ is taken with respect to $\theta$, which follows the posterior distribution $\psi$, and $\theta_1,\theta_2,\ldots$ are i.i.d. random variables following the posterior distribution $\psi$. We note that this nested form may seem to be cumbersome, but it is just the Bayesian risk-averse counterpart of the value function \eqref{eq: value function classic} of the true problem.

We refer to $V^{\psi,\pi}$ as the Bayesian risk-averse value function, and the corresponding model as the Bayesian risk MDP (BRMDP). As studied in \cite{wang2023bayesian} and \cite{lin2023approximate}, the optimal value function $V^{\psi,*}$ of BRMDP can be achieved by some deterministic policy and satisfies the following optimal Bayesian risk Bellman equation:

\begin{equation} \label{eq: BR Bellman}
    V^{\psi,*} (s) = \max_a \left\{ \rho^\psi_\theta \left( r^{\theta}(s,a) + \gamma \mathbb{E}_{s'}^\theta [V^{\psi,*}(s')|s,a] \right)\right\}
\end{equation}
with the Bellman operator defined as 
$$ \mathcal{T}^{\psi} V(s) =  \max_a \left\{ \rho^\psi_\theta \left( r^{\theta}(s,a) + \gamma \mathbb{E}_{s'}^\theta [V(s')|s,a] \right)\right\},$$
which is a contraction mapping with respect to $\|\cdot\|_\infty$, i.e.,
$\|\mathcal{T}^{\psi} V - \mathcal{T}^{\psi} V'\|_\infty \le \gamma \|V-V'\|_\infty$.

For the choice of the posterior distribution, with direct parametrization of the transition matrix, the posterior can be easily computed using a Dirichlet conjugate prior  (see Appendix \ref{sec: Dirichlet} for details).  For the reward model, conjugate priors such as the normal-normal conjugate or the Bernoulli-Beta conjugate are often used to efficiently update the posterior distribution.
% We write $\psi = (\psi^r,\psi^p)$ to denote the (conjugate) posterior distribution for the reward and transition model separately. And $r^\theta \sim \psi^r$, $p^\theta \sim \psi^p$. 
For the choice of risk measure, in this paper, we restrict ourselves to the Conditional Value at Risk (CVaR), which measures the average left-tail risk. Specifically, for a variable $X$, the CVaR with risk level $\alpha$ is defined as  
$$\operatorname{CVaR}_\alpha = \mathbb{E}[X \mid X \leq \operatorname{VaR}_{1-\alpha}(X)],$$  
where $\operatorname{VaR}_\beta(X)$ is the $\beta$-quantile of $X$. 
Note that this definition differs from some conventions that define CVaR as the average right-tail risk. This distinction arises because they consider risk as incurring a large cost (or loss), whereas we treat risk as obtaining a small reward (value function).

\section{Weak Convergence of BRMDP} \label{sec: weak convergence}
% {\color{red} (Can derive for more general parameterization) \\}
In the previous section, we introduced the Bayesian risk-averse formulation BRMDP to address the parametric uncertainty of the true underlying model. In this section, we further present the statistical interpretation of BRMDP, which relates the Bayesian risk-averse value function (the value function of BRMDP) to that of the original true MDP. To streamline the presentation, for all $(s,a) \in \calS\times\calA$, we use Dirichlet conjugate priors (parametrized by $\psi_{s,a} \in \mathbb{N}^{|\calS|}$) to directly estimate $\calP_{s,a}$. We choose this specific approach of direct parametrization because of its expression brevity as well as because it can be applied straightforwardly to any discounted MDP with a finite state and action space. Additionally, we assume that the reward is deterministic, a prevalent assumption in RL studies.

% and known given the current state, action, and the next transition state, that is, $r(s,a) = \mathbb{E}[R(s,a,s')]$, where $R(s,a,s')$ is deterministic and known.  
% Moreover, managing uncertainty in the reward model is typically less complex than addressing uncertainty in the transition model. We concentrate on transition uncertainty and enhance the overall formulation's conciseness.

For a given deterministic policy $\pi$, we have the following assumptions on the observed transition data.
\begin{assumption} \ \label{assump: normality}
    \begin{enumerate}
        \item For each state-action pair $(s,a)$, we have $N_{s,a}$ independent and identically distributed (i.i.d.) observed transition data $\xi_{s,a}^i = (s,a,s'_i)_{i=1}^{N_{s,a}}$, where $s_i' \sim \calP_{s,a}$, and $\{\xi_{s,a}^i\}_{(s,a,i)}$  are independent. 
        \item Let $N = \sum_{s,a} N_{s,a}$, $\forall  s\in\calS$, there exists $\Bar{n}_s > 0$, such that almost surely,
$$ \lim_{N \rightarrow \infty} \frac{N_{s,\pi(s)}}{N} = \bar{n}_{s}.$$
\item \textbf{(Uninformative prior):} The prior distribution for $P_{s,a}$ is a Dirichlet distribution with parameter $\psi^0_{s,a}$ , where $\psi^0_{s,a}(s') = 1, \forall s'\in \mathcal{S}$.
    \end{enumerate}
\end{assumption}
Assumption \ref{assump: normality}.1 indicates that the transition data are i.i.d. for a fixed state-action pair and are independent across different $(s,a)$ pairs. Assumption \ref{assump: normality}.2 states that the number of observations for a state-action pair $(s,\pi(s))$ should increase at a linear rate relative to the total number of observations. This occurs when observations from the real world are obtained using the behavior policy $\pi$ and the MDP under $\pi$ is communicating. Assumption \ref{assump: normality}.3 is not crucial and can be easily extended to any positive prior. We are interested in the relationship between the value function of BRMDP and the original MDP, which is characterized by Theorem \ref{thm: normality BRMDP}.  
Let $\mathcal{P}^\pi = \left(\calP_{s,\pi(s)}\right)_{s\in\calS}$ denote the transition matrix under policy $\pi$, and let $\operatorname{diag}(x)$ denote a diagonal matrix whose diagonal elements are the entries of vector $x$.

\begin{theorem}\textbf{( Weak Convergence} for BRMDP) Suppose Assumption \ref{assump: normality} holds.
For $\alpha \in (0,1]$, 
\label{thm: normality BRMDP}
\begin{align*}
    \sqrt{N}(I - \gamma \calP^\pi) (V^{\psi,\pi} - V^\pi) \Rightarrow \mathcal{N}\left(\gamma \lambda^\pi, \operatorname{diag}((\gamma \sigma^\pi)^2) \right),
\end{align*}
where $\forall s\in \calS$,
\begin{align}
     &\lambda^\pi(s) = \frac{-\sigma^\pi_s }{1-\alpha} \phi(\Phi^{-1}(\alpha)), \notag \\
    &(\sigma_s^\pi)^2 = \frac{1}{\Bar{n}_s}(V^\pi)^\top (\operatorname{diag}(\mathcal{P}_{s,\pi(s)}) - \mathcal{P}_{s,\pi(s)} \mathcal{P}_{s,\pi(s)} ^\top) V^\pi. \label{eq: variance Bars free}
\end{align}
Here $\phi, \Phi$ represents the pdf and cdf of the standard normal distribution.

\end{theorem}

With Theorem \ref{thm: normality BRMDP}, the value function of BRMDP can be expressed as  
\begin{equation} \label{eq: weak interpretation}
    V^{\psi, \pi}=V^\pi-\left(I-\gamma \mathcal{P}^\pi\right)^{-1}\left(\frac{\gamma \lambda^{\pi}}{\sqrt{N}}+\frac{\gamma \sigma^\pi}{\sqrt{N}} Z\right)+o_p\left(\frac{1}{\sqrt{N}}\right),
\end{equation}
where $o_p(\cdot)$ denotes the little-o notation in probability, and $Z \sim \mathcal{N}(0,I_{|\calS|})$ is a standard multivariate normal distribution. For fixed $N$, since $Z$ has mean zero, Equation \eqref{eq: weak interpretation} establishes an equivalence between the value function of the BRMDP and the value function of the original MDP, albeit with the additional negative ``bias" term $\gamma \lambda^\pi(s) = -\frac{\gamma\sigma^\pi_s }{1-\alpha} \phi(\Phi^{-1}(\alpha))$. This bias term encapsulates a risk-averse perspective, ensuring that the Bayesian risk value function remains lower than or equal to the original value function due to risk aversion. 
Moreover, since $\lambda^\pi(s) \rightarrow 0$ as $\alpha \rightarrow 0$, the bias vanishes as the risk level reduces to $0$ (i.e., when the CVaR risk measure reduces to the expectation), meaning that we become risk neutral towards model uncertainty. On the other hand, $\lambda^\pi(s) \rightarrow \infty$ as $\alpha \rightarrow 1$, indicating that the bias increases with stronger risk aversion.

Furthermore, the magnitude of the entire difference, i.e., $V^{\psi,\pi} - V^\pi$, vanishes proportionally to $\frac{1}{\sqrt{N}}$, signifying that the underestimation due to risk aversion decreases with a more precise estimation of the model for any fixed risk attitude $\alpha$. This property endows the BRMDP model with the capacity to naturally adjust its risk aversion as more data are gathered. It demonstrates a significant advantage of BRMDP in online risk-averse reinforcement learning, where the agent can iteratively update the BRMDP based on new information and then solve for an optimal risk-averse policy (of the current BRMDP) that adapts to the data-generation process. Motivated by this observation, in Section \ref{sec: online framework}, we introduce the online framework of Bayesian risk-averse reinforcement learning.

\textbf{Proof Outline of Theorem \ref{thm: normality BRMDP}:} Despite the technical proof details, we provide a proof outline here and defer the complete proof to the appendix. 
 Let $\mathcal{T}^{\psi,\pi}$, $\mathcal{T}^\pi$ be the Bayesian risk Bellman Operator and the classic Bellman operator under policy $\pi$ such that 
    $$\calT^{\psi,\pi} V(s) := r(s,\pi(s)) + \cvar_{\alpha}^{\psi_{s,\pi(s)}}(\mathbb{E}[V(s')|s,a]) \  \ \text{ and } \calT^\pi V(s) := r(s,\pi(s)) + \mathcal{P}_{s,\pi(s)} V^\pi, \ \forall s \in \calS $$
    Then we know 
    $V^{\psi,\pi} = \calT^{\psi,\pi} V^{\psi,\pi}$ and $V^\pi = \calT^\pi V^\pi$.
We can express $V^{\psi,\pi}- V^\pi$ as 
\begin{align}
    &V^{\psi,\pi} - V^\pi \notag\\
    =&\calT^{\psi,\pi} V^{\psi,\pi} - \calT^\pi V^\pi\notag\\
    =& \underbrace{\left(\calT^{\psi,\pi} V^{\pi}  - \calT^\pi V^{\pi}  \right)}_{(I)} + \underbrace{\calT^\pi \left( V^{\psi,\pi} - V^\pi\right)}_{(II)} +\underbrace{\left[\left( \calT^{\psi,\pi}V^{\psi,\pi} - \calT^\pi V^{\psi,\pi}\right) - \left(  \calT^{\psi,\pi} V^\pi - \calT^\pi V^\pi\right) \right]}_{(III)}. \label{eq: Vpsi-V}
\end{align}
We can then show the convergence of the three terms separately.
The second term $(II) = \gamma \mathcal{P}_\pi (V^{\psi,\pi} - V^{\pi})$, where $\mathcal{P}_\pi$ is the transition matrix induced by policy $\pi$, by the definition of Bellman operator. For the third term (III), we can show it is a small error term of order $o\left(\frac{1}{\sqrt{N}}\right)$. 
The first term $(I)$ is equal to 
\begin{align} 
    \calT^{\psi,\pi} V^{\pi}(s)  - \calT^\pi V^{\pi} (s) = \gamma \cvar_\alpha^{\psi_{s,\pi(s)}} \left((p^\top - \calP_{s,\pi(s)}) V^\pi \right),
\end{align}
where $p$ follows a Dirichlet distribution with parameter $\psi_{s,\pi(s)}$.
This indicates that $(I)$ is the CVaR of some Bayesian statistics, whose convergence has been well studied in \cite{wu2018bayesian}. However, to apply their results, the Bayesian parameter space is required to be full rank. In our context, since we use the direct parametrization of the transition probability $\mathcal{P}_{s,\pi(s)}$, the parameter space is the $|\calS|-1$-simplex, which has dimension $|\calS|-1$, less than the number of parameters $|\calS|$. In fact, once we are given values of any $|\calS|-1$ parameters $(p(s))_{s \neq \Bar{s}}$, we know the leftover $p(\Bar{s}) = 1- \sum_{s\neq \Bar{s}}p(s)$. Hence, we can discard one redundant parameter $p(\Bar{s})$ and apply Theorem 4.3 in \cite{wu2018bayesian} to obtain the limiting distribution characterized in terms of $\bar{s}$. Nonetheless, the convergence of $(I)$ should not depend on the choice of $\Bar{s}$, and we further prove the independence of $\Bar{s}$ and obtain the final result \eqref{eq: variance Bars free}.
\hfill $\blacksquare$
% Remark: Notably, although we chose a specific way (direct parametrization) to parametrize the transition distribution, this is for expression brevity and the asymptotic normality can be derived for more general ways of parametrization with similar proof. Indeed, from the above proof of Theorem \ref{thm: normality BRMDP}, in \eqref{eq: Vpsi-V} the second term $(II)$ stays the same no matter the parametrization, and the third term $(III)$ can be proved to converge to $0$ in probability in a similar manner, with assumptions on the posterior distribution that requires certain concentration bounds hold. The first term $(I)$, however, requires extra careful treatment. In short

% v 

\section{Online Bayesian risk-averse reinforcement learning (BRRL)} \label{sec: online framework}
In the previous section, we studied BRMDP from the modeling perspective, showing its weak convergence and the statistical intuition behind it. In this section, we study BRMDP from the algorithmic perspective and develop an algorithm for solving Bayesian risk-averse reinforcement learning (BRRL) in an online RL setting, where the agent interacts with the environment given the current state $s$ by taking an action $a$, then receives a reward $R(s,a)$ and transitions to the next state $s'$. An online RL algorithm needs to specify the choice of action $a$ and how to utilize past learning experience, that is, the past trajectories of rewards and transitions.

The online BRRL framework is as follows. Consider that the entire learning process consists of multiple episodes. Within each episode, we formulate a BRMDP with the posterior updated from data collected in previous episodes. Then, we compute a behavior policy by solving this BRMDP, which is used to interact with the environment and collect a new trajectory of data. To solve the BRMDP, we propose a method based on posterior sampling in the following section.

\subsection{Posterior Sampling for BRMDP}
Recall that $\psi$ is the posterior of the BRMDP at the beginning of the current episode, and the Bellman operator of the BRMDP is  
$$
   \mathcal{T}^{\psi}V (s) =  \max_a \left\{ \rho^\psi_\theta \left( r^{\theta}(s,a) + \gamma \mathbb{E}_{s'}^\theta [V(s')|s,a] \right)\right\}.
$$
One choice for the behavior policy is to compute the optimal policy by finding the fixed point of the above Bellman operator. However, this approach presents two main challenges. First, evaluating the Bellman operator is difficult due to the presence of the risk measure $\rho$. Second, merely selecting the optimal policy leads to a purely exploitative policy, which prevents sufficient exploration of unvisited (or less visited) states and actions. To address both issues, we employ the posterior sampling approach to approximate the exact Bellman operator. This simplifies computation by replacing the CVaR risk measure with its Monte Carlo estimator while also introducing exploration through random sampling from the posterior distribution.

To be specific, let $\theta_1,\ldots,\theta_n$ be $n$ i.i.d. samples from the posterior distributions $\psi$ and $Z_i := r^{\theta_i}(s,a) + \gamma \mathbb{E}^{\theta_i}_{s'} [V(s')]$ be the reward-to-go function under environment parameter $\theta_i$. The Monte Carlo estimator for the CVaR risk measure with risk level $\alpha$ is defined as (see \cite{hong2011monte}): 
\begin{equation} \label{eq: Monte Carlo cvar rl}
    \widehat{C}_\alpha (\{
Z_i\}_{i=1}^n) = \max_x\left\{x -  \frac{1}{n(1-\alpha)} \sum_{i=1}^n (x-Z_i)^+\right\}.
\end{equation}
In fact, the Monte Carlo CVaR estimator $\widehat{C}_\alpha (\{
Z_i \}_{i=1}^n)$ is the CVaR risk measure with risk level $\alpha$ applied to the empirical distribution of the $n$ samples $\{
Z_i\}_{i=1}^n$.
% and use the empirical distribution, $\widehat{\psi}$, of these samples as the approximation of the exact posterior distribution. Then, we obtain a policy by solving a BRMDP whose input posterior distribution is the empirical distribution. With the finite sample approximation, 
The approximate Bayesian risk Bellman operator is then
\begin{equation} \label{eq: approximate BR Bellman}
    \widehat{\mathcal{T}} V (s) :=  \max_a \left\{ \widehat{C}_{\alpha}  (\{
Z_i\}_{i=1}^n) \right\}
   = \max_a \left\{ \widehat{C}_{\alpha} \left( \left\{ r^{\theta_i}(s,a) + \gamma \mathbb{E}^{\theta_i}_{s'} [V(s')] \right \}_{i=1}^n\right)\right\}.
\end{equation}

% the superscript $\widehat{\psi}$ in $\operatorname{CVaR}_{\alpha}^{\widehat{\psi}}  $ means the inside random parameter $\theta$ follows the distribution $\widehat{\psi}$. 
Furthermore, with the finite sample approximation, $\widehat{\mathcal{T}} V$ can be exactly calculated. This enables us to use, e.g., value iteration to obtain the optimal policy $\pi$ from the approximate Bayesian risk Bellman operator satisfying $V^{\psi,\pi} = \widehat{\mathcal{T}} V^{\psi,\pi}$. We then deploy the solved policy $\pi$ for an episode of length $L$, collect new data, and repeat the procedure. We provide the detailed algorithm in Algorithm \ref{alg: BRPS-RL}.

\begin{algorithm}[h]
   \caption{Bayesian Risk-Averse Posterior Sampling for RL (BRPS-RL)}
   \label{alg: BRPS-RL}
\begin{algorithmic}
   \STATE {\bfseries Input:} State space $\calS$, action space $\calA$, reward function $r$, initial state $s_0$, number of episode $T$, length of each episode $L$, 
   %observation batch size $\{n(t)\}_{t=1}^T$ {\color{red} (may not treat as input parameter)}, 
   sample size $n$, risk level $\alpha$, prior distribution $\psi^0$, sample size $n$.
%   \REPEAT
   % \STATE { \bfseries Initialize}  $n \leftarrow \lceil \frac{1}{1-\alpha}\rceil$.
   \FOR{$t=0$ {\bfseries to} $T$}
   \STATE sample $\theta_1,\theta_2,\ldots,\theta_{n} \sim \psi^t$. 
   % \STATE Let $\widehat{\psi}^t$ be the empirical approximation of $\psi^t$ using $\theta_i, i=1,\ldots,n$.
   \STATE Solve for $\pi_t(\cdot)$ using approximate Bayesian risk Bellman operator defined in \eqref{eq: approximate BR Bellman}. 
   \STATE Implement $\pi_t$ and collect trajectory with length $L$.
   \STATE Update posterior distribution $\psi^{t+1}$ using the new observations.
   % \STATE Let $\pi_t(\cdot) \leftarrow \arg\max_{a\in\calA} Q_t(\cdot,a)$, take action $a_t \leftarrow \pi_t(s_t)$ and receive next state $s_{t+1} \sim \calP(\cdot|s_t,a_t)$.
   % \STATE Set $\psi^{t+1}_{s,a}(s') \leftarrow \psi^{t}_{s,a}(s'), \forall s,a,s'$, $\psi^{t+1}_{s_t,a_t}(s_{t+1}) \leftarrow \psi^{t}_{s_t,a_t}(s_{t+1}) + 1$.
   
   % \FOR{{\bfseries all} $(s,a) \in \calS\times\calA$}
   % \STATE Generate $p_1,p_2,\ldots,p_n \sim \psi^t_{s,a}$.
   % \STATE Set $X_i \leftarrow \sum_{s'\in\calS} p_i(s') V_t(s'), i \in [n]$.
   % \STATE Let UCB$_t(s,a) \leftarrow \frac{\gamma}{1-\gamma}\left[ \frac{|\calS|}{\max\{N^t_{s,a},1\}} + 3\sqrt{\frac{1}{\max\{N^t_{s,a},1\}}\log \frac{2n|\calS|^2|\calA|T}{\delta}} \right]$
   % \STATE $Q_{t+1}(s,a) \leftarrow \min\left\{Q_t(s,a), r(s,a) + \gamma \widehat{\rho}_{\psi^t} Q_t (s,a) + \text{UCB}_t(s,a)  \right\}$
   % \STATE $V_{t+1} \leftarrow \max_{a\in\calA} Q_{t+1}(s,a)$
   % \ENDFOR
   
   \ENDFOR
%   \UNTIL{$noChange$ is $true$}
\end{algorithmic}
\end{algorithm}

Notably, the Monte Carlo estimator for CVaR is consistent but biased (see \cite{hong2011monte}), which differs from common risk-neutral RL, where a single sample serves as an unbiased estimator. While increasing the sample size $n$ can reduce the bias, it is more computationally expensive and, more importantly, it can lead to over-exploitation, a common problem in RL. The choice of $n$ will be discussed in Section \ref{sec: regret CMAB} when we analyze the regret.

\section{Bayesian Risk Contextual Multi-Arm Bandit (BRCMAB)} \label{sec: CMAB} 
%In the large literature of reinforcement learning, there is also a stream of work with a special interest in the problem of so-called multi-arm bandit, or more generally, contextual multi-arm bandits (CMAB). A CMAB problem is defined as follows.Suppose we have $K$ arms and a context Space $\mathcal{X}$. Given a context $x$, pulling arm $a \in [K]$ gives us a random reward $R_{x,a}$, with unknown mean $\mu_{x,a}$. Each time we want to pull the arm with the largest $\mu_{s,a}$ given the context variable $x$. A risk-averse decision-maker attempts to exploit more rather than explore. With a Bayesian risk approach, the problem can be derived as a special case of the BRMDP problem, with parameter uncertainty only in the reward model and that $p^\theta_{s,a} = p$, that is, the next state does not depend on the current state and the action taken. We have the value function
So far, we have focused on the discounted RL setting, where the underlying true model is assumed to be a discounted MDP. In the large body of reinforcement learning literature, there is also a stream of work with a special interest in the problem of multi-arm bandits, or more generally, contextual multi-arm bandits (CMAB). CMAB can be regarded as a special case of RL with only a single horizon. In the context of CMAB, the state $s$ is called a context, and taking action $a$ is referred to as pulling an arm $a$. Pulling arm $a$ given context $s$ returns a random reward $R(s,a)$ with expectation $r(s,a)$. Ideally, the decision-maker wants to pull the arm with the largest expected reward given the context each time. That is,  
$$ \pi^*(s) = \arg \max_{a} r(s,a). $$  
On a related note, CMAB can also be derived as a special case of discounted RL, where the current action (pulled arm) has no impact on the future. Specifically, in \eqref{eq: Bellman classic}, we regard the state as the context variable and assume that $\calP_{s,a} = \calP_s$ is independent of the action. Given the optimal value function $V^{*}$, we know the optimal policy $\pi^*$ can be solved by  
$$ \pi^*(s) = \arg\max_a \{ r(s,a) + \gamma \mathbb{E}_{s'} [V^*(s')|s,a]\}. $$ 
Since the second term on the right-hand side does not depend on $a$, we have $\pi^*(s) = \arg\max_a r(s,a)$, which gives us the CMAB objective.

Just as the BRMDP offers a risk-averse perspective on the traditional risk-neutral MDP, we can similarly construct the Bayesian Risk CMAB (BRCMAB) problem. This formulation effectively introduces risk aversion into the conventional CMAB framework by considering a specific instance of the BRMDP, where parameter uncertainty is exclusively confined to the reward model. 
In the BRCMAB setting, the transition probabilities $\calP^\theta_{s,a}$ are assumed to be constant, denoted as $\calP_s$, which signifies that the subsequent state is independent of the action taken. This simplification aligns with the nature of bandit problems, where actions do not affect state transitions. Then, from the Bayesian risk Bellman equation \eqref{eq: BR Bellman}, given the optimal value function $V^{\psi,*}$, the optimal policy $\pi^*$ is solved by  
\begin{equation*} 
    \begin{aligned}
    \pi^*(s) =\arg\max_a \rho^\psi_\theta \left(r^\theta(s,a) + \gamma \mathbb{E}_{s'}[V^{\psi,*}(s')|s,a] \right).
    \end{aligned}
\end{equation*}

Since $\rho$ is a coherent risk measure, it satisfies the property $\rho(X+C) = \rho(X) + C$ if $C$ is a constant. Hence, the optimal policy $\pi^*(s)$ can be solved by  
$$\pi^*(s) =\arg\max_a \rho^\psi (r^\theta(s,a)).$$  
We refer to this formulation as the Bayesian Risk Contextual Multi-Arm Bandit (BRCMAB). In this paper, we study both BRCMAB and the more general BRMDP.

\subsection{BRCMAB with linear payoff}
We now study the BRCMAB problem, which serves as a special case as outlined in Section \ref{sec: CMAB}. Notably, while we assume a finite state space in the BRMDP framework, we relax this assumption in the CMAB problem, allowing the context variable to be continuous. More precisely, we consider CMAB with a linear payoff, a common assumption in the CMAB literature (e.g., \cite{agrawal2014thompson}).

To be specific, given a context $s$, pulling an arm $a$ results in a random reward $R(s,a) = r(s,a) + \varepsilon$, where $r(s,a) = s^\top \theta_a^c$, with $s, \theta_a^c \in \mathbb{R}^d$, representing the expected reward of pulling arm $a$ under context $s$, and $\varepsilon$ is Gaussian noise with mean zero and known variance $\nu^2$.

To estimate the parameter $\theta_a^c$, one can use the Normal-Normal conjugate prior. Specifically, suppose the prior is given as $\mathcal{N}(0,\nu^2 I_d)$. Given $n$ contexts $s_1,\ldots,s_n$ and rewards $R(s_1,a),\ldots,R(s_n,a)$, the posterior of $\theta_a^c$, denoted by $\psi_a$, also follows a normal distribution $\mathcal{N}({\theta}_a,\nu^2V_a^{-1})$, where
\begin{equation} \label{eq: gaussian update} 
    V_a = I_d + \sum_{i=1}^n s_i s_i^\top, \ {\theta}_a = V_a^{-1} \left(\sum_{i=1}^n s_i R(s_i,a) \right). 
\end{equation}

\textbf{Remark:} It is worth noting that the update rule in \eqref{eq: gaussian update} holds only for Gaussian random rewards. In more general contextual bandit scenarios, the posterior typically does not have a closed-form expression, and techniques such as Variational Bayes can be employed to approximate the true posterior, ensuring computational feasibility. This approach involves mapping the true posterior to the family of multivariate normal distributions while retaining the same update rules.

\subsubsection*{Bayesian Risk-Averse Posterior Sampling for CMAB}

To tackle the BRCMAB, similar to BRMDP, we introduce an algorithm inspired by Thompson sampling for the classical risk-neutral CMAB. Thompson sampling selects an arm for pulling by generating a single sample from the posterior distribution, treating it as an (unbiased) estimator of the expected reward, and choosing the arm with the highest estimated value. 
In our BRCMAB, we aim to estimate the Bayesian risk objective with the risk measure set to CVaR at risk level $\alpha \in [0,1)$:
$$\operatorname{CVaR}_\alpha^{\psi_a} \left(r^\theta (s,a)\right) = \operatorname{CVaR}_\alpha^{\psi_a} \left(s^\top \theta \right),$$  
where $\theta \sim \psi_a = \mathcal{N}(\theta_a,\nu^2 V_a)$ and $s^\top \theta \sim  \mathcal{N}(s^\top\theta_a,\nu^2 s^\top V_a s)$. 
Similar to the setting of BRMDP, we adopt a Monte Carlo estimator $\widehat{C}_\alpha (\{r_{a,i}\}_{i=1}^n)$ with $n$ i.i.d. samples $r_{a,1},\dots,r_{a,n} \sim \mathcal{N}(s^\top\theta_a,\nu^2 s^\top V_a s)$, defined as  
\begin{equation} \label{eq: cvar monte carlo}
     \widehat{C}_\alpha (\{r_{a,i}\}_{i=1}^n) = \max_x\left\{x -  \frac{1}{n(1-\alpha)} \sum_{i=1}^n (x - r_{a,i})^+\right\}.
\end{equation}

After obtaining the estimator for CVaR, we pull the arm with the highest CVaR estimate. The detailed algorithm is presented in Algorithm \ref{alg: BRCMAB-Gaussian}.

\begin{algorithm}[H]
   \caption{Bayesian Risk-Averse Posterior Sampling for CMAB (BRPS-CMAB)}
   \label{alg: BRCMAB-Gaussian}
\begin{algorithmic}
   \STATE {\bfseries Input:} K arms,  risk level $\alpha$, contexts set $\{s_t\}_{t=1}^T$, sample size $n$. 
%   \REPEAT
   \STATE { \bfseries Initialize}  $\theta_a \leftarrow 0$, $V_a \leftarrow I_d$.
   \FOR{$t=0$ {\bfseries to} $T$}
   \STATE For $a\in[K]$, sample 
   \begin{equation} \label{algeq: sample}
       r_{a,1},r_{a,2},\ldots,
   r_{a,n}
   \sim \mathcal{N}(s_t^\top\theta_a,\nu^2 s_t^\top V_a s_t)
   \end{equation}
   
   \STATE Pull arm $a_t = \arg\max_a \widehat{C}_\alpha(\{r_{a,i}\}_{i=1}^n)$ and receive reward $R(s_t,a_t)$.
   \STATE Update $(\theta_{a_t},V_{a_t})$ according to \eqref{eq: gaussian update}.    
   \ENDFOR
%   \UNTIL{$noChange$ is $true$}
\end{algorithmic}
\end{algorithm}

\section{Regret Analysis} 
In this section, we analyze the regret of BRPS-CMAB (Algorithm 2) and a variant of BRPS-RL (Algorithm 1). 
\label{sec: regret CMAB}
\subsection{Regret Analysis for BRPS-CMAB}
In the previous section, we presented the online Bayesian risk-averse posterior sampling framework. Recall that in the implementation of the proposed algorithms, the sample size $n$ must be determined for constructing the Bayesian risk (CVaR) estimator. In particular, with a higher risk level $\alpha$ (indicating a stronger risk aversion), we need to estimate a narrower $(1-\alpha)$-tail expectation. This requires a larger sample size, as each sample is less likely to fall into this left-tail region, i.e., the (posterior) probability of $r^{\theta_i}(s,a) \le \operatorname{VaR}_{1-\alpha}^{\psi_a}(r^\theta(s,a))$ is smaller. 
Motivated by this, in the following proof, we choose the sample size as a function of the risk level, $n = \lceil\frac{1}{1-\alpha}\rceil$. Clearly, $n$ increases as $\alpha$ increases, accounting for the increased difficulty of estimating CVaR with a larger risk level. Moreover, this choice of \(n\) also balances the exploitation-exploration trade-off. With stronger risk aversion, more samples are generated, leading to a greater emphasis on exploitation. 
When $\alpha = 0$, the risk-averse CVaR objective reduces to the risk-neutral expectation objective, and the algorithm simplifies to Thompson Sampling.

In addition, with this choice of $n$, we define the CVaR estimator as 
\begin{equation} \label{eq: CVaR estimator modefied}
    \Tilde{C}_\alpha(\{r_{a,i}\}_i) = 
\left\{ 
\begin{aligned}
    & r_{a,1:n} \ &\text{if } \frac{1}{1-\alpha}\in\mathbb{Z}\\
    &r_{a,2:n} \ &\text{if } \frac{1}{1-\alpha} \not\in \mathbb{Z}
\end{aligned}
\right.,
\end{equation}
where $r_{a,k:n}$ is the $k^{th}(k\le n)$ (smallest) order statistic given samples $\{r_{a,i}\}_{i=1}^n $. This definition \eqref{eq: CVaR estimator modefied} of CVaR estimator is a modification of the Monte Carlo estimator \eqref{eq: cvar monte carlo}. Indeed, with $n=\lceil\frac{1}{1-\alpha} \rceil$, \eqref{eq: cvar monte carlo} can be expressed as 
$$ \widehat{C}_\alpha (\{
r_{a,i}\}_{i=1}^n) =\frac{1}{n(1-\alpha)}r_{a,1:n} + \left(1-\frac{1}{n(1-\alpha)}\right) r_{a,2:n}.$$
When $\frac{1}{1-\alpha} \in \mathbb{Z}$, $\Tilde{C}_\alpha (\{
r_{a,i}\}_{i=1}^n) = \widehat{C}_\alpha (\{
 r_{a.i}\}_{i=1}^n) $, which is the same as the Monte Carlo estimator. Otherwise 
$\Tilde{C}_\alpha(\{
r_{a,i}\}_{i=1}^n) > \widehat{C}_\alpha (\{
r_{a,i}\}_{i=1}^n) $, where we increases the Monte Carlo estimator by $\frac{1}{n(1-\alpha)}\left( r_{a,2:n}-r_{a,1:n}\right)$. We want to point out that such modification is only for technical proof reason and is not required when implementing Algorithm \ref{alg: BRCMAB-Gaussian}. 

The modified CVaR estimator satisfies the following property as shown in Lemma \ref{lem: order statistic}, which is crucial in the regret analysis and can be of independent interests.
\begin{lem} \label{lem: order statistic}
    Assume the random variable $X \ge 0$ has a cdf $F$ and a pdf $f$. Then the following holds:
    \begin{enumerate}
        \item $\mathbb{E} \left[ X_{1:n}\right] \ge \operatorname{CVaR}_{\frac{1}{n}}(X)$ for $n \ge 1$;
        \item $\mathbb{E} \left[ X_{2:n}\right] \ge \operatorname{CVaR}_{\frac{2}{n}}(X)$ for $n\ge2$.
    \end{enumerate}
    As a result, we have
    $$ \mathbb{E} \left[ \Tilde{C}_\alpha \left(\{X_i\}_{i=1}^n \right)\right] \ge \operatorname{CVaR}_\alpha(X).$$
\end{lem}
\subsection*{Bayesian Risk Regret (BR-Regret)}
In this section, we consider the performance metric of Bayesian risk regret, a risk-averse definition of regret.  
Let $\mathcal{I}_a(t) = \{ \tau \le t: a_\tau = a \}$ be the set of iterations when arm $a$ is pulled up to time $t$. Let $\psi^t_a$ denote the posterior distribution for $\theta_a^c$ at iteration $t$, which follows a normal distribution with mean $\theta_{a,t}$ and covariance matrix $\nu^2 V_{a,t}$, where  
\begin{equation} \label{eq: gaussian update t}
        V_{a,t} = I_d + \sum_{\tau \in \mathcal{I}_a(t)} s_\tau s_\tau^\top, \quad {\theta}_{a,t} = V_{a,t}^{-1} \left(\sum_{\tau \in \mathcal{I}_a(t)} s_\tau R(s_\tau,a) \right). 
\end{equation}  
We introduce the Bayesian risk regret, defined as:  
\begin{equation*}
     \text{BR-Regret}(T) = \mathbb{E} \left[\sum_{t=1}^T  r^t(s_t, a^*_t) - r^t(s_t,a_t) \right],
\end{equation*}
where $r^t(s,a) = \operatorname{CVaR}_\alpha^{\psi_a^t} (s^\top \theta)$ is the Bayesian risk-averse expected reward with the current posterior $\psi^t_a$, and $a_t^* = \arg\max_a r^t(s_t,a)$. The expectation here is taken with respect to all randomness, including the random actions, rewards, and contextual variables (if the contextual variables are random).  
This definition of BR-Regret emanates from our primary objective and can be interpreted as a risk-averse adaptation of the conventional Bayes Regret (e.g., see \cite{russo2018tutorial}), with two key differences. The Bayes Regret is defined as  
\begin{equation*}
\text{BayesRegret}(T) = \mathbb{E} \left[ \sum_{t=1}^T \left(r^\theta(s_t,a^*_t) - r^\theta(s_t,a_t)\right)\right],
\end{equation*}
where $\theta$ is the parameter in the linear payoff, and the expectation is taken with respect to all randomness, including the random action $a_t$, context $s_t$ (if random), and the parameter $\theta$. The optimal action $a^*_t$ in this case depends on the parameter $\theta$, which is integrated over its prior distribution.  
The first key difference in BR-Regret is that the optimal action $a^*_t$ depends on the current posterior distribution $\psi^t$ instead of the reward parameter $\theta$. The second difference is that we impose CVaR on the expected reward (as in the definition of $r^t(s,a)$), rather than on the expectation as in BayesRegret. Both modifications reflect the decision maker's risk attitude in the decision-making process.

To prove the regret bound, we additionally make the following assumption. 
\begin{assumption} \label{assump: CMAB BR-regret} \ 
    \begin{enumerate}
        \item  The reward and the contexts satisfy $R(s,a) \in [0,1]$ and $s\ge 0,\|s\|_2\in[0,1]$, for all $s$ and $a$. 
        \item The posterior distribution for $s_t^\top \theta_a^c$ at time $t$ is a truncated normal distribution with support $[0,1]$, whose mean $s_t^\top \theta_{a,t}$ and variance $\nu^2 s_t^\top V_{a,t} s_t$, where $\theta_{a,t}, V_{a,t}$ are specified in \eqref{eq: gaussian update t}.
    \end{enumerate}
   
\end{assumption}

Assumption \ref{assump: CMAB BR-regret}.1 requires that both the random reward and the context variable are bounded in $[0,1]$. This assumption does not sacrifice generality as long as the reward is bounded (by a linear change of variables), which often holds in practice.  
% In general, if we have $R_{x,a} \in [0,\Bar{r}]$, the BR-Regret bound is scaled by $\Bar{r}$.  
Assumption \ref{assump: CMAB BR-regret}.2 can be justified from the Variational Bayes perspective. Since we assume $R(s,a) \in [0,1]$, the true posterior distribution of $s_t^\top \theta_a^c$ will not follow the normal distribution $\mathcal{N}(s_t^\top \theta_{a,t},\nu^2 s_t^\top V_{a,t} s_t)$, meaning that we no longer have a conjugate prior.  
For a general reward distribution, the posterior distribution may not belong to any parametric distribution family, even if the prior is from a parametric family. In such cases, the posterior cannot be represented by a fixed-dimensional parameter, and computing or sampling from it requires approximation methods (e.g., MCMC), which can be computationally expensive.  
To mitigate this, we approximate the true posterior distribution by a truncated normal distribution (known as the variational distribution) with the same parameters specified in \eqref{eq: gaussian update t}. We denote this truncated normal distribution by $\mathcal{N}^+(s_t^\top \theta_{a,t},\nu^2 s_t^\top V_{a,t} s_t)$. A random variable $X \sim \mathcal{N}^+(\mu,\sigma^2)$ has the same distribution as $\min\{1,\max\{Y,0\}\}$, where $Y \sim \mathcal{N}(\mu,\sigma^2)$.  
We refer to Algorithm \ref{alg: BRCMAB-Gaussian}, with $\mathcal{N}(s_t^\top \theta_{a,t},\nu^2 s_t^\top V_{a,t} s_t)$ replaced by $\mathcal{N}^+(s_t^\top \theta_{a,t},\nu^2 s_t^\top V_{a,t} s_t)$, as Algorithm \ref{alg: BRCMAB-Gaussian}'.

Now we are ready to give the result on BR-Regret.
\begin{theorem} \label{thm: BR-regret MAB} \textbf{(BR-Regret)} Suppose Assumption \ref{assump: CMAB BR-regret} holds and the risk level is set to $\alpha\ge \frac{1}{2}$. Then, for Algorithm $\ref{alg: BRCMAB-Gaussian}'$,
$$ \operatorname{BR-Regret}(T) \le \left(\nu\sqrt{2(1-\alpha)\ln (KT^2) } + \frac{\nu}{1-\alpha}(\frac{1}{\sqrt{2\pi}}+\phi( \Phi^{-1}(\alpha)))\right) 5\sqrt{dKT\ln T} + 1.$$
\end{theorem}

Theorem \ref{thm: BR-regret MAB} indicates that the BR-Regret upper bound of Algorithm \ref{alg: BRCMAB-Gaussian}$'$ increases in the order of $O\left(\sqrt{T} \ln T\right)$ with respect to the total number of iterations $T$. Furthermore, it also increases as the risk level $\alpha$ increases, following an order of $O\left(\frac{1}{1-\alpha}\right)$.

\textbf{Proof Outline: } We give a proof outline here and defer the complete proof into the Appendix. By Lemma \ref{lem: order statistic}, we have $r^t(s_t,a^*_t) \le \mathbb{E}[\widehat{r}^t(s_t,a^*_t)|\psi^t_{a^*_t}]$, where $\widehat{r}^t(s_t,a) = \Tilde{C}_\alpha (\{
r_{a,i}\}_{i=1}^n)$ is the CVaR estimator with samples $r_{a,i}\sim \mathcal{N}^+(s_t^\top \theta_{a,t},\nu^2 s_t^\top V_{a,t} s_t)$.  Hence, we can upper bound BR-Regret(T) as 
$$ \operatorname{BR-Regret}(T) \le \sum_{t=1}^T \mathbb{E}\left[ \widehat{r}^t(s_t, a^*_t) - r^{t}(s_t,a_t)\right] \le \sum_{t=1}^T \mathbb{E}\left[ \widehat{r}^t(s_t, a_t) - r^t(s_t,a_t)\right],$$
where the second inequality holds since $a_t$ maximizes $\Tilde{C}_\alpha (\{
r_{a,i}\}_{i=1}^n)$ over $\calA$. With such an upper bound, we transfer the difference of the Bayesian risk objective value between the optimal arm and the pulled arm ($ r^t(s_t, a^*_t) - r^t(s_t,a_t)$) into the difference between the true Bayesian risk objective value and its estimation of only the pulled arm. We then construct the following concentration event which holds with probability at least $1-\frac{1}{T}$:
$$ \mathbf{D} = \left\{\widehat{r}^t(s_t,a) \le r^t(s_t,a) + C\sqrt{s_t^\top (V_a^t)^{-1} s_t}, \forall a\in[K],t\in[T] \right\},$$
where  $C =\left(\nu\sqrt{2(1-\alpha)\ln (KT^2) } + \frac{\nu}{1-\alpha}(\frac{1}{\sqrt{2\pi}}+\phi( \Phi^{-1}(\alpha)))\right)$ is a constant.
    Then, we can upper bound BR-Regret(T) as
    $$
    \begin{aligned}
       \operatorname{BR-Regret}(T) \le& (1-\frac{1}{T})\sum_{t=1}^T C \sqrt{s_t^\top (V_{a_t}^t)^{-1} s_t} + \frac{1}{T} T   \\
       \le &  5C\sqrt{dKT\ln T} + 1
    \end{aligned},
   $$
   where the second inequality holds according to proof of Theorem 2 in \cite{lin2022risk}.  
\hfill $\blacksquare$

\subsection*{Conventional Regret}
Besides BR-Regret, we are also interested in the following regret measure, defined in the conventional way:
\begin{equation*}
    \operatorname{Regret}(T) = \sum_{t=1}^T \left(r(s_t,a^*_t) - r(s_t,a_t)\right).
\end{equation*}
Here, $a^*_t = \arg\max_a r(s_t,a)$ is the optimal arm for context $s_t$ in the original problem. Our goal is to demonstrate that, despite our algorithm being designed for the Bayesian risk formulation, it ultimately converges toward the original objective of identifying the true optimal arm.
To obtain the regret bound, a slight adjustment is applied to Algorithm \ref{alg: BRCMAB-Gaussian}. Specifically, within the sampling process, instead of sampling from the posterior distribution as in \eqref{algeq: sample}, we replace $\nu$ with $\nu_t = \frac{3}{4} \sqrt{ d \ln \left(\frac{4t}{\delta}\right)}$ at iteration $t$.  
% where $R$ is the sub-Gaussian parameter for the random reward such that $R(s,a)$ is $R$-subGaussian $\forall s,a$. In the context of rewards in the range of $[0,1]$, $R$ can be set to $\frac{1}{4}$.  
This approach of replacing $\nu$ with $\nu_t$ was proposed in \cite{agrawal2014thompson} to maintain a sufficiently large exploration rate, ensuring that the probability of the algorithm getting stuck on a sub-optimal arm due to insufficient data remains bounded away from zero.  
We refer to this modified algorithm as Algorithm \ref{alg: BRCMAB-Gaussian}$''$. The regret bound under the conventional regret measure is given by the following theorem.

\begin{theorem} \textbf{(conventional regret)} \label{thm: bandit conventional regret bound}
     Suppose Assumption \ref{assump: CMAB BR-regret}.1 holds. Then for Algorithm $\ref{alg: BRCMAB-Gaussian}''$, with probability $1-\delta$,
    $$  \operatorname{Regret}(T)  \le \frac{3(4\sqrt{\pi}e)^{\frac{2-\alpha}{1-\alpha}}}{2} \sqrt{d\ln{\frac{4T}{\delta}}\ln{\frac{2T^2}{1-\alpha}} } \left( 15\sqrt{dKT\ln{T}} + 6\sqrt{2\ln{\frac{2}{\delta}T}}\right).$$
    What's more,
    $$\mathbb{E}[ \operatorname{Regret}(T)] \le  \frac{3(4\sqrt{\pi}e)^{\frac{2-\alpha}{1-\alpha}}}{2}\sqrt{d\ln{{4T^2}}\ln{\frac{2T^2}{1-\alpha}} } \left( 15\sqrt{dKT\ln{T}} + 6\sqrt{2T\ln{2T^2}}\right)$$
\end{theorem}
Theorem \ref{thm: bandit conventional regret bound} indicates the 
 conventional regret bound increases in an order $O\left(\sqrt{T \ln^3T}\right)$ in terms of $T$, which is the same as in \cite{agrawal2014thompson}. In addition, it also increases as the risk level $\alpha$ increases in the order of $ O\left(e^{\frac{1}{1-\alpha}}\sqrt{\ln{\frac{1}{1-\alpha}}}\right)$.

The proof of Theorem \ref{thm: bandit conventional regret bound} extends the previous work of \cite{agrawal2014thompson}, which addressed risk-neutral Thompson sampling for CMAB. Notably, the key distinction between our modified algorithm and theirs lies in the sampling procedure for the expected reward. Specifically, their approach generates a single sample of $r(s_t,a)$ from the posterior for each arm $a$ and selects the arm with the highest sampled value, whereas we generate multiple samples to compute an estimator of the Bayesian risk objective (CVaR objective) and pull the arm with the largest estimator. A similar line of proof follows if we can establish the desired properties of our Bayesian risk objective estimators, analogous to those demonstrated for the single-sample case in \cite{agrawal2014thompson}.

Below we discuss some comparisons between the two types of regrets:
\begin{enumerate}
    \item  \textbf{Algorithm Implementation:} When implementing the algorithms in practice, Theorem \ref{thm: bandit conventional regret bound} requires specifying the confidence value $\delta$ to obtain the high-probability bound (or $T$ to derive the expected regret bound) and imposing a substantial sampling variance $\nu_t$ to achieve a sub-linear (traditional) regret bound. However, this can lead to slower convergence in practice. In contrast, Theorem \ref{thm: BR-regret MAB} does not require such additional parameters. Empirical studies in Section \ref{sec: numerical MAB} indicate that Algorithm \ref{alg: BRCMAB-Gaussian}, without modifying the value of $\nu$, is generally more efficient.

    \item \textbf{Regret Bound Behavior:} As $\alpha \rightarrow 1$, the conventional regret grows at an order of $O\left(e^{\frac{1}{1-\alpha}}\sqrt{\ln\frac{1}{1-\alpha}}\right)$, which is significantly faster than the order $O\left(\frac{1}{1-\alpha}\right)$ of BR-Regret. Consequently, for a more risk-averse attitude (larger $\alpha$), a much smaller upper bound on BR-Regret can be attained compared to the conventional regret. This aligns well with the algorithm's design goal, which aims to pull the arm that maximizes the Bayesian risk-expected performance. Moreover, the BR-Regret bound also increases more slowly in $T$—at an order of $O\left(\sqrt{T} \ln T\right)$—compared to the conventional regret bound, which grows at an order of $O\left(\sqrt{T \ln^3 T}\right)$.

\end{enumerate}
\subsection{Regret Analysis for BRPS-RL}
Let $s_{t,1},\ldots,s_{t,L}$ be the states observed when implementing $\pi_t$. Similar to the BRCMAB, we can define the BR-Regret and the conventional regret respectively as  
\begin{equation*}
    \operatorname{BR-Regret}\left(T\right)=\mathbb{E}\left[\sum_{t=1}^T \sum_{\ell=1}^L \left(V^{\psi^t,*}(s_{t,\ell})-V^{\psi^t,\pi_t}(s_{t,\ell})\right) \right],
\end{equation*}
and  
\begin{equation*}
    \operatorname{Regret}(T) =  \sum_{t=1}^T \sum_{\ell=1}^L \left(V^*(s_{t,\ell}) - V^{\pi}(s_{t,\ell})\right).
\end{equation*}
Deriving regret bounds for both types in the RL setting poses significant challenges due to two primary technical difficulties. The first is the nested formulation of BRMDP, which involves the multi-layer application of the CVaR risk measure. This structure prevents the use of Lemma \ref{lem: order statistic} to establish a monotonicity property between the value functions of the BRMDP with the true posterior distribution \(\psi\) and its finite-sample approximation. Such a monotonicity property is a crucial step in proving Theorem \ref{thm: BR-regret MAB} in the bandit setting. 
%Unlike the BRCMAB setting, where the regret bound can be derived by bounding \(r^t(s_t,a^*_t)\) using its Monte Carlo estimator \(\mathbb{E}[\widehat{r}^{t}(s_t,a_t^*)|\psi^t_{a_t^*}]\), the nested structure in BRMDP prevents a straightforward result of the form \(\mathbb{E}[V^{\widehat{\psi},*}] \ge V^{\psi,*}\). Consequently, directly bounding the difference between policies becomes extremely difficult.
The second difficulty arises from the non-linearity of the Bellman operator, which precludes the common technique of iteratively applying the Bellman operator to decompose the value function differences (i.e., regret for a single iteration) as a summation of a vanishing (discounted) sequence of differences between Bellman operators.

The inability to analyze the regret without employing the aforementioned techniques stems from the challenge of characterizing the ``exploration rate" of Algorithm \ref{alg: BRPS-RL}. However, this issue can be addressed by introducing an exploration bonus term to the value function estimate. This approach is inspired by \cite{dong2022online}, which studied online distributionally robust RL.  
The details of the modified algorithm can be found in Appendix \ref{sec: regret RL}. By adding the exploration bonus term, we obtain the following result, with details deferred to Appendix \ref{sec: regret RL}

\begin{theorem} \label{thm: UCBVI}
    With Probability $1-\delta$, the regret of Algorithm \ref{alg: Online BRVI} satisfies
    \begin{align*}
        \text{Regret}(T) \le &\frac{6\gamma}{(1-\gamma)^2}\sqrt{\frac{|\calS|^2}{2}\log{\frac{4(3-2\alpha)|\calS|^2|\calA|T}{(1-\alpha)\delta}}} \min\{ |\calS|\sqrt{T+1},\sqrt{|\calS|(T+1)\ln{(T+1)}}\}\\
     & + \gamma\frac{2|\calS|+2}{(1-\gamma)^2} + \frac{2\gamma}{(1-\gamma)^2}\sqrt{2T\ln{\frac{2}{\delta}}}.
    \end{align*}
In particular, setting $\delta = \frac{1}{T}$, we obtain
\begin{align*}
        \mathbb{E} \left[\text{Regret}(T)\right] \le &\frac{6\gamma}{(1-\gamma)^2}\sqrt{\frac{|\calS|^2}{2}\log{\frac{4(3 - 2\alpha)|\calS|^2|\calA|T^2}{1-\alpha}}} \min\{ |\calS|\sqrt{T+1},\sqrt{|\calS|(T+1)\ln{(T+1)}}\}\\
     & + \gamma\frac{2|\calS|+2}{(1-\gamma)^2} + \frac{2\gamma}{(1-\gamma)^2}\sqrt{2T\ln{2T}} + \frac{1}{1-\gamma}.
    \end{align*}

\end{theorem}

% Nevertheless, this approach is problematic because it conflicts with the risk-averse attitude central to the Bayesian risk framework.

% The second difficulty is also faced in the context of distributionally robust RL. To our best knowledge, only \cite{dong2022online} took a distributionally robust approach and considered an online RL setting where the behavior policy (policy that is used to collect data) is changing and controlled by the decision maker. To derive the regret bound, they add a (sufficiently large) bonus term to the (state-action) value function estimation to force the exploration. However, such approach is against the robust (risk-averse) principle. Both the distributionally robust approach in \cite{dong2022online} and our Bayesian risk approach tend to underestimate the value function by either considering the worst-case scenario or imposing a risk measure. Adding the exploration term violates such risk-averse attitudes.  In fact, for our approach, we can also obtain a sublinear regret bound by adding such a bonus term for exploration, which can be found in the Appendix. 

The effectiveness of Algorithm \ref{alg: BRPS-RL} will be further demonstrated through numerical experiments in Section \ref{sec: numerical}.

\section{Numerical Study} \label{sec: numerical}
\subsection{Gaussian Contextual Bandits with linear Payoff} \label{sec: numerical MAB}
We first compare Algorithm \ref{alg: BRCMAB-Gaussian} (denoted as BRPS-CMAB) with the classical Thompson Sampling algorithm for CMAB (denoted as TS-CMAB) with linear payoff, as proposed by \cite{agrawal2014thompson}. We consider a Gaussian bandit setting with a linear payoff, where the contextual variable $s \sim \operatorname{Uniform}[0,1]^3$, and the set of arms is $\mathcal{K} = \{1,\ldots,10\}$.  
For each arm $i \in \mathcal{K}$ and context $s$, the reward follows $r(s,i) \sim \mathcal{N}(s^\top \theta_i^c,1)$, where $\theta_i^c =(0.5,0.5+\operatorname{sin}i,0.5+\operatorname{cos}i)$. For all implemented algorithms, the initial prior distribution for $\theta_i^c, i \in \mathcal{K}$, is $\mathcal{N}(\mathbf{0},\nu^2 \mathrm{I})$ with $\nu =1$.  
We also denote by BRPS-CMAB$_\nu$ Algorithm \ref{alg: BRCMAB-Gaussian}'' with $\nu = \frac{3}{4}\sqrt{6\ln (2T)}$ to obtain the regret bound in Theorem \ref{thm: bandit conventional regret bound}, and TS-CMAB$_\nu$ as Thompson Sampling with $\nu =  \frac{3}{4}\sqrt{6\ln T}$ to obtain the regret bound as stated in Theorem 1 of \cite{agrawal2014thompson}.

We first plot the result with four aforementioned algorithms using both BR-Regret and the conventional regret criteria. For BRPS-CMAB and BRPS-CMAB$_\nu$, we set the risk level $\alpha = 0.8$. For both regrets, the x-axis represents the iterations and the y-axis represents the average cumulative regret calculated by running 200 macro-replications. The strips show the $95\%$ confidence interval.
\begin{figure}[H]
\centering
\caption{BR-Regret and conventional regret}
\begin{subfigure}[t]{0.45\linewidth}
    \includegraphics[width=\textwidth]{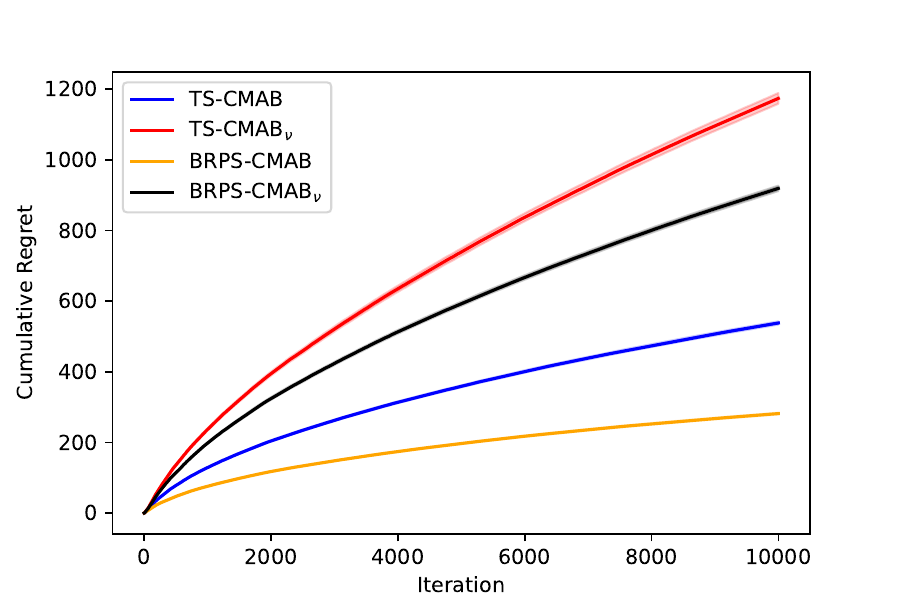}
    \caption{Comparison using BR-Regret}
    \label{fig: bandit BR-regret nu}
\end{subfigure}
 \begin{subfigure}[t]{0.45\textwidth}
\includegraphics[width=\textwidth]{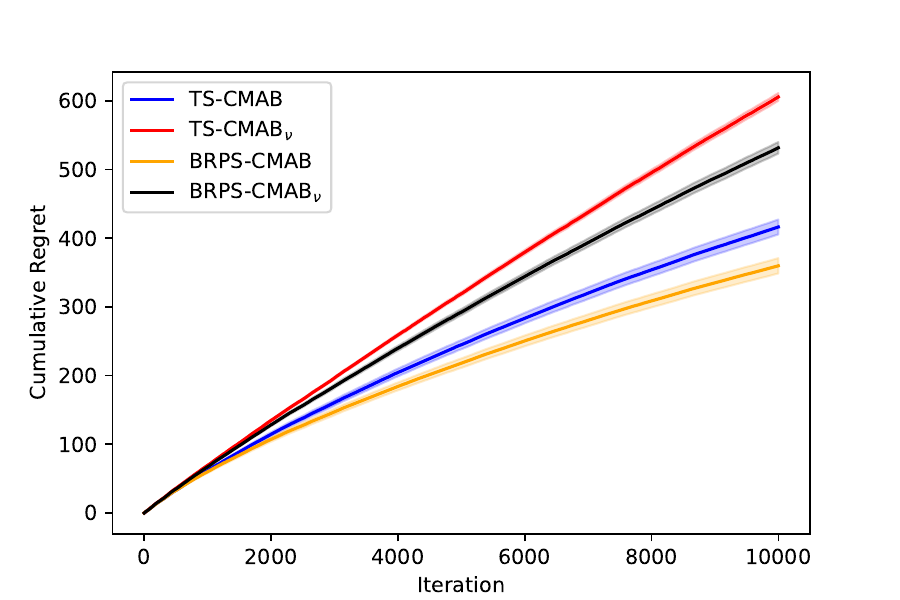}
    \caption{Comparison using conventional regret}
    \label{fig: bandit conventional regret nu}
\end{subfigure}  
\end{figure}
Figures \ref{fig: bandit BR-regret nu} and \ref{fig: bandit conventional regret nu} indicate that both TS-CMAB$_\nu$ and BRPS-CMAB$_\nu$ are significantly outperformed by BRPS-CMAB and TS-CMAB. The underperformance of TS-CMAB$_\nu$ and BRPS-CMAB$_\nu$ can be attributed to an excessive enlargement of the variance in their posterior distributions (due to $\nu = O(\sqrt{\ln T})$), leading to overly aggressive exploration and substantially higher cumulative regrets.  
Consequently, our subsequent analysis will focus exclusively on comparing the performance of TS-CMAB and BRPS-CMAB, which have consistently demonstrated superior performance over TS-CMAB$_\nu$ and BRPS-CMAB$_\nu$.

\begin{figure}[h]
    \centering
    \caption{Comparison of BRCMAB and TS using BR-Regret}
    \label{fig: bandit BR-Regret}
\begin{subfigure}[t]{0.32\textwidth}
\includegraphics[width=\linewidth]{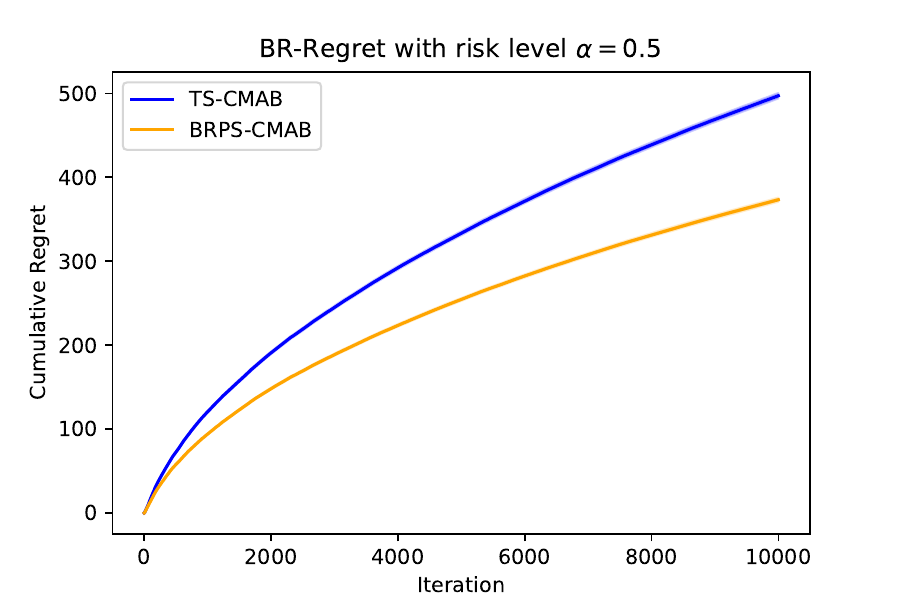}
\end{subfigure}
\begin{subfigure}[t]{0.32\textwidth}
\includegraphics[width=\linewidth]{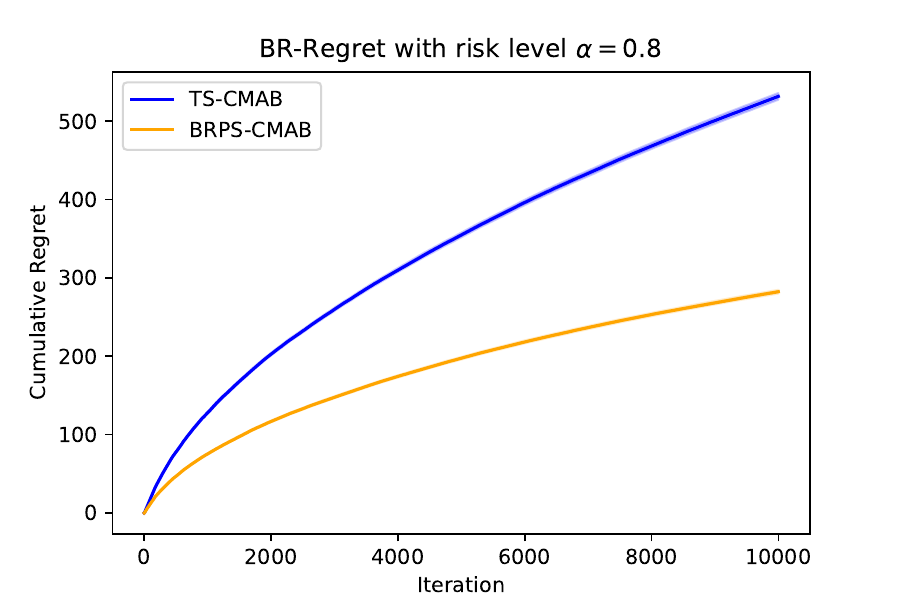}
\end{subfigure}
\begin{subfigure}[t]{0.32\textwidth}
\includegraphics[width=\linewidth]{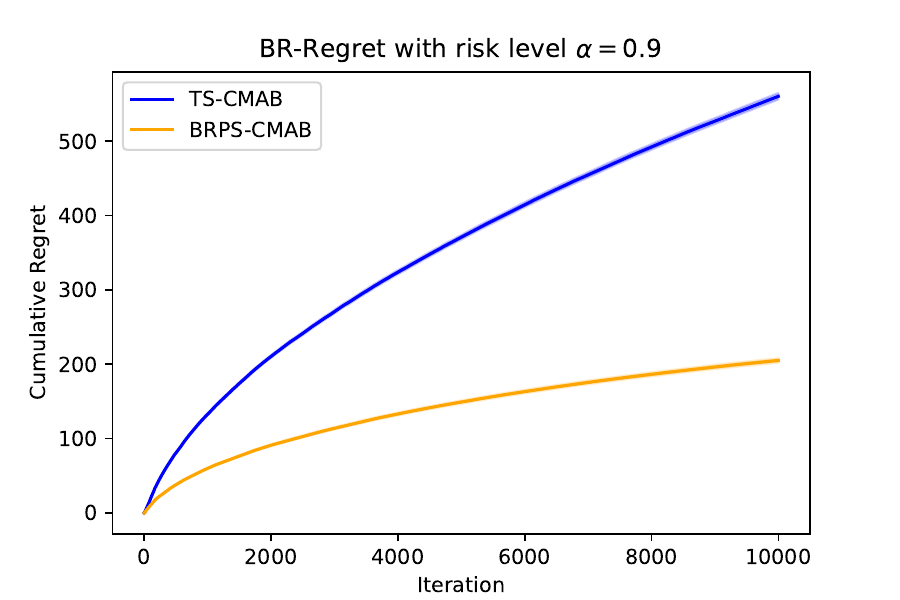}
\end{subfigure}
\end{figure}

Next, we explore the impact of the risk level on the performance of BRPS-CMAB. In Figure \ref{fig: bandit BR-Regret}, we present the results using BR-Regret, where the risk level $\alpha$ varies in $\{0.5, 0.8, 0.9\}$.

Figure \ref{fig: bandit BR-Regret} illustrates that when using BR-Regret as the performance metric, BRPS-CMAB consistently outperforms TS-CMAB across three distinct risk levels. This observation is expected since BR-Regret defines the optimal value function as the Bayesian risk-averse value function with a specified risk level—an objective that BRPS-CMAB is designed to achieve.  
Additionally, as the risk level decreases (indicating lower risk aversion), the performance gap between TS-CMAB and BRPS-CMAB narrows. This is attributed to the fact that TS-CMAB can be viewed as the risk-neutral variant of BRPS-CMAB, effectively operating at a risk level of $\alpha = 0$.

\begin{figure}[h]
    \centering
\includegraphics[width = 0.6\linewidth]{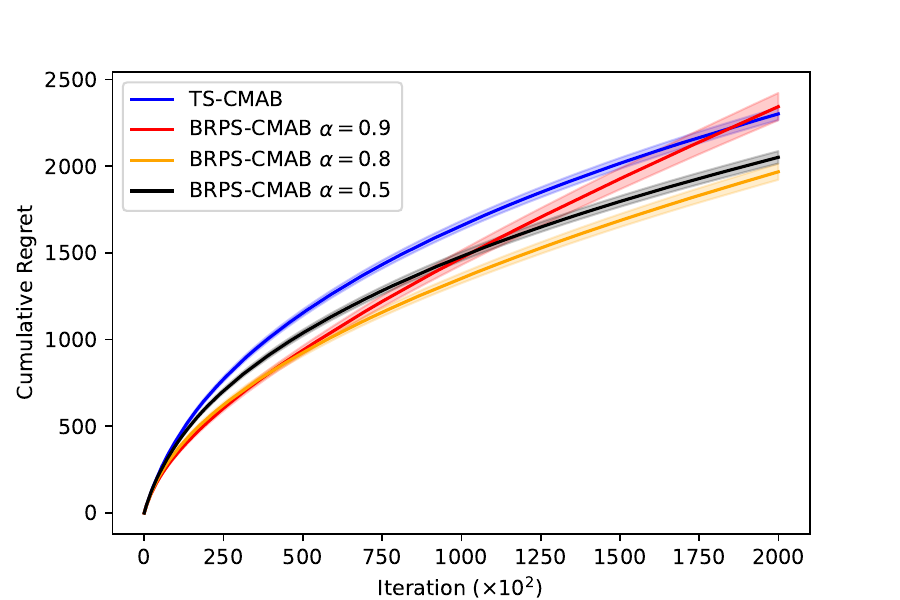}
    \caption{Comparison of conventional regret with different risk levels}
    \label{fig: bandit multi risk}
\end{figure}

Third, in Figure \ref{fig: bandit multi risk}, we compare TS-CMAB and BRPS-CMAB using the conventional regret metric.

Figure \ref{fig: bandit multi risk} shows that for $\alpha = 0.5, 0.8$, BRPS-CMAB outperforms TS-CMAB, whereas for $\alpha = 0.9$, BRCMAB eventually exhibits the highest cumulative regret. This outcome is influenced by the role of the risk level $\alpha$ in the exploitation-exploration trade-off.  
With a larger $\alpha$ (indicating a more risk-averse attitude), the algorithm tends to prioritize exploitation, adopting a safer policy that selects the arm with lower tail risk (i.e., a larger CVaR value of the reward). Notably, although $\alpha = 0.9$ results in the highest regret after $2\times 10^5$ iterations, it achieves the lowest cumulative regret in the early iterations, around $5\times 10^4$, due to its risk-averse approach when the dataset is small.

\subsection{OpenAI Benchmark problem: Frozen Lake} \label{sec: frozen lake}
In this section, we test on a variant of the OpenAI problem, a simple game namely ``\href{ https://gymnasium.farama.org/environments/toy_text/frozen_lake/}{Frozen Lake}". 
\begin{figure}[H]
    \centering
    \includegraphics[width = 0.4\linewidth]{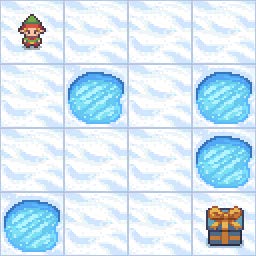}
    \caption{Frozen Lake}
    \label{fig: frozen}
\end{figure}
The game environment is a "$4\times 4$" grid world (see Figure \ref{fig: frozen}) with a goal located at the bottom-right corner, i.e., location [3,3]. There are also four holes in the ice, positioned at locations [1,1], [1,3], [2,3], and [3,0]. The game starts with the player at the top-left corner, location [0,0].  
The player can take one of four actions: "left," "right," "up," or "down" until reaching the goal. Upon reaching the goal, the game restarts with the same grid world, but the player is placed randomly in a location other than a hole or the goal, with an unknown probability.  
The lake is slippery, so the player may occasionally move perpendicular to the intended direction. For instance, when attempting to move left, the actual movement could be left, up, or down, with some unknown probability. If the player falls into a hole, they can attempt to move out. With probability $p_h$, the player moves in their intended direction. Otherwise, with probability $1-p_h$, they remain in the same location despite the attempted move. The probability $p_h$ is also unknown to the player.

For the MDP formulation (discounted, infinite-horizon MDP), the state space consists of 16 possible locations of the player, while the action space comprises the four possible moves the player can take. The reward is $1$ if the goal is reached and $0$ otherwise. The discount factor is set to $\gamma = 0.8$.  
The transition kernel depends on the probability of unintended actions due to the slippery ice, the probability $p_h$, and the distribution of the player's location when the game is restarted. All three distributions are unknown, and we impose a Dirichlet conjugate prior on each finite discrete distribution. Each action taken by the player consumes one time step (horizon), and the relocation of the player after reaching the goal also takes one horizon.  
When solving for the BRMDP, the MDP environment is sampled by drawing from the posterior distributions of the three unknown (finite-support) distributions. The true probability of perpendicular moves is set to $0.25$ for each perpendicular direction, and the distribution of player relocation follows a discrete uniform distribution with 11 support points.  
The probability $p_h$ is set to $\{0.1, 0.2, 0.6, 0.8\}$. A smaller value of $p_h$ represents a riskier environment, as the player is more likely to remain stuck in a hole for a longer time if they fall in. Conversely, a larger value of $p_h$ represents a safer environment, as the penalty for falling into a hole is reduced.

\begin{figure}[h]
    \caption{Regret Comparison for different $p_h$}
    \label{fig: fl regret}
    \begin{minipage}{.48\linewidth}
  \centering
  \includegraphics[width = \linewidth]{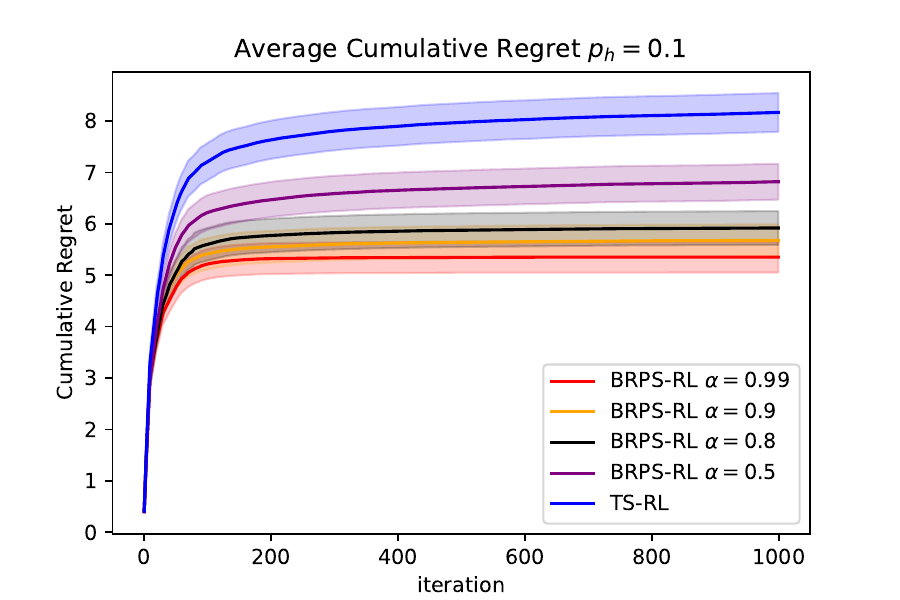}
    % \captionof{figure}
    {$p_h = 0.1$}
    % \label{fig: frozen lake p_h=0.1}
\end{minipage}
~
\begin{minipage}{.48\linewidth}
\centering\includegraphics[width = \linewidth]{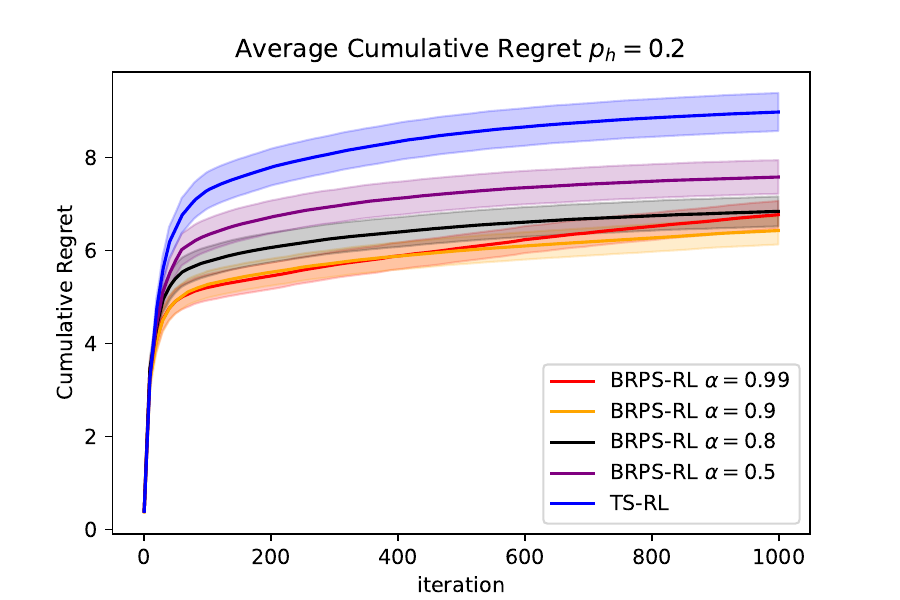}
    % \captionof{figure}
    {$p_h = 0.2$}
    % \label{fig: frozen lake p_h=0.2}
\end{minipage}

\begin{minipage}{.48\linewidth}
  \centering
  \includegraphics[width = \linewidth]{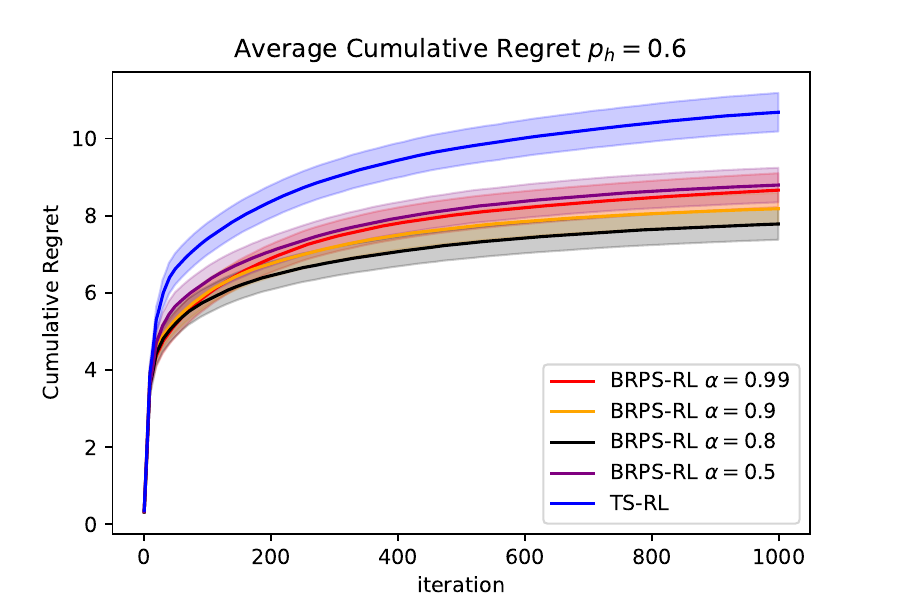}
    % \captionof{figure}
    {$p_h = 0.6$}
    % \label{fig: frozen lake p_h=0.6}
\end{minipage}
~
\begin{minipage}{.48\linewidth}
\centering\includegraphics[width = \linewidth]{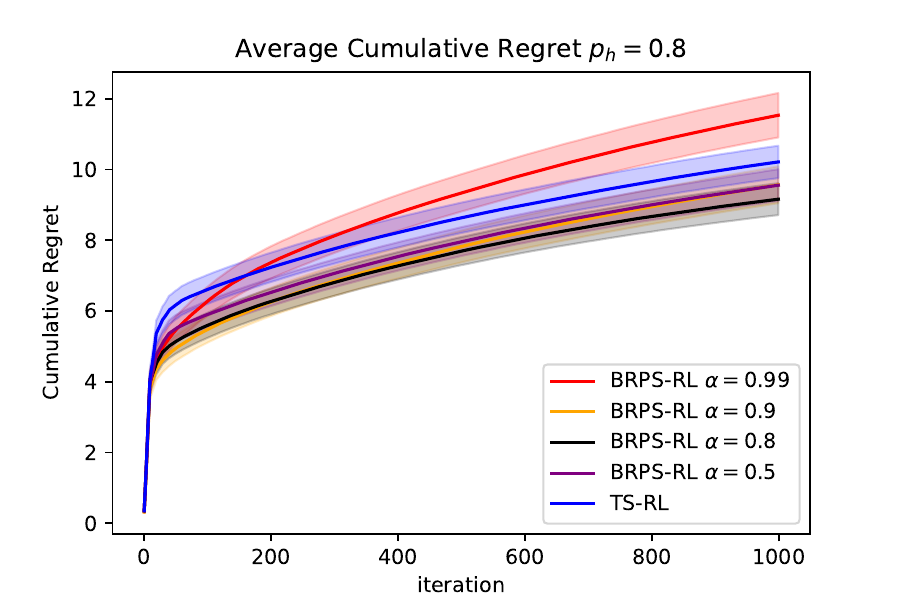}
    % \captionof{figure}
    {$p_h = 0.8$}
    % \label{fig: frozen lake p_h=0.8}
\end{minipage}

\end{figure}

We first present the results using the conventional regret metric. We evaluate the performance of Thompson Sampling for RL (see \cite{osband2013more}, denoted as TS-RL) and Algorithm \ref{alg: BRPS-RL} (denoted as BRPS-RL) with different risk levels, ranging in $\{0.5, 0.8, 0.9, 0.99\}$.  
For implementation details, we set the length of each stage to $L = 10$. To compute the optimal policy for each stage, after generating $n = \lceil \frac{1}{1-\alpha} \rceil$ samples, we apply Q-value iteration with 100 iterations. The original value function of the policy—i.e., the value function when deploying the policy in the true environment—is computed directly, since the true environment is a finite-state, finite-action MDP.  
The empirical expected regret is estimated by running 500 macro-replications. Figure \ref{fig: fl regret} presents the results for $p_h = 0.1, 0.2, 0.6, 0.8$.

% \begin{minipage}{.48\linewidth}
%   \centering
%   \includegraphics[width = \linewidth]{figure/frozen_lake/expected_regret.png}
%     \captionof{figure}{Expected Regret}
%     \label{fig: fl_regret}
% \end{minipage}
% ~
% \begin{minipage}{.48\linewidth}
% \centering\includegraphics[width = \linewidth]{figure/frozen_lake/stage_wise_std.png}
%     \captionof{figure}{Iteration-wise standard error (log)}
%     \label{fig: fl_std}
% \end{minipage}
\subsubsection*{Result:} 
\begin{enumerate}
    \item In Figure \ref{fig: fl regret}, BRPS-RL with risk levels $0.5, 0.8, 0.9$ outperforms Thompson Sampling due to its risk aversion to model uncertainty and preference for exploitation. This demonstrates its efficiency in online reinforcement learning when the risk level is appropriately set.

    \item Across different values of $p_h$, the optimal choice of risk levels (among the four considered) are $0.99, 0.9, 0.8, 0.8$ for $p_h = 0.1, 0.2, 0.6, 0.8$, respectively, which decreases as $p_h$ increases.  
This behavior arises because, when implementing BRPS-RL with a high risk level, the algorithm prioritizes avoiding holes, even when the player has little information about the true value of $p_h$. When $p_h$ is small, the player tends to remain stuck in a hole for a long time on average. Thus, avoiding the hole is more favorable, and BRPS-RL with a larger risk level performs better.  
However, when $p_h$ is large, the player can easily escape after falling into a hole. In this case, taking an action that leads to a hole may be beneficial if it provides a shorter path. Moreover, falling into the hole facilitates learning the true value of $p_h$. This exploration can be advantageous when $p_h$ is large, as the player can safely determine that falling into a hole poses minimal risk.
. 
    % Generally speaking, when the underlying environment is more adversarial (i.e., the optimal value function under the true parameter is relatively small compared to other parameters considered), Algorithm \ref{alg: BRPS-RL} with a larger risk level will perform better since 
    \item Although setting the risk level too high can lead to large cumulative regret in the long term, it can still be beneficial in the short term. This is because the algorithm avoids taking actions that may be significantly worse than others, a phenomenon similar to what is observed in the CMAB example.  
This insight suggests that even if the environment is safe to explore (as in the case of larger $p_h$), a higher risk level can still be favorable when the available data is limited—in other words, when model uncertainty is high. Such scenarios are common in offline reinforcement learning, though our focus remains on online reinforcement learning.

    \item Beyond regret as a performance metric, computational time is also of interest. Roughly speaking, the computational effort can be measured by the number of (parametrized) environments sampled from the posterior distribution at each stage. According to our algorithm, this number is essentially the inverse of $1-\alpha$.  
Thompson Sampling can be viewed as a special case where the risk level is $\alpha = 0$. Table \ref{table: running time} presents the average computational time (in seconds) for different algorithms, along with their corresponding standard errors.

    \renewcommand{\arraystretch}{1.5}
    \begin{table}[h] 
        \centering
        \begin{tabular}{c|c|c|c|c|c}
        \hline
             \multirow{ 2}{*}{Algorithms} &  {BRPS-RL} & {BRPS-RL}  & {BRPS-RL} & {BRPS-RL} & \multirow{ 2}{*}{TS-RL}\\ 
             & $\alpha = 0.99$&$\alpha = 0.9$&$\alpha = 0.8$&$\alpha = 0.5$&\\
             \hline
             Computational time (s)&$253.43(0.86)$ &$33.88(1.46)$ &$20.95(0.95)$ &$13.01(0.09)$ &$10.45(0.10)$ \\
             \hline
        \end{tabular}
        \vskip 0.1in
        \caption{Computational time of different algorithms}
    \label{table: running time}
\end{table}
    \renewcommand{\arraystretch}{1}
    
   Table \ref{table: running time} indicates that despite the improved performance with a smaller risk level, solving an optimal policy requires more computational effort. For instance, when setting the risk level to $0.99$, the regret is only slightly better than that with a risk level of $0.9$ for $p_h = 0.1$, but the computation time is approximately eight times longer.  
In practice, the computational cost should also be taken into account when selecting the risk level.

\end{enumerate}

\subsection{Verification of asymptotic normality in Theorem \ref{thm: normality BRMDP}.} 

In this section, we numerically study BRMDP from the modeling perspective. Specifically, we investigate the behavior of the Bayesian risk value function $V^{\psi,\pi}$ under varying amounts of data used to estimate the posterior distribution $\psi$.  
For this analysis, we continue utilizing the Frozen Lake example but adopt a different approach than in Section \ref{sec: frozen lake}. Previously, we estimated the Bayesian posterior distribution for probabilities such as the likelihood of a wrong action due to slippery ice, the relocation distribution, and $p_h$. Now, we aim to directly estimate the transition probabilities $\calP_{s,a}$ for all state-action pairs to validate the exact limiting distribution proposed in Theorem \ref{thm: normality BRMDP}.

In this evaluation, data are collected along a single trajectory governed by the behavior policy deemed optimal ($\pi^*$) in the actual environment with $p_h = 0.1$. The optimal policy $\pi^*$ is outlined in Table \ref{tab: optimal policy}, with red arrows indicating the locations of holes.  
The average number of data points, $\Bar{n}_s$, as stated in Theorem \ref{thm: normality BRMDP}, corresponds to the stationary distribution under $\pi^*$.

To compute the Bayesian risk value function $V^{\psi,\pi^*}$, we generate $n = 5000$ samples from the posterior $\psi$, which is estimated from a dataset of size $N$. We then perform 100 steps of value iteration to derive the value function.  
We denote the Bayesian risk value function obtained from a dataset of size $N$ as $V_N$ and the optimal value function under the true environment as $V^*$.

We illustrate the empirical density function of $\sqrt{N}(V_N - V^*)[s_{10}]$ by conducting 500 replications. Additionally, we plot the density function of the theoretical limiting distribution as described in Theorem \ref{thm: normality BRMDP}.  
A Quantile-Quantile (Q-Q) plot of these 500 replications will also be provided. The dataset sizes $N$ considered in our experiments range from $\{10^4, 5\times 10^4, 10^5, 10^6\}$.
 %This comprehensive analysis aims to elucidate the dynamics and precision of the Bayesian risk value function under varying data volumes.
\begin{table}[h] \large
    \centering
    \begin{tabular}{c|c|c|c}
  $\downarrow$ &$\rightarrow$ &$\downarrow$ &$\leftarrow$ \\ \hline
$\downarrow$ &$\textcolor{red}{\downarrow}$ &$\downarrow$ &$\textcolor{red}{\leftarrow}$ \\ \hline
$\rightarrow$ &$\downarrow$ &$\downarrow$ &$\textcolor{red}{\downarrow}$ \\ \hline
$\textcolor{red}{\rightarrow}$ &$\rightarrow$ &$\rightarrow$ &$*$ 
    \end{tabular}
    \caption{Optimal policy}
    \label{tab: optimal policy}
\end{table}

\begin{figure}[h]
    \caption{Density function}
    \label{fig: pdf normality}

\begin{minipage}{.48\linewidth}
  \centering
  \includegraphics[width = \linewidth]{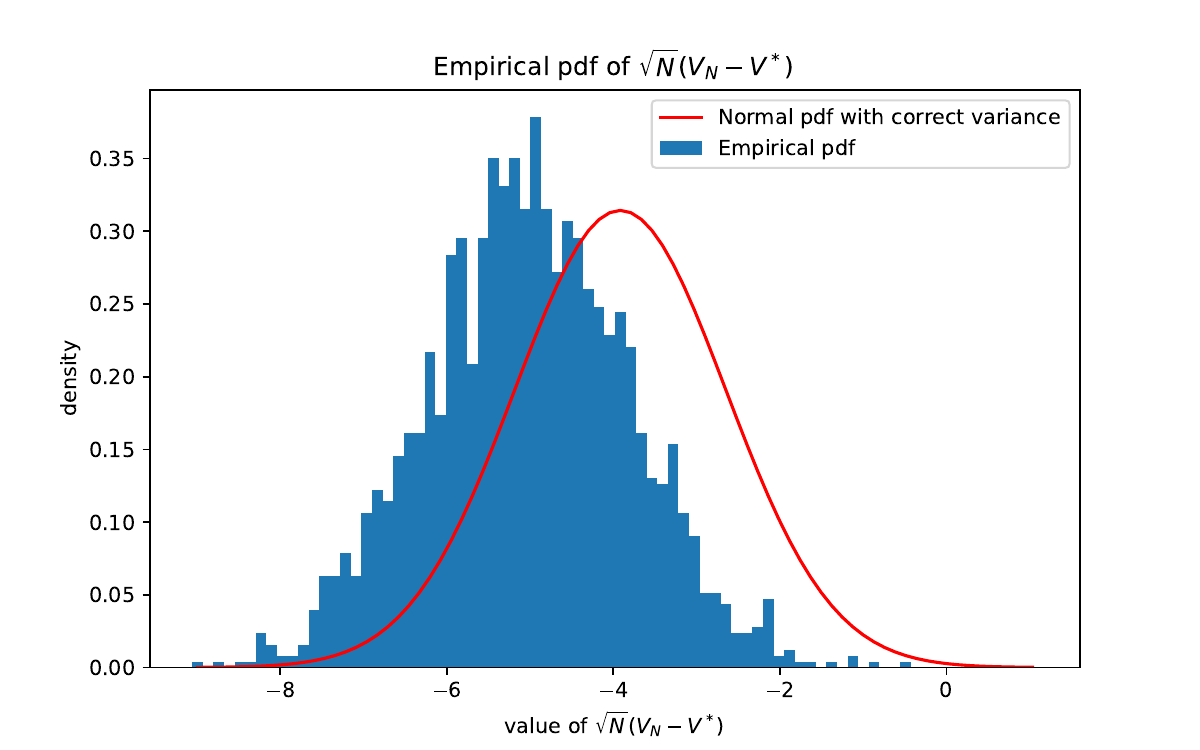}
    % \captionof{figure}
    {$N = 10^4$}
    % \label{fig: normality N=10^4}
\end{minipage}
~
\begin{minipage}{.48\linewidth}
\centering\includegraphics[width = \linewidth]{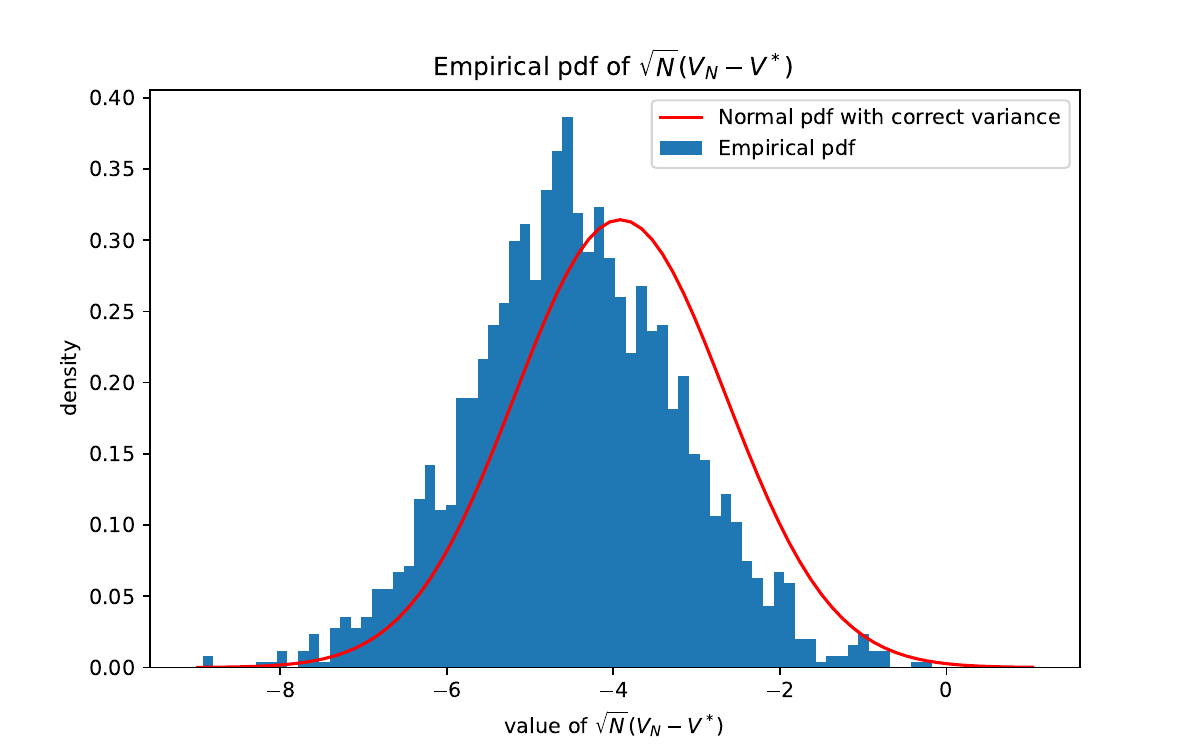}
    % \captionof{figure}
    {$N = 5\times 10^4$}
    % \label{fig: normality N=5*10^4}
\end{minipage}

\begin{minipage}{.48\linewidth}
  \centering
  \includegraphics[width = \linewidth]{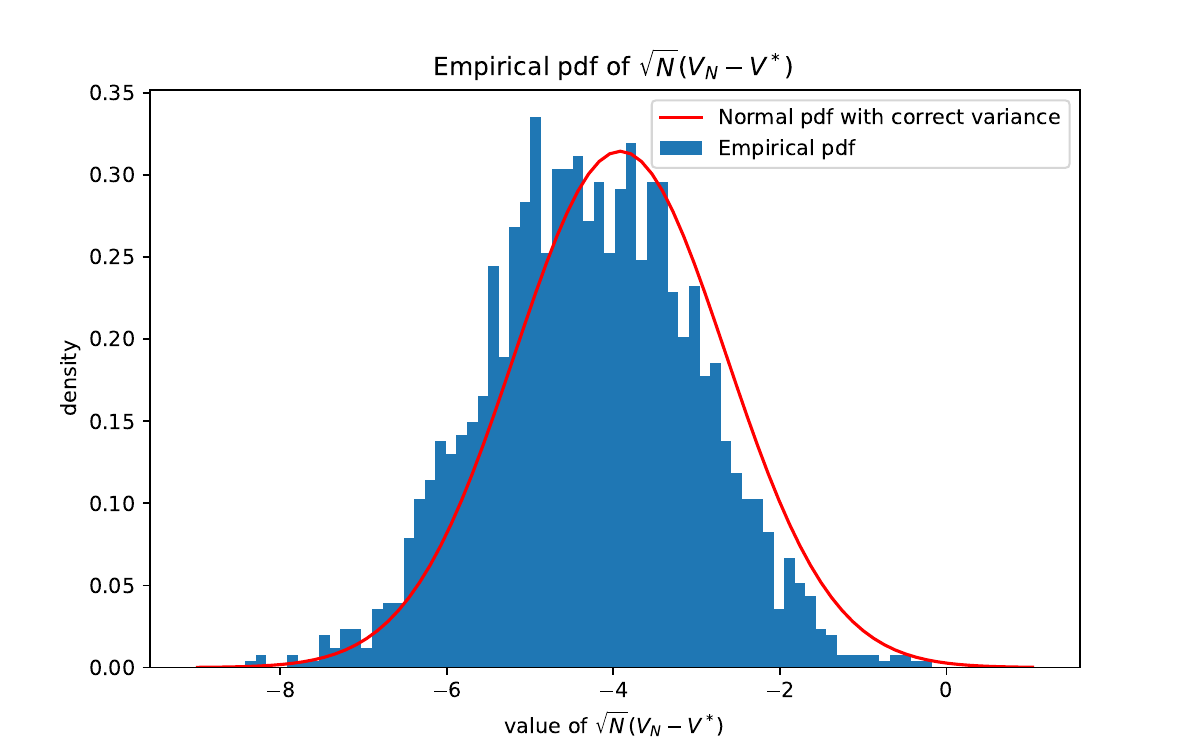}
    % \captionof{figure}
    {$N = 10^5$}
    % \label{fig: normality N=10^5}
\end{minipage}
~
\begin{minipage}{.48\linewidth}
\centering\includegraphics[width = \linewidth]{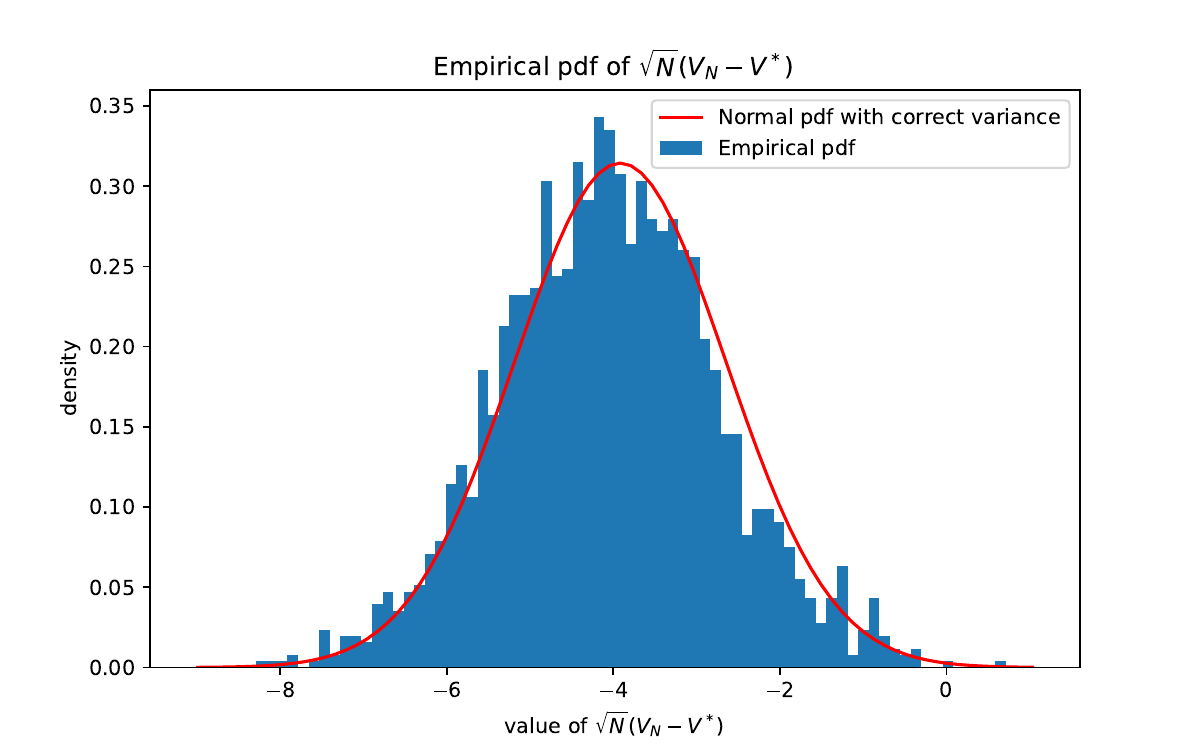}
    % \captionof{figure}
    {$N =  10^6$}
    % \label{fig: normality N=10^6}
\end{minipage}
\end{figure}

\begin{figure}[h]
    \caption{Q-Q plot}
    \label{fig: qq plot}

    \begin{minipage}{.48\linewidth}
  \centering
  \includegraphics[width = \linewidth]{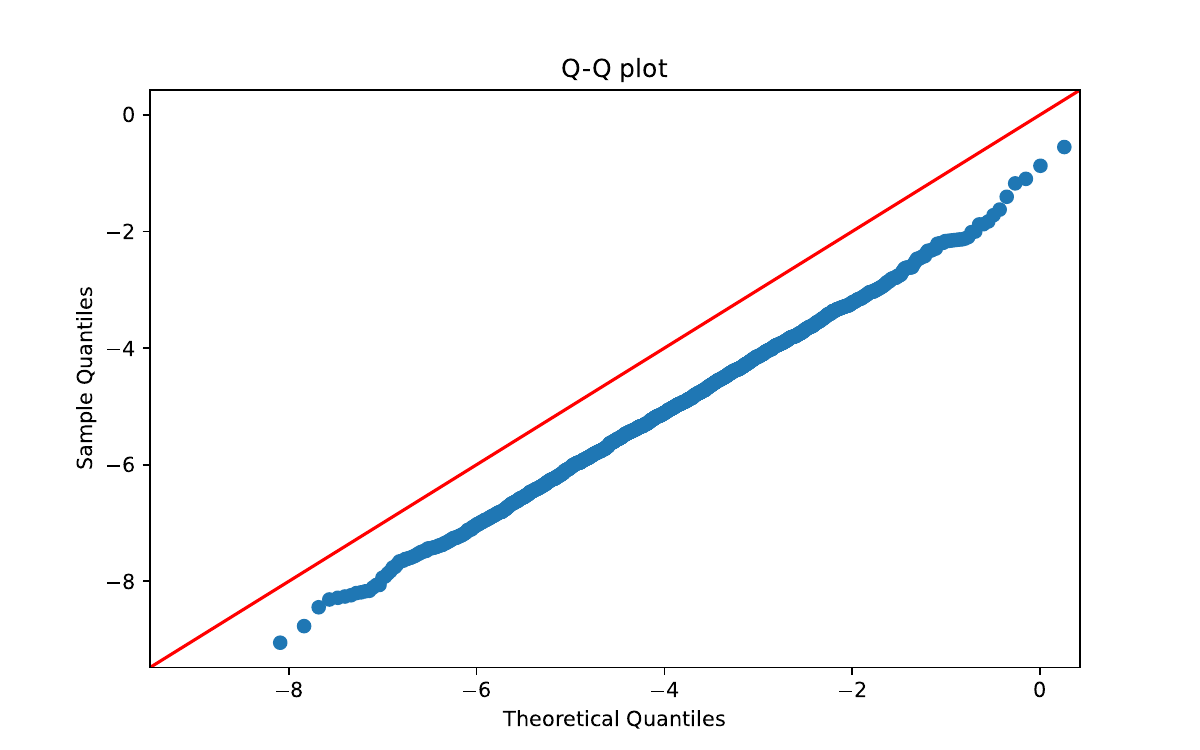}
    % \captionof{figure}
    {$N = 10^4$}
    % \label{fig: qq N=10^4}
\end{minipage}
~
\begin{minipage}{.48\linewidth}
\centering\includegraphics[width = \linewidth]{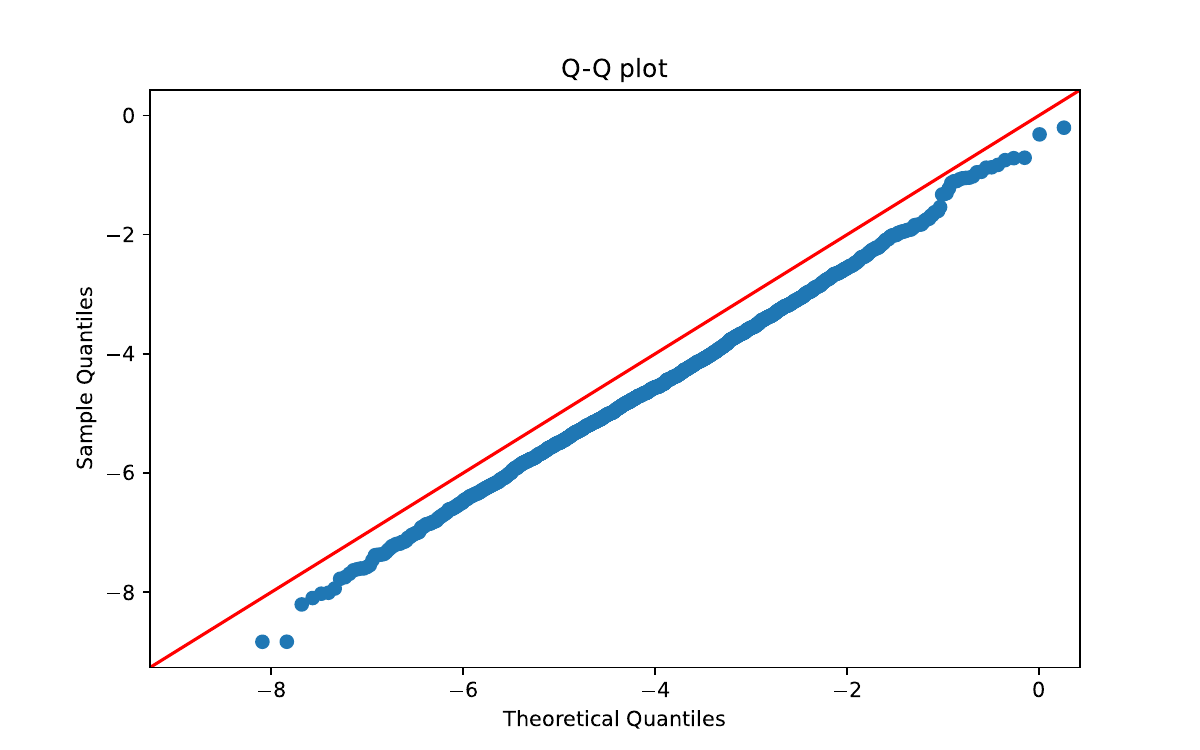}
    % \captionof{figure}
    {$N = 5\times 10^4$}
    % \label{fig: qq N=5*10^4}
\end{minipage}

\begin{minipage}{.48\linewidth}
  \centering
  \includegraphics[width = \linewidth]{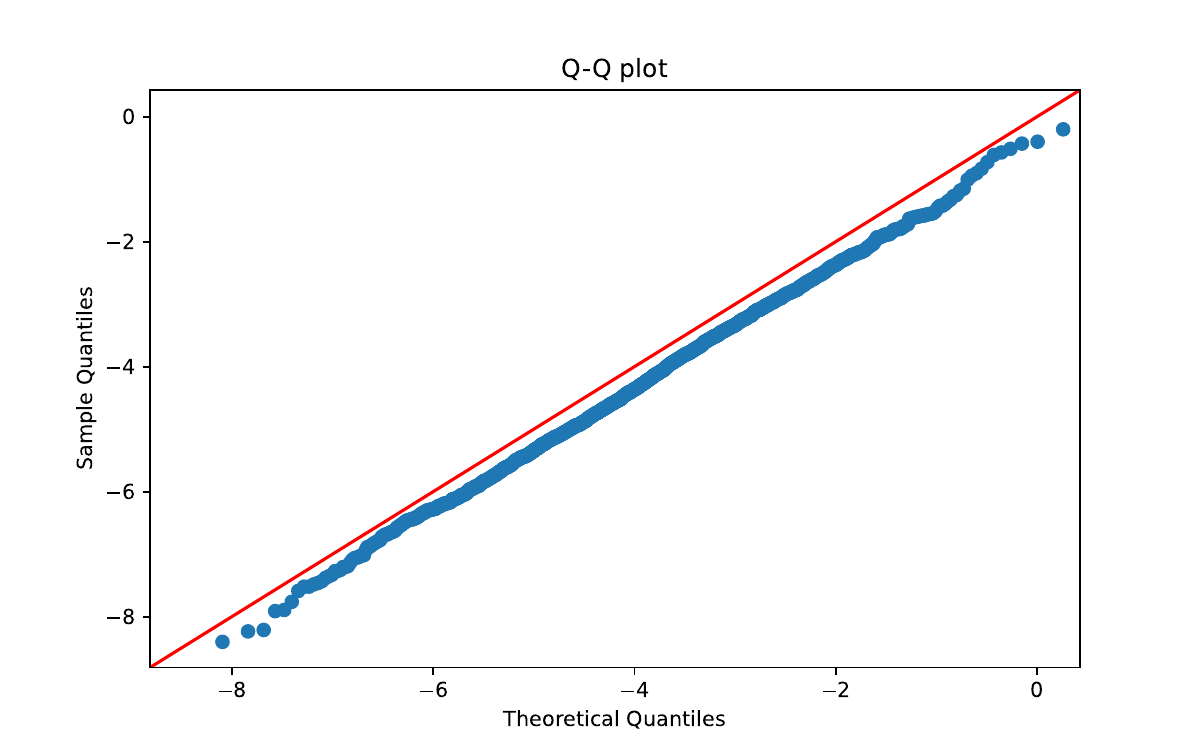}
    % \captionof{figure}
    {$N = 10^5$}
    % \label{fig: qq N=10^5}
\end{minipage}
~
\begin{minipage}{.48\linewidth}
\centering\includegraphics[width = \linewidth]{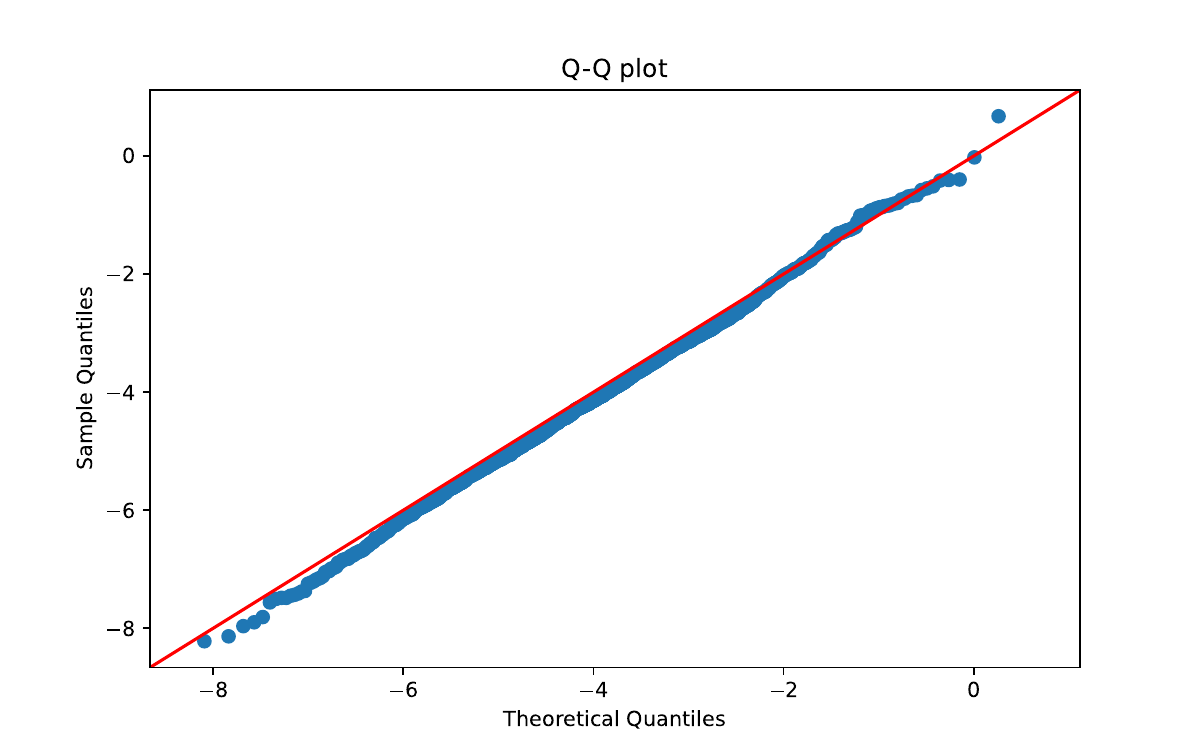}
    % \captionof{figure}
    {$N =  10^6$}
    % \label{fig: qq N=10^6}
\end{minipage}
\end{figure}

Both Figure \ref{fig: pdf normality} and Figure \ref{fig: qq plot} indicate that as the dataset size $N$ approaches infinity, the normalized error between the Bayesian risk value function and the true optimal value function, $\sqrt{N}(V_N - V^*)$, converges to a normal distribution with the mean and variance specified in Theorem \ref{thm: normality BRMDP}, thereby verifying its correctness.

\section{Conclusion}
In this paper, we study Bayesian risk-averse reinforcement learning (RL) from both the modeling and an algorithmic perspective.  
From the perspective of model formulation, the value function of the Bayesian risk Markov Decision Process (BRMDP) asymptotically equals the value function of the unknown true Markov Decision Process plus a negative error term (on average), which increases with stronger risk aversion (as determined by the user-specified risk level) but decreases as more data become available. This indicates that BRMDP exhibits flexibility in risk management and can adaptively hedge against parametric uncertainty throughout the data collection process.  
From the algorithmic perspective, we provide two online procedures using posterior sampling: one for the general RL setting and another for the contextual multi-arm bandit (CMAB) setting. Compared with the classic Thompson Sampling algorithm (also known as posterior sampling), our proposed algorithms allow for a more flexible risk attitude and demonstrate more robust performance due to their risk aversion to model uncertainty, particularly when the amount of data is small.

\acks{The authors gratefully acknowledge the support by the Air Force Office of Scientific Research under Grant FA9550-22-1-0244, the National Science Foundation under Grant NSF-DMS2053489 and the NSF AI Institute for Advances in Optimization under Grant
NSF-2112533.}


\newpage

\small
\bibliography{ref}

\appendix
\include{appendix}

\end{document}

%% file: appendix.tex
% \appendixpage
% \ECSwitch
\section{Definition of Coherent Risk Measure} \label{sec: coherent risk measure}
A risk measure $\rho$ is defined as a mapping that maps a random variable to a real number. Let $Z_1, Z_2$ be two random variables defined in the same probability space. A risk measure is called a coherent risk measure if it satisfies the following properties:
\begin{enumerate} 
    \item (Normalization) $\rho(0) = 0$.
    \item (Monotonicity) If $Z_1\le Z_2$ almost surely, then $\rho(Z_1) \le \rho(Z_2)$.
    \item (Sup-additivity) $\rho(Z_1+Z_2) \ge \rho(Z_1) + \rho(Z_2)$.
    \item (Positive homogeneity) $\rho(\alpha Z_1) = \alpha \rho(Z_1)$ for $\alpha \ge 0$.
    \item (Translation invariance) $\rho(Z_1 + a) = \rho(Z_1) + a$ for $a\in \mathbb{R}$.
\end{enumerate}

\section{Dirichlet Posterior on State-action Pair} \label{sec: Dirichlet}
A Dirichlet distribution is parameterized by a count vector $\psi=\left(\psi_1, \ldots, \psi_k\right)$, where $\psi_i \geq 1$, such that the density of probability distribution $p=\left(p_1, \ldots, p_k\right)$ is defined as $f(p \mid \psi) \propto \prod_{i=1}^k p_i^{\psi_i-1}$. For each $(s,a) \in \calS \times \calA$, we impose a Dirichlet prior with parameter $\psi_{s,a} = (\psi_{s,a}(s'))_{s'\in\calS}$ on the unknown transition probability $p_{s,a} = (p_{s,a}(s'))_{s'\in\calS}$. After we observe the transition $(s,a,s')$ for $o_{s,a,s'}$ times $\forall s'\in\calS$, the posterior distribution of $p_{s,a}$ is also a Dirichlet distribution with parameter $\psi_{s,a} + \mathbf{o}_{s,a} = (\psi_{s,a}(s')+o_{s,a,s'})_{s'\in\calS}$, where $\mathbf{o}_{s,a} = (o_{s,a,s'})_{s'\in \calS}$.

% \iffalse
\section{Online Bayesian Risk-Averse Posterior Sampling with Exploration bonus} \label{sec: regret RL}
{ In this section, we present the online Bayesian risk-averse value iteration algorithm in Algorithm \ref{alg: Online BRVI}, where value iteration is continuously applied with an additional bonus term to update the estimate of the BRMDP value function.  
We restrict our focus to transition uncertainty and directly model the unknown transition probabilities using a Dirichlet conjugate prior. The detailed algorithm is provided in Algorithm \ref{alg: Online BRVI}.

\begin{algorithm}[H]
   \caption{Online Bayesian risk-averse Value Iteration}
   \label{alg: Online BRVI}
\begin{algorithmic}
   \STATE {\bfseries Input:} State space $\calS$, action space $\calA$, reward function $r$, initial state $s_0$, termination stage $T$, 
   %observation batch size $\{n(t)\}_{t=1}^T$ {\color{red} (may not treat as input parameter)}, 
     risk level $\alpha$. 
%   \REPEAT
   \STATE { \bfseries Initialize} $\psi^0_{s,a}(s') \leftarrow 1, \forall s,a,s' $, $Q_0(s,a) = V_0(s) \leftarrow \frac{1}{1-\gamma}$, $\forall s,a$, sample size $n\leftarrow \lceil \frac{1}{1-\alpha}\rceil$.
   \FOR{$t=0$ {\bfseries to} $T$}
   \STATE Let $\pi_t(\cdot) \leftarrow \arg\max_{a\in\calA} Q_t(\cdot,a)$, take action $a_t \leftarrow \pi_t(s_t)$ and receive next state $s_{t+1} \sim \calP(\cdot|s_t,a_t)$.
   \STATE Set $\psi^{t+1}_{s,a}(s') \leftarrow \psi^{t}_{s,a}(s'), \forall s,a,s'$, $\psi^{t+1}_{s_t,a_t}(s_{t+1}) \leftarrow \psi^{t}_{s_t,a_t}(s_{t+1}) + 1$.
   
   \FOR{{\bfseries all} $(s,a) \in \calS\times\calA$}
   \STATE Generate $p_1,p_2,\ldots,p_{n} \sim \psi^t_{s,a}$.
   % \STATE Set $X_i \leftarrow \sum_{s'\in\calS} p_i(s') V_t(s'), i \in [n]$.
   \STATE Set $\widehat{\rho}_{s,a}^t V_t  \leftarrow \widehat{C}_\alpha\left( \left\{\sum_{s'\in\calS} p_i(s') V_t(s')\right\}_{i=1}^n\right) $
   \STATE Let UCB$_t(s,a) \leftarrow \frac{3\gamma}{1-\gamma} \sqrt{\frac{|\calS|^2}{2(N_{s,a}^t + 1)}\log{\frac{4(n+1)|\calS|^2|\calA|T}{\delta}}}$
   \STATE $Q_{t+1}(s,a) \leftarrow \min\left\{Q_t(s,a),  r(s,a) + \gamma \widehat{\rho}_{s,a}^t V_t + \text{UCB}_t(s,a)  \right\}$
   \STATE $V_{t+1} \leftarrow \max_{a\in\calA} Q_{t+1}(s,a)$
   \ENDFOR
   
   \ENDFOR
%   \UNTIL{$noChange$ is $true$}
\end{algorithmic}
\end{algorithm}
Unlike Algorithm \ref{alg: BRPS-RL}, where selecting an action requires solving the current BRMDP model to optimality, Algorithm \ref{alg: Online BRVI} only requires a single step of value iteration using the Bayesian risk-averse Bellman operator $\calT^\psi$.  
However, to run Algorithm \ref{alg: Online BRVI}, a pre-specified risk level $\delta$ and the maximum stage $T$ are required.

For Algorithm \ref{alg: Online BRVI}, we define regret in the traditional sense as:
\begin{equation}
\textbf{Regret}(T) = \sum_{t=1}^T V^*(s_t) - V^\pi_t(s_t),
\end{equation}
where $\pi = (\pi_1, \pi_2, \ldots)$ is any non-stationary policy, and  
\begin{equation}
V_t^\pi(s) = \mathbb{E} \left[\sum_{i=0}^{\infty} \gamma^i r\left(s_{t+i}, a_{t+i}\right) \mid s_t = s \right].
\end{equation}
This definition of regret is closely related to the concept of ``complexity of exploration" and is commonly used in the literature on discounted reinforcement learning (see \cite{dong2019q,kakade2003sample,lattimore2012pac}).  

We present the following result on the performance of Algorithm \ref{alg: Online BRVI}.

\begin{theorem} \label{thm: UCBVI}
    With Probability $1-\delta$, the regret of Algorithm \ref{alg: Online BRVI} satisfies
    \begin{align*}
        \text{Regret}(T) \le &\frac{6\gamma}{(1-\gamma)^2}\sqrt{\frac{|\calS|^2}{2}\log{\frac{4(3-2\alpha)|\calS|^2|\calA|T}{(1-\alpha)\delta}}} \min\{ |\calS|\sqrt{T+1},\sqrt{|\calS|(T+1)\ln{(T+1)}}\}\\
     & + \gamma\frac{2|\calS|+2}{(1-\gamma)^2} + \frac{2\gamma}{(1-\gamma)^2}\sqrt{2T\ln{\frac{2}{\delta}}}.
    \end{align*}
In particular, set $\delta = \frac{1}{T}$, we obtain
\begin{align*}
        \mathbb{E} \left[\text{Regret}(T)\right] \le &\frac{6\gamma}{(1-\gamma)^2}\sqrt{\frac{|\calS|^2}{2}\log{\frac{4(3 - 2\alpha)|\calS|^2|\calA|T^2}{1-\alpha}}} \min\{ |\calS|\sqrt{T+1},\sqrt{|\calS|(T+1)\ln{(T+1)}}\}\\
     & + \gamma\frac{2|\calS|+2}{(1-\gamma)^2} + \frac{2\gamma}{(1-\gamma)^2}\sqrt{2T\ln{2T}} + \frac{1}{1-\gamma}.
    \end{align*}

\end{theorem}
}
% \fi
\section{Techinal Proof}
\subsection{Proof of Theorem \ref{thm: normality BRMDP}}
% \begin{lem} \label{lem: Bernstein-von}
%     (Bernstein-von Mises). Let the experiment $\left(P_\theta: \theta \in \Theta\right)$ be differentiable in quadratic mean at $\theta_0$ with nonsingular Fisher information matrix $I_{\theta_0}$, and suppose that for every $\varepsilon>0$ there exists a sequence of tests $\psi_n$ such that
% $$
% P_{\theta_0}^n \psi_n \rightarrow 0, \quad \sup _{\left\|\theta-\theta_0\right\| \geq \varepsilon} P_\theta^n\left(1-\psi_n\right) \rightarrow 0
% $$
% Furthermore, let the prior measure be absolutely continuous in a neighborhood of $\theta_0$ with a continuous positive density at $\theta_0$. Then the corresponding posterior distributions satisfy
% $$
% \left\|P_{\sqrt{n}\left(\bar{\Theta}_n-\theta_0\right) \mid X_1, \ldots, X_n}-N\left(\Delta_{n, \theta_0}, I_{\theta_0}^{-1}\right)\right\| \stackrel{P_{\theta_0}^n}{\rightarrow} 0
% $$
% \end{lem} 

\begin{lem} \label{lem: concentration expon}
    Assume the prior Dirichlet distribution is set to $\psi_{s,a}(s') = 1$, $\forall s,a,s'$. Given observed counting data $\{N_{s,a}(s')\}$ and $\bar{N}_{s,a} = \sum_{s'}N_{s,a}(s')$. Let $\bar{\psi}_{s,a}(s') = \frac{\psi_{s,a}(s')}{\sum_{s''} \psi_{s,a}(s'')} = \frac{N_{s,a}(s')+1}{\sum_{N_{s,a}(s'')} + |\calS|}$. Then we have for $\epsilon > \frac{|\calS|}{\bar{N}_{s,a}}$
    $$ \mathbb{P}\left(  \|\bar{\psi}_{s,a} - P_{s,a}\|_\infty 
 > \epsilon \right)\le 2 |\calS|\mathrm{e}^{-2(\epsilon-\frac{|\calS|}  {\bar{N}_{s,a}})^2\bar{N}_{s,a}}.$$
\end{lem}
\proof{}
   $$ \mathbb{P}\left(  \|\bar{\psi}_{s,a} - P_{s,a}\|_\infty 
 > \epsilon \right) =  \mathbb{P}\left( \bigcup_{s' \in \calS}
 |\bar{\psi}_{s,a}(s') - P_{s,a}(s')| 
 > \epsilon \right) \le \sum_{s' \in \calS} \mathbb{P}\left(  |\bar{\psi}_{s,a}(s') - P_{s,a}(s')| 
 > \epsilon \right). $$ 
 For each $s'$, let $s_i$ denote the $i^{th}$ 
 transition observation for state-action pair $(s,a)$.Then $N_{s,a}(s') = \sum_{i=1}^{\bar{N}_{s,a}} \mathbf{1}_{s_i = s}$. By Hoeffding inequality, 
 $$ \mathbb{P} (|\frac{N_{s,a}(s')}{\bar{N}_{s,a}} - \calP_{s,a}(s')| > \epsilon ) \le 2\mathrm{e}^{-2\bar{N}_{s,a}\epsilon^2}.$$
 Further notice 
 $$\bar{\psi}_{s,a}(s') - \frac{N_{s,a}(s')} {\bar{N}_{s,a}} = \frac{\bar{N}_{s,a} - |\calS|N_{s,a}(s')}{(\bar{N}_{s,a} +|\calS|)\bar{N}_{s,a}}. $$
 We have $$ \frac{\bar{N}_{s,a} - |\calS|N_{s,a}(s')}{(\bar{N}_{s,a} +|\calS|)\bar{N}_{s,a}} \le \frac{\bar{N}_{s,a}}{(\bar{N}_{s,a} + \calS)\bar{N}_{s,a}} \le \frac{|\calS|}{\bar{N}_{s,a}}.$$
And 
$$ \frac{\bar{N}_{s,a} - |\calS|N_{s,a}(s')}{(\bar{N}_{s,a} +|\calS|)\bar{N}_{s,a}} \ge -\frac{ |\calS|N_{s,a}(s')}{(\bar{N}_{s,a} +|\calS|)\bar{N}_{s,a}} \ge -\frac{|\calS|}{\bar{N}_{s,a}}.$$
Hence 
$$ \mathbb{P}\left(  |\bar{\psi}_{s,a}(s') - P_{s,a}(s')| 
 > \epsilon \right) \le \mathbb{P}\left(  |\frac{N_{s,a}(s')}{\bar{N}_{s,a}} - P_{s,a}(s')| 
 > \epsilon -\frac{|\calS|}{\bar{N}_{s,a}}\right) \le 2\mathrm{e}^{-2\left(\epsilon-\frac{|\calS|}{\bar{N}_{s,a}}\right)^2\bar{N}_{s,a}}.$$
\hfill $\blacksquare$

\begin{lem} \label{lem: using chebyshev}
    Given $\|\bar{\psi}_{s,a}-\calP_{s,a}\|_\infty < \epsilon < 1-\alpha$, we have for $\epsilon < \delta < 1-\alpha$ and an arbitrary vector $V$,
    $$\left|\cvar_\alpha^{\psi_{s,a}}\left((p^\top-\calP_{s,a})V^\pi \right)\right| \le |\calS|\|V\|_\infty \{ \delta + 2\frac{|\calS|}{1-\alpha}\frac{(\delta-\epsilon)^2}{\bar{N}_{s,a}}\}$$
\end{lem}

\proof{}
    Since $p\sim$ Dirichlet($\psi_{s,a}$), the marginal distribution of $p(s')$ has variance less than $\frac{1}{\bar{N}_{s,a}}$. Then with a union bound we know 
    \begin{equation*}
        \mathbb{P}\left(\|p-\bar{\psi}_{s,a} \|_\infty \ge \delta \right) \le |\calS| \frac{\delta^2}{\bar{N}_{s,a}}.   \end{equation*}
        Given $\|\bar{\psi}_{s,a}-\calP_{s,a}\|_\infty < \epsilon < 1-\alpha$ and $\epsilon < \delta$,
        we have 
        \begin{align*}
            \mathbb{P} \left(\|p - \calP_{s,a}\|_\infty \ge \delta \right)
            \le  \mathbb{P} \left(\|p - \bar{\psi}_{s,a}\|_\infty \ge \delta - \epsilon \right) 
            \le  |\calS| \frac{(\delta-\epsilon)^2}{\bar{N}_{s,a}}.
        \end{align*}
        Let $U = \{p| \|p - \calP_{s,a}\|_\infty \ge \delta \}$, and $W = \{p| (p^\top-\calP_{s,a})V^\pi \le \text{VaR}_{1-\alpha}^{\psi_{s,a}}((p^\top-\calP_{s,a})V^\pi) \}$. Then we can write 
        \begin{align*}
            \cvar_\alpha^{\psi_{s,a}}\left((p^\top-\calP_{s,a})V^\pi \right) 
            =&\frac{1}{1-\alpha} \left\{ \mathbb{E}_{W\cap U} \left[(p^\top-\calP_{s,a})V^\pi\right] +  \mathbb{E}_{W\cap U^c} \left[(p^\top-\calP_{s,a})V^\pi\right]\right\}\\
            \le &\frac{1}{1-\alpha} \left\{ 1-\alpha \delta |\calS| \|V\|_\infty + |\calS| \frac{(\delta-\epsilon)^2}{\bar{N}_{s,a}} (2|\calS|\|V\|_\infty)\right\}\\
            =&|\calS| \|V\|_\infty \{ \delta + 2\frac{|\calS|}{1-\alpha}\frac{(\delta-\epsilon)^2}{\bar{N}_{s,a}}\}
        \end{align*}
\hfill $\blacksquare$

 \noindent\textbf{Proof of Theorem \ref{thm: normality BRMDP}:}
    Let $\mathcal{T}^{\psi,\pi}$ be the Bellman Operator such as 
    $$\calT^{\psi,\pi} V(s) = r(s,\pi(s)) + \cvar_{\alpha}^\psi(\mathbb{E}[V(s')]). $$
    Then we know 
    $V^{\psi,\pi} = \calT^{\psi,\pi} V^{\psi,\pi}$ and $V^\pi(s) = \calT^\pi V:= r(s,\pi(s)) + \mathcal{P}_{s,\pi(s)} V^\pi$.

Write
\begin{align*}
    V^{\psi,\pi} - V^\pi 
    =&\calT^{\psi,\pi} V^{\psi,\pi} - \calT^\pi V^\pi\\
    =& \underbrace{\left(\calT^{\psi,\pi} V^{\pi}  - \calT^\pi V^{\pi}  \right)}_{(I)} + \underbrace{\calT^\pi \left( V^{\psi,\pi} - V^\pi\right)}_{(II)} +\underbrace{\left[\left( \calT^{\psi,\pi}V^{\psi,\pi} - \calT^\pi V^{\psi,\pi}\right) - \left(  \calT^{\psi,\pi} V^\pi - \calT^\pi V^\pi\right) \right]}_{(III)}.
\end{align*}
The second term $(II) = \gamma \mathcal{P}_\pi (V^{\psi,\pi} - V^{\pi})$.
The first term $(I)$ can be written as 
\begin{align*}
    \calT^{\psi,\pi} V^{\pi}(s)  - \calT^\pi V^{\pi} (s) = \gamma \cvar_\alpha^{\psi_{s,\pi(s)}} \left((p^\top - \calP_{s,\pi(s)}) V^\pi \right).
\end{align*} 
Fix an arbitrary state $\Bar{s} \in \calS$, since $p(\Bar{s}) = 1-\sum_{s'\neq \Bar{s}} p(s')$ and $\mathcal{P}_{s,\pi(s)}(\Bar{s}) = 1 - \sum_{s' \neq \Bar{s}} \calP_{s',\pi(s')}$, we have 
$$(I) = \gamma \operatorname{CVaR}_\alpha^{\psi_{s,\pi(s)}}\left((p^\top_{-\Bar{s}} - 
\mathcal{P}^\pi_{s,-\Bar{s}})(V^\pi_{-\Bar{s}} - V^\pi_{+\Bar{s}})\right).$$
Define $H(p_{-\Bar{s}}) = (p_{-\Bar{s}}^\top - \calP^\pi_{s,-\Bar{s}}) (V^\pi_{-\Bar{s}} - V^\pi_{+\Bar{s}})$, which is simply a linear function of variable $p_{\Bar{s}}$. Furthermore, we have
$$\mathbb{P}( s'|s,\pi(s))= \left\{
\begin{aligned}
    & \calP_{s,\pi(s)} (s') \quad & \text{ if } s' \neq \Bar{s}\\
    &1 - \sum_{s'' \neq \Bar{s}} \calP_{s,\pi(s)}(s'') & \text{ if } s' = \Bar{s}
\end{aligned}
\right.$$
Then, apply Theorem 4.3 in \cite{wu2018bayesian}, we obtain
$$\sqrt{{N}_{s,\pi(s)}}\cvar_\alpha^{\psi_{s,\pi(s)}} \left(H(p_{-\Bar{s}})\right) \Rightarrow \mathcal{N} (\lambda'_s, (\sigma'_s)^2),$$
where 
$\lambda'_s =-\frac{\sigma_s}{1-\alpha} \phi(\Phi^{-1}(\alpha)), \ (\sigma'_s)^2 = (V^\pi_{-\Bar{s}} - V^\pi_{+\Bar{s}})^\top \left(\operatorname{diag}\left(\calP^\pi_{s,-\Bar{s}}\right) - \calP^\pi_{s,-\Bar{s}} (\calP^\pi_{s,-\Bar{s}})^\top \right) (V^\pi_{-\Bar{s}} - V^\pi_{+\Bar{s}}). $
Furthermore, since $\frac{{N}_{s,\pi(s)}}{N} \rightarrow \bar{n}_s$. We obtain
\begin{align*}
    \sqrt{N}\cvar_\alpha^{\psi_{s,\pi(s)}} \left(H(p)\right) = \frac{\sqrt{N}}{\sqrt{N_{s,\pi(s)}}}\sqrt{N_{s,\pi(s)}}\cvar_\alpha^{\psi_{s,\pi(s)}} \left(H(p)\right) \Rightarrow \mathcal{N} (\lambda^\pi_s,(\sigma^\pi_s)^2).
\end{align*}
Next, we show the third term satisfies $\sqrt{N}(III) \rightarrow 0$ in probability. Indeed,
\begin{align*}
    \calT^{\psi,\pi} V^{\psi,\pi}(s) - \calT^\pi V^{\psi,\pi} 
    =& \gamma \cvar_\alpha^{p\sim \psi_{s,\pi(s)}} \left( p^\top V^{\psi,\pi} \right) - \gamma \calP_{s,\pi(s)} V^{\psi,\pi}\\ 
    =&\gamma \cvar_\alpha^{p\sim \psi_{s,\pi(s)}} \left[ (p^\top - \mathcal{P}_{s,\pi(s)}) V^{\psi,\pi} \right],
\end{align*}
where the second inequality holds since $\gamma \calP_{s,\pi(s)} V^{\psi,\pi}$ is a constant. Similarly, we also have 
\begin{align*} 
    \calT^{\psi,\pi} V^{\pi}(s) - \calT^\pi V^{\pi} 
    =\gamma \cvar_\alpha^{p\sim \psi_{s,\pi(s)}} \left[ (p^\top - \mathcal{P}_{s,\pi(s)}) V^{\pi} \right].
\end{align*}
Combining together, we have
\begin{align}
    \text{(III)}  
    = & \gamma \cvar_\alpha^{p\sim \psi_{s,\pi(s)}} \left[ (p^\top - \mathcal{P}_{s,\pi(s)}) V^{\psi,\pi} \right] - \gamma \cvar_\alpha^{p\sim \psi_{s,\pi(s)}} \left[ (p^\top - \mathcal{P}_{s,\pi(s)}) V^{\pi} \right] \label{eq: (III).1}\\
    \le & -\gamma \cvar_\alpha^{p\sim\psi_{s,\pi(s)}} \left( (p^\top - \mathcal{P}_{s,\pi(s)}) \left( V^\pi - V^{\psi,\pi}\right) \right)\notag \\
    \le & \gamma\left|\cvar_\alpha^{p\sim\psi_{s,\pi(s)}} \left( (p^\top - \mathcal{P}_{s,\pi(s)}) \left( V^\pi - V^{\psi,\pi}\right) \right)\notag\right |
    % \le &\gamma \mathbb{E}_{p\sim\psi_{s,\pi(s)}} \left[(p^\top - \mathcal{P}_{s,\pi(s)}) \left( V^\pi - V^{\psi,\pi}\right) \right] \notag\\
    % =&\gamma \left( \bar{\psi}_{s,\pi(s)} - \calP_{s,\pi(s)} \right) \left( V^\pi - V^{\psi,\pi}\right)\notag
    \end{align}
The inequality holds since for for CVaR with left-tail risk,
$ \operatorname{CVaR}(X) + \cvar(Y) \le \cvar(X+Y)$. 
%  Furthermore, CVaR is law-invariant, meaning we can pick the common random variable $p\sim\psi_{s,\pi(s)}$. The second inequality holds since CVaR is the left-tail conditional expectation.

Now we prove $\sqrt{{N}_{s,\pi(s)}}$(III) converge to zero in probability.
First notice 
\begin{align*}
    \|V^{\psi,\pi} - V^\pi\|_\infty 
    =& \|\calT^{\psi,\pi} V^{\psi,\pi} - \calT^{\psi,\pi} V^\pi + \calT^{\psi,\pi} V^\pi - \calT^\pi V^\pi \|_\infty\\
    \le &  \|\calT^{\psi,\pi} V^{\psi,\pi} - \calT^{\psi,\pi} V^\pi \|_\infty + \|\calT^{\psi,\pi} V^\pi - \calT^\pi V^\pi \|_\infty,
\end{align*}
This implies $ \|V^{\psi,\pi} - V^\pi\|_\infty \le \frac{1}{1-\gamma} \|\calT^{\psi,\pi} V^\pi - \calT^\pi V^\pi \|_\infty$.
For each $s \in \calS$, we have 
$$\calT^{\psi,\pi} V^\pi (s) - \calT^\pi V^\pi (s) = \gamma \cvar_\alpha^{\psi_{s,\pi(s)}} \left( (p^\top - \calP_{s,\pi(s)} ) V^\pi\right).$$

By Lemma \ref{lem: concentration expon}, we know for an arbitrary positive $\varepsilon$, consider $N$ large enough such that $\bar{N}^{\frac{3}{4}}_{s,\pi(s)} > \frac{|\calS|}{\varepsilon}$ for all $s\in \calS$. Then by Lemma $1$, we know
$$ \mathbb{P} \left(\|\bar{\psi}_{s,\pi(s)} - \calP_{s,\pi(s)}\|_\infty > \frac{\varepsilon}{{N}_{s,\pi(s)}^{\frac{1}{4}}}\right) \le 2|\calS|\mathrm{e}^{-2\left({N}_{s,\pi(s)}^{-\frac{1}{4}}\varepsilon-\frac{|\calS|}{{N}_{s,\pi(s)}}\right)^2 {N}_{s,\pi(s)}}$$
Consider Event $E = \{ \psi|\|\bar{\psi}_{s,\pi(s)} - \calP_{s,\pi(s)}\|_\infty \le \frac{\varepsilon}{{N}_{s,\pi(s)}^{\frac{1}{4}}}, \ \forall s\in \calS\}$. With a union bound we have 
$$\mathbb{P} (E) \ge 1 - 2|\calS|\sum_{s\in\calS}\mathrm{e}^{-2\left({N}_{s,\pi(s)}^{-\frac{1}{4}}\varepsilon-\frac{|\calS|}{{N}_{s,\pi(s)}}\right)^2 {N}_{s,\pi(s)}} $$
by Lemma \ref{lem: using chebyshev} where the parameter $\epsilon$ is set to ${N}_{s,\pi(s)}^{-\frac{1}{4}}\varepsilon $, $\delta$ is set to $2{N}_{s,\pi(s)}^{-\frac{1}{4}}\varepsilon $ and $\|V^\pi\|_\infty \le \frac{1}{1-\gamma}$, we obtain for all $s\in \calS$
$$ \left|\cvar_\alpha^{\psi_{s,a}}\left((p^\top-\calP_{s,\pi(s)})V^\pi \right)\right| \le \frac{|\calS|}{1-\gamma} \{ 2{N}_{s,\pi(s)}^{-\frac{1}{4}} \varepsilon + 2\frac{|\calS|}{1-\alpha}\frac{\varepsilon^2}{\bar{N}^\frac{1}{2}_{s,\pi(s)}}\}$$
This means $\|V^{\psi,\pi} - V^\pi\|_\infty \le \frac{2|\calS|}{(1-\gamma)^2} \max_{s \in \calS} \left\{ {N}_{s,\pi(s)}^{-\frac{1}{4}}\varepsilon + \frac{|\calS|}{1-\alpha} \frac{\varepsilon^2}{{N}_{s,\pi(s)}^{\frac{1}{2}}}   \right\} $.
We have conditioned on $E$,

\begin{align*}
    \sqrt{N}\text(III) \le& \sqrt{N} \gamma\left|\cvar_\alpha^{p\sim\psi_{s,\pi(s)}} \left( (p^\top - \mathcal{P}_{s,\pi(s)}) \left( V^\pi - V^{\psi,\pi}\right) \right)\notag\right | \\
    \le & \frac{4|\calS|^2\gamma}{(1-\gamma)^2} \max_{s \in \calS} \left\{\left(\frac{N}{{N}_{s,\pi(s)}}\right)^{\frac{1}{4}}\varepsilon + \frac{|\calS|}{1-\alpha} \frac{\varepsilon^2 N^{\frac{1}{4}} }{{N}_{s,\pi(s)}^{\frac{1}{2}}}  \right\}^2,
\end{align*}  
by Lemma \ref{lem: using chebyshev} with $V = V^{\psi,\pi} - V^\pi$, $\epsilon = {N}_{s,\pi(s)}^{-\frac{1}{4}}\varepsilon $ and $\delta = 2{N}_{s,\pi(s)}^{-\frac{1}{4}}\varepsilon $ 

% \begin{align*}
%     \sqrt{N}\text(III) \le& \sqrt{N} \gamma \left( \bar{\psi}_{s,\pi(s)} - \calP_{s,\pi(s)} \right) \left( V^\pi - V^{\psi,\pi}\right) \\
%     \le & \sqrt{N} \gamma \left\| \bar{\psi}_{s,\pi(s)} - \calP_{s,\pi(s)} \right\|_\infty \left\| V^\pi - V^{\psi,\pi}\right\|_1\\
%     \le &\sqrt{N} \gamma |\calS|\left\| \bar{\psi}_{s,\pi(s)} - \calP_{s,\pi(s)} \right\|_\infty \left\| V^\pi - V^{\psi,\pi}\right\|_\infty \\
%     \le & \gamma |\calS|N^{\frac{1}{4}}\left\| \bar{\psi}_{s,\pi(s)} - \calP_{s,\pi(s)} \right\|_\infty N^{\frac{1}{4}}\left\| V^\pi - V^{\psi,\pi}\right\|_\infty\\
%     \le & \gamma |\calS| \varepsilon \dot (1-\gamma)^{-1}\frac{2|\calS|}{1-\gamma} \max_{s \in \calS} \left\{\left(\frac{N}{{N}_{s,\pi(s)}}\right)^{\frac{1}{4}}\varepsilon + \frac{|\calS|}{\alpha} \frac{\varepsilon^2 N^{\frac{1}{4}} }{{N}_{s,\pi(s)}^{\frac{1}{2}}}   \right\} \\
% \end{align*}  
By assumption $\frac{N_{s,\pi(s)}}{N} \rightarrow \bar{n}_s$, For $N$ large enough, we can further bound the above inequality by 
\begin{align*}
    \sqrt{N}\text(III) \le& \frac{4|\calS|^2\gamma}{(1-\gamma)^2}  \left\{\underline{n}^{\frac{1}{4}}\varepsilon + \frac{|\calS|}{1-\alpha} \varepsilon^2 \underline{n} ^{\frac{1}{4}} \right\}^2, 
\end{align*}
with $\underline{n} = \max_s \frac{1}{\bar{n}_s}$.
By switching the positions of two terms in \eqref{eq: (III).1}, we can also obtain conditions on $E$, 
\begin{align*}
    \sqrt{N}\text(III) \ge& -\frac{4|\calS|^2\gamma}{(1-\gamma)^2}  \left\{\underline{n}^{\frac{1}{4}}\varepsilon + \frac{|\calS|}{1-\alpha} \varepsilon^2 \underline{n} ^{\frac{1}{4}} \right\}^2
\end{align*}
Hence,
\begin{align*}
    &\mathbb{P}\left( |\sqrt{N} \text{(III)}| \le \frac{4|\calS|^2\gamma}{(1-\gamma)^2}  \left\{\underline{n}^{\frac{1}{4}}\varepsilon + \frac{|\calS|}{1-\alpha} \varepsilon^2 \underline{n} ^{\frac{1}{4}} \right\}^2 \right) \\
    \ge  &1 - 2|\calS|\sum_{s\in\calS}\mathrm{e}^{-2\left({N}_{s,\pi(s)}^{-\frac{1}{4}}\varepsilon-\frac{|\calS|}{{N}_{s,\pi(s)}}\right)^2 {N}_{s,\pi(s)}} \rightarrow 1,
\end{align*} 
as $N\rightarrow \infty$. Notice the term $\frac{4|\calS|^2\gamma}{(1-\gamma)^2}  \left\{\underline{n}^{\frac{1}{4}}\varepsilon + \frac{|\calS|}{1-\alpha} \varepsilon^2 \underline{n} ^{\frac{1}{4}} \right\}^2 $ is increasing and only depends on $\varepsilon$. Hence, we have proved $|\sqrt{N} \text{(III)}|$ converge to $0$ in probability.

Hence, we have 
$$ (\mathrm{I} - \gamma \mathcal{P}_\pi)(V^{\psi,\pi}- V^\pi) \Rightarrow \mathcal{N}(\gamma \lambda^\pi,\gamma^2 \operatorname{diag}((\sigma^\pi)^2) )$$
where $(\sigma^\pi)^2$ is in terms of \eqref{eq: variance Bars free}.

Finally, we show that $\sigma^\pi$ does not depend on the choice of $\Bar{s}$, which only requires extra effort of algebraic calculation. For simplicity, we drop the superscript $\pi$ in the following proof. We also write $\mathcal{P}(s') = \mathcal{P}_{s,\pi(s)}$ for simplicity since we fix both $s$ and $\pi$. To begin with,
\begin{align*}
    &(V_{-\Bar{s}} - V_{+\Bar{s}})^\top \left(\operatorname{diag}\left(\calP_{s,-\Bar{s}}\right) - \calP_{s,-\Bar{s}} (\calP_{s,-\Bar{s}})^\top \right) (V_{-\Bar{s}} - V_{+\Bar{s}}) \\
    = & \underbrace{(V_{-\Bar{s}} )^\top \left(\operatorname{diag}\left(\calP_{s,-\Bar{s}}\right) - \calP_{s,-\Bar{s}} (\calP_{s,-\Bar{s}})^\top \right) V_{-\Bar{s}}}_{A} - \underbrace{2 (V_{-\Bar{s}})^\top \left(\operatorname{diag}\left(\calP_{s,-\Bar{s}}\right) - \calP_{s,-\Bar{s}} (\calP_{s,-\Bar{s}})^\top \right) V_{+\Bar{s}}}_{B}\\
    &+ \underbrace{( V_{+\Bar{s}})^\top \left(\operatorname{diag}\left(\calP_{s,-\Bar{s}}\right) - \calP_{s,-\Bar{s}} (\calP_{s,-\Bar{s}})^\top \right)  V_{+\Bar{s}}}_{C}.
\end{align*}
We then have
$$ A = \underbrace{\sum_{s'\neq \Bar{s}} \calP(s') V(s')^2 }_{A.1}- \underbrace{\sum_{s_1,s_2 \neq \Bar{s}} \calP(s_1)\calP(s_2) V(s_1)V(s_2)}_{A.2}.$$
\begin{align*}
    C =& \sum_{s'\neq \Bar{s}} \mathcal{P}(s') V(\Bar{s})^2 - \sum_{s_1,s_2 \neq \Bar{s}} \calP(s_1) \calP(s_2) V(\Bar{s})^2 \\ 
    =& V(\Bar{s})^2 (1 - \calP(\Bar{s})) - V(\Bar{s})^2 (1-\calP(\Bar{s}))^2\\
    =& \underbrace{V(\Bar{s})^2 \calP(\Bar{s})}_{C.1} - \underbrace{V(\Bar{s})^2 \calP(\Bar{s})^2}_{C.2}
\end{align*}
\begin{align*}
    B = & 2\sum_{s'\neq \Bar{s}} \calP(s') V(s') V(\Bar{s}) - 2\sum_{s_1,s_2\neq \Bar{s}} \calP(s_1) \calP(s_2) V(s_1) V(\Bar{s}) \\ 
    = & 2\sum_{s'\neq \Bar{s}} \calP(s') V(s') V(\Bar{s}) - 2\sum_{s_1\neq \Bar{s}} \calP(s_1) V(s_1) V(\Bar{s}) (1-\calP(\Bar{s})) \\
    =& 2 \sum_{s'\neq \Bar{s}} \calP(s')\calP(\Bar{s}) V(s') V(\Bar{s})
\end{align*}
We have 
$$A.1 + C.1 = \sum_{s'\in \calS} \calP(s') V(s')^2 = V^\top \operatorname{diag}(\calP) V,$$
$$ -A.2 - C.2 - B = \sum_{s_1,s_2\in\calS} \calP(s_1)\calP(s_2)V(s_1)V(s_2) = -V^\top \calP \calP^\top V.$$
Hence, we have 
$$(V_{-\Bar{s}} - V_{+\Bar{s}})^\top \left(\operatorname{diag}\left(\calP_{s,-\Bar{s}}\right) - \calP_{s,-\Bar{s}} (\calP_{s,-\Bar{s}})^\top \right) (V_{-\Bar{s}} - V_{+\Bar{s}}) = 
V^\top \left( \operatorname{diag} (\calP) - \calP \calP^\top \right) V.
$$
This gives us the expression \eqref{eq: variance Bars free} and completes the proof.
\hfill $\blacksquare$

\subsection{Proof of Lemma \ref{lem: order statistic}}
\proof{}
    For 1, since 
    \begin{equation*}
        \mathbb{E} \left[ X_{1:n}\right] = \int_0^\infty nxf(x)(1-F(x))^{n-1} 
        \mathrm{d}x
    \end{equation*}
    and 
    \begin{equation*}
        \text{CVaR}_{\frac{1}{n}} (X) = n\int_0^{F^{-1}\left(\frac{1}{n}\right)} xf(x)\mathrm{d}x.
    \end{equation*}
    Then we have 
    \begin{align*}
        &\frac{1}{n}\left(\mathbb{E} \left[ X_{1:n}\right] - \text{CVaR}_{\frac{1}{n}} (X) \right)\\
        =&\int_0^{F^{-1}\left(\frac{1}{n}\right)} x f(x) \left[ (1-F(x))^{n-1} -1 \right] \mathrm{d}x+\int_{F^{-1}\left(\frac{1}{n}\right)}^\infty x f(x)(1-F(x))^{n-1} \mathrm{d}x\\
        \ge & F^{-1}\left(\frac{1}{n}\right) \left\{\int_0^{F^{-1}\left(\frac{1}{n}\right)} f(x) \left[ (1-F(x))^{n-1} -1 \right] \mathrm{d}x +\int_{F^{-1}\left(\frac{1}{n}\right)}^\infty f(x)(1-F(x))^{n-1} \mathrm{d}x\right\}\\
        =& F^{-1}\left(\frac{1}{n}\right) \left\{\int_0^{F^{-1}\left(\frac{1}{n}\right)} \left[ (1-F(x))^{n-1} -1 \right] \mathrm{d}F(x)+\int_{F^{-1}\left(\frac{1}{n}\right)}^\infty (1-F(x))^{n-1} \mathrm{d}F(x)\right\}\\
        =& F^{-1}\left(\frac{1}{n}\right) \left[ \int_0^\infty (1-F(x))^{n-1}\mathrm{d}F(x) -\frac{1}{n} \right]\\
        =&F^{-1}\left( \left. \frac{1}{n}\right) \left[ -\frac{(1-F(x))^{n}}{n} \right|_{x=0}^{x=\infty}  - \frac{1}{n}\right] \\
        =& 0 .
    \end{align*}
    For 2, if $n = 2$, then $X_{2:2} = \max_{i=1,2} X_i \ge X_1$. Hence $\mathbb{E} [X_{2:n}] \ge \mathbb{E} [X_1] \ge \text{CVaR}_\alpha(X)$ for all $0\le 1-\alpha \le 1$. Otherwise for $n\ge 3$,
    $$ \mathbb{E} [X_{2:n}] = n(n-1)\int_0^\infty xf(x)F(x)(1-F(x))^{n-2}\mathrm{d}x$$
    and 
    $$ \text{CVaR}_{\frac{2}{n}}(X) = \frac{n}{2} \int_0^{F^{-1}\left(\frac{2}{n}\right)} xf(x)\mathrm{d} x.$$
    Then, 
    \begin{align}
        \frac{1}{n}\left(\mathbb{E} \left[ X_{2:n}\right] - \text{CVaR}_{\frac{2}{n}} (X) \right) \notag
        =&\int_0^{F^{-1}\left(\frac{2}{n}\right)} x f(x) \left[ (n-1)F(x)(1-F(x))^{n-2} -\frac{1}{2} \right] \mathrm{d}x\notag\\
        &+\int_{F^{-1}\left(\frac{2}{n}\right)}^\infty x f(x)(n-1)F(x)(1-F(x))^{n-2} \mathrm{d}x \label{eq:1}
    \end{align}
    Notice $F(x)(1-F(x))^{n-2}$ attains its maximum at $F(x) = \frac{1}{n-1} \le \frac{2}{n}$ for $n\ge 2$. Hence,
    $$ (n-1)F(x)(1-F(x))^{n-2} \le (1-\frac{1}{n-1})^{n-2}. $$
    Let $g(x) = (x-2)\log(1-\frac{1}{x-1})$. We have 
    $$ g'(x) = \log(1-\frac{1}{x-1}) + \frac{1}{x-1} = \frac{1}{x-1} - \log(1+\frac{1}{x-2})$$
    and
    $$g''(x) = -\frac{1}{(x-1)^2} + \frac{1}{(x-1)(x-2)} < 0 \ \text{for} x \ge 3.$$
    Furthermore since $ \lim_{x\rightarrow \infty} g'(x) = 0$, we know $g'(x) \le 0$ for $x \ge 3$ and hence, $g(x) \le g(3) = \log\frac{1}{2}$. Hence
    $$ (n-1)F(x)(1-F(x))^{n-2} -\frac{1}{2} \le (1-\frac{1}{n-1})^{n-2} -\frac{1}{2} \le 0$$
    always holds for $n\ge 3$. Then, we  further have 
    \begin{align*}
        \eqref{eq:1} \ge &F^{-1}\left(\frac{2}{n}\right) \int_0^{F^{-1}\left(\frac{2}{n}\right)} \left[ (n-1)F(x)(1-F(x))^{n-2} -\frac{1}{2} \right]\mathrm{d}F(x)\\
        &+  F^{-1}\left(\frac{2}{n}\right) \int^\infty_{F^{-1} \left(\frac{2}{n}\right)} (n-1)F(x)(1-F(x))^{n-2} \mathrm{d}F(x) \\
        = &\int_0^\infty (n-1)F(x)(1-F(x))^{n-2} \mathrm{d}F(x) - \frac{1}{n}\\
        =& \left. -F(x)(1-F(x))^{n-1} \right|_0^\infty - \left.\frac{1}{n} (1-F(x))^{n} \right|_0^\infty - \frac{1}{n} \\
        =& 0 
    \end{align*}
\hfill $\blacksquare$

% \textbf{Proof of Theorem \ref{thm: concentrate t}}
% \proof{}
%     \begin{align*}
%         \mathbb{P} (X>x) = & \frac{\Gamma(\frac{n+1}{2}}{\sqrt{n\pi}\Gamma(\frac{n}{2})} \int_{x}^\infty \frac{1}{\left( 1+\frac{t^2}{n}\right)^{\frac{n+1}{2}}} \mathrm{d}t \\
%         \le & \frac{\Gamma(\frac{n+1}{2}}{\sqrt{n\pi}\Gamma(\frac{n}{2})} \int_{x}^\infty  \frac{n}{2} \frac{t}{x} \frac{1}{\left( 1+\frac{t^2}{n}\right)^{\frac{n+1}{2}}} \mathrm{d}(1+\frac{t^2}{n})\\
%         \le&- \sqrt{\frac{n}{\pi}} \frac{\Gamma(\frac{n+1}{2})}{\Gamma(\frac{n}{2})} \frac{1}{n-1} \left. \frac{1}{\left( 1+\frac{t^2}{n}\right)^{\frac{n-1}{2}}} \right|_{x}^\infty\\
%         = & \frac{\sqrt{n}}{\sqrt{\pi}(n-1)} \frac{\Gamma(\frac{n+1}{2})}{\Gamma(\frac{n}{2})} \left( 1+\frac{x^2}{n}\right)^{-\frac{n-1}{2}}.
%     \end{align*}
%     Notice 
%     $ (1+\frac{x^2}{n})^{-(n+1)} < \mathrm{e}^{-x^2}$ holds for all $n >0$, which leads to $(1+\frac{x^2}{n})^{-\frac{(n+1)}{2}} < \mathrm{e}^{-\frac{x^2}{2}} $. Hence, we obtain 
%     $$ \mathbb{P} (X>x) \le \frac{\sqrt{n}}{\sqrt{\pi}(n-1)} \frac{\Gamma(\frac{n+1}{2})}{\Gamma(\frac{n}{2})} \mathrm{e}^{-\frac{x^2}{2}}  \left( 1+\frac{x^2}{n}\right).$$
%     Furthermore, if $n$ is even, then 
%     $\frac{\Gamma(\frac{n+1}{2})}{\Gamma(\frac{n}{2})} \le \frac{\Gamma(\frac{n}{2}+1)}{\Gamma(\frac{n}{2})} = \frac{n}{2}. $ This further leads to $$ \mathbb{P} (X>x) \le \frac{n\sqrt{n}}{2\sqrt{\pi}(n-1)}  \mathrm{e}^{-\frac{x^2}{2}}  \left( 1+\frac{x^2}{n}\right).$$
% \hfill $\blacksquare$

\subsection{Proof of Theorem \ref{thm: BR-regret MAB}}
% Let $\widehat{ r}^t_{s_t,a} = \hat{ r}_{s_t,a}^{{\psi}_{a}^t} $ denote the Monte Carlo estimator of $ r_{s_t,a}^t  =  r_{s_t,a}^{\psi^t_a}$ as in the algorithm with $ n=\lceil\frac{1}{1-\alpha}\rceil$ samples, and $a_t^*$ be the optimal arm in $t^{th}$ pull with context $s_t$ and posterior $\psi^t$. 
The following Lemma \ref{lem: CVaR truncated normal} and Lemma \ref{lem: hat CVaR truncated normal} are needed to prove Theorem \ref{thm: BR-regret MAB}.
\begin{lem} \label{lem: CVaR truncated normal}
    For $X\sim \mathcal{N}^+ ( r,\sigma^2)$, with $1 \ge  r \ge 0$ and $\alpha \ge \frac{1}{2}$, we have
   \begin{equation*}
       \text{CVaR}_\alpha (X) \ge  r - (\frac{1}{\sqrt{2\pi}} + \phi(\Phi^{-1}(\alpha)))\frac{\sigma}{1-\alpha}.
   \end{equation*}
\end{lem}
\proof{}
    For any $\beta > 0$, Let $Y \sim \mathcal{N}( r,\sigma^2)$ and $X = \max\{0,\min\{Y,1\}\}$. 
    
    VaR$_\beta(Y) = \Phi^{-1}(\beta)\sigma +  r$. Then 
    $ \text{VaR}_\beta(X) = \left\{
    \begin{aligned}
        &0 ,&\text{ if }\beta < \Phi(-\frac{ r}{\sigma})\\
        & \Phi^{-1}(\beta)\sigma +  r, & \text{ if } \Phi(-\frac{ r}{\sigma})\le \beta \le \Phi(\frac{1- r}{\sigma})\\
        &1, & \text{ if } \beta > \Phi(\frac{1- r}{\sigma})
    \end{aligned}
    \right.$
    Since $\Phi(\frac{1- r}{\sigma}) > \Phi(0) = \frac{1}{2} \ge 1-\alpha$, we have
    \begin{align*}
        &1-(1-\alpha) \text{CVaR}_\alpha (X) = \int_{0}^{1-\alpha} \text{VaR}_\beta (X) \mathrm{d}\beta\\
        =&\int_{0}^{\Phi(-\frac{ r}{\sigma})} 0 \mathrm{d}\beta + \int_{\Phi(-\frac{ r}{\sigma})}^{1-\alpha} (\Phi^{-1}(\beta)\sigma +  r )\mathrm{d}\beta  = (1-\alpha) r  - \Phi(-\frac{ r}{\sigma}) r + \sigma\int_{-\frac{ r}{\sigma}}^{\Phi^{-1}(\alpha)} y\phi(y)\mathrm{d}y \\ 
        \ge&  (1-\alpha) r  - \Phi(-\frac{ r}{\sigma}) r + \sigma\int_{-\infty}^{\Phi^{-1}(\alpha)} y\phi(y)\mathrm{d}y  = (1-\alpha) r  - \Phi(-\frac{ r}{\sigma}) r - \frac{\sigma}{\sqrt{2\pi}}\int_{-\infty}^{\Phi^{-1}(\alpha)}\mathrm{d}\mathrm{e}^{-\frac{y^2}{2}} \\
        =& (1-\alpha) r  - \Phi(-\frac{ r}{\sigma}) r - \left.\frac{\sigma}{\sqrt{2\pi}}\mathrm{e}^{-\frac{y^2}{2}} \right|_{-\infty}^{\Phi^{-1}(\alpha)} = (1-\alpha) r  - \Phi(-\frac{ r}{\sigma}) r - \phi( \Phi^{-1}(\alpha))
    \end{align*}
Let $Z$ denote a standard normal random variable, then we have for an arbitrary $z > 0$,
$$ \mathbb{P}(Z \le -z) = \frac{1}{2} \mathbb{P}(|Z| \ge z) \le \frac{\mathbb{E}[|Z|]}{2z} = \sqrt{\frac{2}{\pi}}\frac{1}{2z} = \frac{1}{\sqrt{2\pi}z}.$$
Hence, for the second term, 
$    \Phi(-\frac{ r}{\sigma}) r = \mathbb{P} (Z\le - \frac{ r}{\sigma})  r \le \frac{\sigma}{\sqrt{2\pi}}
$.
We have 
$ \text{CVaR}_\alpha(X) \ge  r - \frac{\sigma}{1-\alpha}(\frac{1}{\sqrt{2\pi}}+\phi( \Phi^{-1}(\alpha))) $
\hfill $\blacksquare$
\begin{lem} \label{lem: hat CVaR truncated normal}
    For $X\sim \mathcal{N}^+ ( r,\sigma^2)$, with $1 \ge  r \ge 0$ and $\alpha \ge \frac{1}{2}$, $\{X_i\}_{i=1}^n$ being i.i.d samples of $X$ and $n=\lceil\frac{1}{1-\alpha}\rceil$ we have with probability $1-\delta$,
   \begin{equation*}
       \widehat{\text{C}}_\alpha (\{X_i\}_{i=1}^n) \le  r + \sqrt{2(1-\alpha)\sigma^2\ln \frac{1}{\delta}}.
   \end{equation*}
\end{lem}
\proof{}
    Let $Y \sim \mathcal{N}( r,\sigma^2)$ and $X = \max\{0,\min\{Y,1\}\}$. We know $|X- r| \le |Y- r|$ always holds.
    Then,
    \begin{align*}
    &\mathbb{P} (\widehat{\text{C}}_\alpha (\{X_i\}_{i=1}^n)\le  r + \epsilon) 
    \ge \mathbb{P} (X_{1:n} \le  r+\epsilon) \\
    =&  \mathbb{P} (\cup\{X_i \le  r+\epsilon\} )
    =  1 - \mathbb{P} (\cap \{X_i >  r+\epsilon \}) \\
    =& 1 - \Pi_{i=1}^n\mathbb{P}(X_i >  r+ \epsilon) 
    \ge  1 - \Pi_{i=1}^n\mathbb{P}(Y_i >  r+ \epsilon) 
    \ge  1 - \mathrm{e}^{-\frac{n\epsilon^2}{2\sigma^2}}
    \end{align*}
    %   \begin{align*}
    % \mathbb{P} (\widehat{\text{C}}_\alpha (\{X_i\}_{i=1}^n)\le  r + \epsilon) 
    % \ge& \mathbb{P} (X_{1:n} \le  r+\epsilon) \\
    % = & \mathbb{P} (\cup\{X_i \le  r+\epsilon\} )\\
    % = & 1 - \mathbb{P} (\cap \{X_i >  r+\epsilon \}) \\
    % = &1 - \Pi_{i=1}^n\mathbb{P}(X_i >  r+ \epsilon) \\
    % \ge & 1 - \Pi_{i=1}^n\mathbb{P}(Y_i >  r+ \epsilon) \\
    % \ge & 1 - \mathrm{e}^{-\frac{n\epsilon^2}{2\sigma^2}}
    % \end{align*}
    Set $\delta = \mathrm{e}^{-\frac{n\epsilon^2}{2\sigma^2}}$. Then $\epsilon = \sqrt{\frac{2\sigma^2}{n}\ln \frac{1}{\delta}} \le\sqrt{2(1-\alpha) \sigma^2\ln \frac{1}{\delta}} $. This completes the proof
\hfill $\blacksquare$
\textbf{Proof of Theorem \ref{thm: BR-regret MAB}:}
By Lemma \ref{lem: order statistic}, conditioned on $\psi^t$, we have
\(
    \mathbb{E}[\widehat{r}^t(s_t,a) | \psi^t_a] \ge  r^t_{s_t,a}
\).
Let $$\text{BR-regret(t)} = \mathbb{E} [  r^t(s_t, a^*_t) -  r^t(s_t,a_t)] \le  \mathbb{E} [\hat r^{t}(s_t,a^*_t) -  r^t(s_t,a_t)] \le \mathbb{E} [\hat{ r}^{t}(s_t,a_t) -  r^t(s_t,a_t)] $$

The left is to bound $ \widehat{ r}^t(s_t,a_t) -  r^t(s_t,a_t)$, for which purpose we derive the following results.
With Lemma \ref{lem: CVaR truncated normal}, we have 
    $$ r^{t}(s_t,a) \ge s_t^\top \theta_a^t - \frac{\nu\sqrt{s_t^\top (V_a^t)^{-1} s_t}}{1-\alpha}(\frac{1}{\sqrt{2\pi}}+\phi( \Phi^{-1}(\alpha))).$$
    For $\widehat{ r}^t(s_t,a)$, from Lemma \ref{lem: hat CVaR truncated normal}, let $ r = s_t^\top \theta_a^t, \sigma^2 = \nu^2 s_t^\top (V_a^t)^{-1}s_t$ and $\delta = \frac{1}{T^2 K}$, then we know
    \begin{align*} 
    &\mathbb{P}\left(\widehat{ r}^t(s_t,a) \le s_t^\top\theta_a^t + \nu\sqrt{2\alpha s_t^\top (V_a^t)^{-1} s_t\ln (KT^2) }\right)\\
    = &\mathbb{E}\left[\mathbb{P}(\widehat{ r}^t(s_t,a) \le s_t^\top\theta_a^t + \nu\sqrt{2\alpha s_t^\top (V_a^t)^{-1} s_t\ln (KT^2) }| \mathcal{H}_t )\right] 
    \le  \frac{1}{T^2K},
    \end{align*}
    where $\mathcal{H}_t = \{r(s_\tau,a_\tau), \tau = 1,\ldots,t-1\} $ denotes the past observations. 
    With a union bound we have with probability $1-\frac{1}{T}$,
    $$ \widehat{ r}^t(s_t,a) \le s_t^\top\theta_a^t + \nu\sqrt{2(1-\alpha) s_t^\top (V_a^t)^{-1} s_t\ln (KT^2) }, \forall a\in[K],t\in[T] $$
    Then, With probability $1-\frac{1}{T}$, the following event holds,
    $$ D = \left\{\widehat{ r}^t(s_t,a) -  r^t(s_t,a) \le \left(\nu\sqrt{2(1-\alpha)\ln (KT^2) } + \frac{\nu}{1-\alpha}(\frac{1}{\sqrt{2\pi}}+\phi( \Phi^{-1}(\alpha)))\right) \sqrt{s_t^\top (V_a^t)^{-1} s_t}, \forall a \in[K], t\in [T]\right\}$$
    Then we have with probability $1-\frac{1}{T}$
    $$ \sum_{t=1}^T \left(\widehat{ r}^t(s_t,a_t) -  r^t(s_t,a_t)\right) \le \left(\nu\sqrt{2(1-\alpha)\ln (KT^2) } + \frac{\nu}{1-\alpha}(\frac{1}{\sqrt{2\pi}}+\phi( \Phi^{-1}(\alpha)))\right) \sum_{t=1}^T\sqrt{s_t^\top (V_{a_t}^t)^{-1} s_t}$$
    According to the proof of Theorem 2 in \cite{lin2022risk}, we have
    $ \sum_{t=1}^T \sqrt{s_t^\top (V_{a_t}^t)^{-1} s_t} \le 5\sqrt{dKT\ln T}.$
    This implies with probability $1-\frac{1}{T}$,
    $$ \sum_{t=1}^T \hat{ r}^t(s_t,a_t)  -  r^t(s_t,a_t) \le \left(\nu\sqrt{2(1-\alpha)\ln (KT^2) } + \frac{\nu}{1-\alpha}(\frac{1}{\sqrt{2\pi}}+\phi( \Phi^{-1}(\alpha)))\right) 5\sqrt{dKT\ln T}. $$
    Furthermore, notice $\hat{ r}^t(s_t,a_t) -  r^t(s_t,a_t) \le 1-0$ by the definition. We have
    $$ \text{BR-Regret}(T) \le \left(\nu\sqrt{2(1-\alpha)\ln (KT^2) } + \frac{\nu}{1-\alpha}(\frac{1}{\sqrt{2\pi}}+\phi( \Phi^{-1}(\alpha)))\right) 5\sqrt{dKT\ln T} + 1.$$
\hfill $\blacksquare$
\subsection{Proof of Theorem \ref{thm: bandit conventional regret bound}}
The following lemmas are needed to prove Theorem \ref{thm: bandit conventional regret bound}.
\begin{lem}(Lemma 1 in \cite{agrawal2014thompson}) \label{lem: theta}
    Define the concentration set $C^\theta(t) := \{|\theta_a^t - \theta_a^c| \le \ell(t) \sigma_a(t), \forall a \in [K] \}$. The $C^\theta(t)$ holds true with probability $1-\frac{\delta}{4t^2}$, where $\ell(t) = R \sqrt{d \ln \left(\frac{4t^3}{\delta}\right)}+1$.
\end{lem}

\begin{lem}
Define the concentration set $C^ r(t) := \{ |\widehat{ r}^t(s_t,a) - s_t^\top \theta_a^t| \le h(t) \sigma_a(t), \forall a \in [K] \}$. Then $C^ r(t)|\mathcal{F}_{t-1}$ holds true with probability at least $1-\frac{1}{t^2}$, where $\mathcal{F}_{t-1}$ is the filteration generated by $\{x_\tau,a_\tau,r(s_\tau,a_\tau)m, \tau \in [t-1], s_t  \}$ and $h(t) =  \sqrt{2\ln\frac{2 t^2}{1-\alpha}}\nu_t $.    
\end{lem}
\proof{}
\begin{align*}
    &\mathbb{P} (|\widehat{ r}^t(s_t,a) - s_t^\top \theta_a^t| \le \varepsilon |\mathcal{F}_t)
    \ge  \mathbb{P} (\cap_{i\in[n]} | r_{a,i} - s_t^\top \theta_a^t| \le \varepsilon | \mathcal{F}_t) \\
    = & \mathbb{P}^n (| r_{a,1} - s_t^\top \theta_a^t| \le \varepsilon | \mathcal{F}_t) 
    = (1 -  \mathbb{P} (| r_{a,1} - s_t^\top \theta_a^t| \ge \varepsilon | \mathcal{F}_t))^n\\
    \ge& 1 - n\mathbb{P} (| r_{a,1} - s_t^\top \theta_a^t| \ge \varepsilon | \mathcal{F}_t)
    \ge 1 - 2n \exp{(-\frac{\varepsilon^2}{2\nu_t^2\sigma_{s_t,a}^2} )}
\end{align*}
Set $2n\exp{(-\frac{\varepsilon^2}{2\nu_t^2\sigma_{s_t,a}^2} )} = \frac{1}{t^2}$, we have $\varepsilon = \sqrt{2\ln{2nt^2}}\nu_t\sigma_{s_t,a} \le \sqrt{2\ln\frac{ 2t^2(2-\alpha)}{1-\alpha}}\nu_t \sigma_{s_t,a}$. 
\hfill $\blacksquare$
\begin{definition} (Saturated set)
    Let $S(t) = \{a\in[K], s_t^\top \theta_{a_t} - s_t^\top \theta_a^c > g_t \sigma_{s_t,a}  \}$ be the Saturated set, where $g_t = h(t) + \ell(t)$.
\end{definition}
\begin{lem}
Conditioned on $\mathcal{F}_{t-1}$ and that $C^\theta$ is true,
    $$
\mathbb{P}\left(\widehat{ r}^t(s_t,a^*_t)>s_t^T \theta_{a^*t}^t \mid \mathcal{F}_{t-1}\right) \geq p, \text{ where }p = \frac{1}{(4\sqrt{\pi}\mathrm{e})^{\frac{2-\alpha}{1-\alpha}}}.
$$
\end{lem}
\proof{}
Denote by $  r^t_{a,i}$ the $i^{th}$ samples for arm $a$ at iteration $t$ by Algorithm $\ref{alg: BRCMAB-Gaussian}''$. We have
    \begin{align*}
        \mathbb{P}\left(\widehat{ r}^t(s_t,a^*_t)>s_t^T \theta_{a^*t}^t \mid \mathcal{F}_{t-1}\right) 
        \ge &\mathbb{P}\left(\cap_{i\in[n]} \{ r^t_{a^*_t,i}>s_t^T \theta_{a^*t}^t\} \mid \mathcal{F}_{t-1}\right) \\
        \ge &\mathbb{P}\left(\cap_{i\in[n]} \{\frac{ r^t_{a^*_t,i}-s_t^\top \theta_{a^*_t}^t}{\nu_t\sigma_{s_t,a^*_t}}>\frac{s_t^T \theta_{a^*t}^t-s_t^\top \theta_{a^*_t}^t}{\nu_t\sigma_{s_t,a^*_t}}\} \mid \mathcal{F}_{t-1}\right)\\
        = &\mathbb{P}^n\left( \frac{ r^t_{a^*_t,1}-s_t^\top \theta_{a^*_t}^t}{\nu_t\sigma_{s_t,a^*_t}}>\frac{s_t^T \theta_{a^*t}^t-s_t^\top \theta_{a^*_t}^t}{\nu_t\sigma_{s_t,a^*_t}}\mid \mathcal{F}_{t-1}\right)\\
        \ge & \frac{1}{4^n\sqrt{\pi^n}Z^n_t}\mathrm{e}^{-nZ_t^2},
    \end{align*}
    where
    $ Z_t = |\frac{s_t^T \theta_{a^*t}^t-s_t^\top \theta_{a^*_t}^t}{\nu_t\sigma_{s_t,a^*_t}}| \le \frac{\ell(t)}{\nu_t}\le 1.$
    And the last inequality uses the anti=concentration bound for Gaussian random variables. That is, let $Z \sim \mathcal{N}(0,1)$. Then for $z > 0$, 
    $ \mathbb{P} ( |Z| > z ) \ge \frac{1}{2 \sqrt{\pi} z} e^{-z^2 / 2}.
$
Hence, we then have 
$$ \mathbb{P}\left(\widehat{ r}^t(s_t,a^*_t)>s_t^T \theta_{a^*t}^t \mid \mathcal{F}_{t-1}\right) \ge \frac{1}{(4\sqrt{\pi}e)^{n}}\ge \frac{1}{(4\sqrt{\pi}e)^{\frac{2-\alpha}{1-\alpha}}}$$
\hfill $\blacksquare$

The rest of the proof follows from \cite{agrawal2014thompson}(arxiv), with different $g_t$, $h(t)$, $p$ and $\nu_t$.

\begin{lem} (Lemma 3 in \cite{agrawal2014thompson})
Conditioned on $\mathcal{F}_{t-1}$ and that $C^\theta$ is true.  
$$
\mathbb{P}\left(a(t) \notin S(t) \mid \mathcal{F}_{t-1}\right) \geq p-\frac{1}{t^2},
$$  
\end{lem}
\begin{lem}
    Let regret(t) $:= s_t^\top \theta^c_{a^*_t} - s_t^\top \theta^c_{a_t}$ and regret'(t) $:=$regret(t)$\mathbf{1}_{C^\theta(t)}$. Furthermore, let 
    $$s_t = \text{regret}^{\prime}(t)-\frac{3 g_t}{p} s_{a(t)}(t)-\frac{2 g_t}{p t^2},\quad Y_t = \sum_{\tau=1}^t s_t.$$ Then, $Y_t$ is a super-martingale with filtration $\mathcal{F}_t$.
    
\end{lem}
 Then, notice $|s_t|$ is bounded by $1+\frac{3}{p}g_t + \frac{2}{pt^2}g_t \le \frac{6}{p}g_t$. Applying the Azumma Hoeffeding theorem, we obatin with probaility $1-\frac{\delta}{2}$, 
 \begin{equation*}
     \sum_{t=1}^T \operatorname{regret}^{\prime}(t) \leq \sum_{t=1}^T \frac{3 g_t}{p} \sigma^t_{s_t,a_t}+\sum_{t=1}^T \frac{2 g_t}{p t^2}+\sqrt{2\left(\sum_t \frac{36 g_t^2}{p^2}\right) \ln \left(\frac{2}{\delta}\right)} 
 \end{equation*}

Notice $g_t \le g_T = h(T) + \ell(T) \le \frac{3}{2} \sqrt{d\ln{\frac{4T}{\delta}}\ln{\frac{2T^2}{1-\alpha}}} $. Furthermore, with 
$\sum_{t=1}^T \sigma_{s_t,a_t}^t \le 5\sqrt{dKT\ln T}$
and $\sum_{t=1}^T \frac{1}{t^2} \le 2$. We have with probability $1-\frac{\delta}{2}$
$$ \sum_{t=1}^T \operatorname{regret}^{\prime}(t)  \le \frac{3(4\sqrt{\pi}e)^{\frac{2-\alpha}{1-\alpha}}}{2} \sqrt{d\ln{\frac{4T}{\delta}}\ln{\frac{2T^2}{1-\alpha}} } \left( 15\sqrt{dKT\ln{T}} + 6\sqrt{2\ln{\frac{2}{\delta}T}}\right).$$
Furthermore, with Lemma \ref{lem: theta} and a union bound, we have with probability $1-\sum_{t=1}^T\frac{\delta}{4t^2} \ge 1-\frac{\delta}{2}$, $\cap_{t\in[T]}C^\theta(t)$ hold true. Then we know With probability $1-\delta$,

$$ \sum_{t=1}^T \operatorname{regret}(t)  \le \frac{3(4\sqrt{\pi}e)^{\frac{2-\alpha}{1-\alpha}}}{2} \sqrt{d\ln{\frac{4T}{\delta}}\ln{\frac{2T^2}{1-\alpha}} } \left( 15\sqrt{dKT\ln{T}} + 6\sqrt{2\ln{\frac{2}{\delta}T}}\right).$$

To obtain the expected regret, set $\delta = \frac{1}{T^2}$ and notice $r_{x,a} \in [0,1]$,  we have 
$$ \mathbb{E} [\sum_{t=1}^T \operatorname{regret}(t) ] \le  \frac{3(4\sqrt{\pi}e)^{\frac{2-\alpha}{1-\alpha}}}{2}\sqrt{d\ln{{4T^3}}\ln{\frac{2T^2}{1-\alpha}} } \left( 15\sqrt{dKT\ln{T}} + 6\sqrt{2T\ln{2T^2}}\right) $$

% \iffalse

\section{Proof of Theorem \ref{thm: UCBVI}}

\begin{lem} \label{lem: hat rho concentration}
    With Probability $1-\delta/2$, $\forall s\in\calS,a\in\calA,t\in[T]$ and $V$ such that $\|V\|_\infty \le 1$, 
    \begin{equation*}
    \|\hat{\rho}^t_{s,a} V - P_{s,a}^c V\|_\infty \le  3\sqrt{\frac{|\calS|^2}{2(N_{s,a}^t + 1)}\log{\frac{4(n+1)|\calS|^2|\calA|T}{\delta}}}
    \end{equation*}
\end{lem}
\proof{}
Notice When $N_{s,a}^t = 0$, the Lemma holds trivially. Hence we only consider $N_{s,a}^t \ge 1$. For an arbitrary $\delta>0$, let $\varepsilon = \sqrt{\frac{1}{2(N_{s,a}^t+1)}\ln{\frac{4(n+1)|\calS|^2|\calA|T}{\delta}}}$.
First notice 
$$ \mathbb{P} \left( \|\hat{\rho}^t_{s,a} V - P_{s,a}^c V\|_\infty > 3|\calS| \varepsilon \right) = \mathbb{E}[ \mathbb{P} \left( \|\hat{\rho}^t_{s,a} V - P_{s,a}^c V\|_\infty > 3|\calS| \varepsilon \Big|N_{s,a}^t\right)] $$
Let $p_1,\ldots,p_n$ be the i.i.d. samples from posterior $\psi_{s,a}^t$ conditioned on $\psi_{s,a}^t$. Let $\Bar{\psi}^t_{s,a} = \psi_{s,a}^t / \sum_{s'\in\calS} \psi_{s,a}^t(s')$ We have 
\begin{align*}
        &\mathbb{P} \left( \|\hat{\rho}^t_{s,a} V - P_{s,a}^c V\|_\infty > 3|\calS| \varepsilon \Big|N_{s,a}^t\right) \\
         = & \mathbb{P} \left(\bigcup_{i\in[n]} \|p_i- P_{s,a}^c \|_\infty >  3|\calS| \varepsilon \Big| N_{s,a}^t\right)\\
        \le & \mathbb{P} \left( \bigcup_{i\in[n]} \|p_i- P_{s,a}^c \|_\infty >  3|\calS| \varepsilon \Big|N_{s,a}^t \right) \\
         \le & \mathbb{P} \left( \bigcup_{i\in[n]} \|p_i- \Bar{\psi}_{s,a}^t \|_\infty > 3|\calS| \varepsilon - \| \Bar{\psi}_{s,a}^t - P_{s,a}^c \|_\infty \Big|N_{s,a}^t \right) \\
          \le  & \mathbb{P} \left( \bigcup_{i\in[n]} \|p_i- \Bar{\psi}_{s,a}^t \|_\infty > |\calS|\varepsilon \Big|N_{s,a}^t \right) + \mathbb{P} \left( \| \Bar{\psi}_{s,a}^t - P_{s,a}^c \|_\infty > 2 |\calS| \varepsilon \Big|N_{s,a}^t\right)
          \\
          \le  & \sum_{i\in[n]} \mathbb{P} \left(  \|p_i- \Bar{\psi}_{s,a}^t \|_\infty > |\calS|\varepsilon \Big|N_{s,a}^t \right) + \mathbb{P} \left( \| \Bar{\psi}_{s,a}^t - P_{s,a}^c \|_\infty > 2 |\calS| \varepsilon \Big|N_{s,a}^t\right)
        %   \\
        % \ge & \mathbb{P} \left( \cap_{i\in[n]} \cap_{s'\in\calS} |p_i(s') - P_{s,a}^c(s')| \le \varepsilon |\Big| \| \Bar{\psi}_{s,a}^t - P_{s,a}^c \|_\infty \le 2 |\calS| \varepsilon, \right) \mathbb{P} \left( \| \Bar{\psi}_{s,a}^t - P_{s,a}^c \|_\infty \le 2 |\calS| \varepsilon \Big|N_{s,a}^t\right) \\
        % \ge& 1 - \sum_{i\in[n],s\in\calS} \mathbb{P} \left( |p_i(s')- P_{s,a}^c(s')| > \varepsilon |\psi_{s,a}^t \right) \\
        % \ge &1 - 2n|\calS| \mathrm{e}^{-2(N_{s,a}^t+1)\varepsilon^2}
\end{align*}
Since the distribution of $p_i$ only depends on $\psi_{s,a}^t$, hence  we can further write 
\begin{align*}
    & \mathbb{P} \left(  \|p_i- \Bar{\psi}_{s,a}^t \|_\infty > |\calS|\varepsilon \Big|N_{s,a}^t \right) \\
    = &\mathbb{E} \left[ \mathbb{P} \left( \|p_i- \Bar{\psi}_{s,a}^t \|_\infty >|\calS|\varepsilon \Big|\psi_{s,a}^t\right)\Big|N_{s,a}^t \right]
\end{align*}
Then,
\begin{align*}
    &\mathbb{P} \left( \|p_i- \Bar{\psi}_{s,a}^t \|_\infty > |\calS|\varepsilon \Big|\psi_{s,a}^t\right)\\
    \ge& \sum_{s\in\calS} \mathbb{P} \left( |p_i(s')- \Bar{\psi}_{s,a}^t(s')| > \varepsilon \Big|\psi_{s,a}^t \right) \\
    \le & 2|\calS| \mathrm{e}^{-2(N_{s,a}^t+1)\varepsilon^2}.
\end{align*}
 The second inequality holds since for a Dirichlet random vector $p$ with parameter $\psi \in \mathbb{R}^{|\calS|}$, we have $x^\top p$ is $\frac{1}{4\sum_{s'\in\calS}\psi(s')}$-sub-Gaussian as long as $x$ belongs to the $|\calS|-1$ dimension simplex \cite{marchal2017sub}. Then, take $x = e_i$, which is the vector with $i$th element to be 1 and others to be 0. We have $p(s')$ is $\frac{1}{4\sum_{s'\in\calS}\psi(s')}$-sub-Gaussian, which leads to $  \mathbb{P} \left( |p_i(s') - \Bar{\psi}_{s,a}^t(s') | \le \varepsilon | \psi_{s,a}^t\right) \le 2\mathrm{e}^{-2(N_{s,a}^t+|\calS|) \varepsilon^2} \le2\mathrm{e}^{-2(N_{s,a}^t+1) \varepsilon^2} $

Next, by Lemma \ref{lem: concentration expon}, since $|\calS|\varepsilon > \frac{|\calS|}{N_{s,a}^t}$, we have 
\begin{align*}
    &\mathbb{P} \left( \| \Bar{\psi}_{s,a}^t - P_{s,a}^c \|_\infty > 2 |\calS| \varepsilon \Big|N_{s,a}^t\right) \\
    \le & 2 |\calS| \mathrm{e}^{-2  \left(2|\calS| \varepsilon - \frac{|\calS|}{N_{s,a}^t}\right)^2 N_{s,a}^t} \\
    \le & 2|\calS| \mathrm{e}^{-2  |\calS|^2 \varepsilon^2 N_{s,a}^t}\\
    \le & 2|\calS| \mathrm{e}^{-2 (N_{s,a}^t +1) \varepsilon^2 }.
\end{align*}
The last inequality holds as long as $N^t_{s,a} \ge 1$ and $|\calS| \ge 2 $.
Then we obtain 
$$\mathbb{P} \left( \|\hat{\rho}^t_{s,a} V - P_{s,a}^c V\|_\infty \ge 3|\calS| \varepsilon \Big|N_{s,a}^t\right) \le 2(n+1)|\calS| \mathrm{e}^{-2 (N_{s,a}^t +1) \varepsilon^2 }$$
Set $ 2(n+1)|\calS| \mathrm{e}^{-2(N_{s,a}^t+1)\varepsilon^2} = \frac{\delta}{2|\calS||\calA|T}$, we obtain $\varepsilon = \sqrt{\frac{1}{2(N_{s,a}^t+1)}\ln{\frac{4(n+1)|\calS|^2|\calA|T}{\delta}}}$. Then we obtain with probability $1-\frac{\delta}{2|\calS||\calA|T}$, $$ \|\hat{\rho}^t_{s,a} V - P_{s,a}^c V\|_\infty \le 3\sqrt{\frac{|\calS|^2}{2(N_{s,a}^t + 1)}\log{\frac{4(n+1)|\calS|^2|\calA|T}{\delta}}}, \quad \forall \|V\|_\infty \le 1. $$
Further with a union bound, we obtain the desired result.  
\hfill $\blacksquare$
\begin{lem} \label{lem: upper bound}
    Let $Q_t$ denote the Q function given by the algorithm. Then we have with probability $1-\delta/2$, $Q_t(s,a) \ge Q^*(s,a)$ holds for all $s\in\calS,a\in\calA,t\in[T]$.
\end{lem}
\proof{}
Let $C^p$ be the concentration event in Lemma \ref{lem: hat rho concentration}. We know with probability $1-\frac{\delta}{2}$, $C^p$ holds. We prove $Q_t(s,a) \ge Q^*(s,a)$ by induction. When $t=0$, then $Q_0(s,a) = \frac{1}{1-\gamma} \ge Q^*(s,a)$. Assume this holds for $Q_t$. Then for $t+1$, if $Q_{t+1} = Q_t$, then $Q_{t+1} \ge Q_t$ by induction. Otherwise, 
$$ Q_{t+1}(s,a) = r(s,a) + \gamma \hat{\rho}_{s,a}^t V_t + \text{UCB}_t(s,a).$$
Then we have 
\begin{align*}
    Q_{t+1}(s,a) - Q^*(s,a) = & \gamma \rho_{s,a}^tV_t - \gamma P_{s,a}^cV^* + \text{UCB}_t(s,a) \\
    =& \gamma (\hat{\rho}_{s,a}^t V_t - \hat{\rho}_{s,a}^t V^*) + \gamma(\hat{\rho}V^* - P_{s,a}^vV^*) + \text{UCB}_t(s,a)
\end{align*}
Since $V_t \ge V^*$ by induction and that $\rho$ preserves the order, we know $\hat{\rho}_{s,a}^t V_t - \hat{\rho}_{s,a}^t V^* \ge 0 $. Furthermore, when $C^p$ holds, by Lemma \ref{lem: hat rho concentration},   we know   
$$\gamma(\hat{\rho}V^* - P_{s,a}^cV^*) \ge - \frac{3\gamma}{1-\gamma} \sqrt{\frac{|\calS|^2}{2(N_{s,a}^t + 1)}\log{\frac{4(n+1)|\calS|^2|\calA|T}{\delta}}} = -\text{UCB}_t(s,a)$$
Hence, we have $Q_{t+1}(s,a)-Q^*(s,a) \ge 0$.
\hfill $\blacksquare$

\begin{lem} \label{lem: sum N^t}
    $$ \sum_{t=0}^T \frac{1}{\sqrt{N_{s_t,a_t}^t+1}} \le \min\left\{
        |\calS|\sqrt{T+1},\sqrt{|\calS|(T+1)\ln{(T+1)}}  \right\} $$
\end{lem}
We can now derive the overall regret. 

\proof{}
    Assume the concentration event in both Lemma \ref{lem: upper bound} holds true. By Lemma \ref{lem: upper bound}, we know with probability $1-\frac{\delta}{2}$, $V^* \le V_t$. 
    Then 
    \begin{align*}
        \text{Regret}(T) \le &\sum_{t=1}^T V_t(s_t) - V_t^\pi(s_t) \\
        \le & \sum_{t=1}^T \left[r(s_t,a_t) + \gamma \hat{\rho}^t_{s_t,a_t} V_{t-1} + \text{UCB}_{t-1}(s_t,a_t) - r(s_t,a_t) - \gamma P_{s_t,a_t}^c V^\pi_{t+1}\right]\\
        = & \sum_{t=1}^T\gamma (\hat{\rho}_{s_t,a_t}^t V_{t-1} - P_{s_t,a_t}^cV_{t-1}) + \sum_{t=1}^T\gamma (P_{s_t,a_t}^c (V_{t-1}-V_{t+1}^\pi)) + \sum_{t=1}^T\text{UCB}_{t-1}(s_t,a_t) \\
        =& \gamma \underbrace{\sum_{t=1}^T(\hat{\rho}_{s_t,a_t}^t V_{t-1} - P_{s_t,a_t}^cV_{t-1})}_{(I)} + 
        \gamma\underbrace{\sum_{t=1}^T\left[ P_{s_t,a_t}^c (V_{t-1}-V_{t+1}^\pi) - (V_{t-1}(s_{t+1}) - V_{t+1}^\pi(s_{t+1}) \right]}_{(II)}\\
        & +\gamma \underbrace{\sum_{t=1}^T[V_{t-1}(s_{t+1}) - V_{t+1}^\pi(s_{t+1})]}_{(III)} + \sum_{t=1}^T\text{UCB}_{t-1}(s_t,a_t) &
        \hfill \blacksquare
    \end{align*}

To bound (I), by lemma \ref{lem: hat rho concentration}, we know $\hat{\rho}_{s_t,a_t}^t V_{t-1} - P_{s_t,a_t}^cV_{t-1} \le \frac{3}{1-\gamma}\sqrt{\frac{|\calS|^2}{2(N_{s_t,a_t}^t + 1)}\log{\frac{4(n+1)|\calS|^2|\calA|T}{\delta}}} $. Hence, 
\begin{align*}
    (I) \le & \sum_{t=1}^T \frac{3}{1-\gamma}\sqrt{\frac{|\calS|^2}{2(N_{s_t,a_t}^t + 1)}\log{\frac{4(n+1)|\calS|^2|\calA|T}{\delta}}} \\
    \le & \min\{ |\calS|\sqrt{T+1},\sqrt{|\calS|(T+1)\ln{(T+1)}}\}\frac{3}{1-\gamma}\sqrt{\frac{|\calS|^2}{2}\log{\frac{4(n+1)|\calS|^2|\calA|T}{\delta}}}.
\end{align*}
To bound (II), notice (II) is the summation of a martingale difference sequence given filtration $\mathcal{H}_t = \{s_\tau,a_\tau, \tau\in[t] \}$.
Furthermore, each term in (II) can be bounded by $\frac{2}{1-\gamma}$. By Azuma-Hoeffeding inequality, we obtain
$$ \mathbb{P}\left( \sum_{t=1}^T\left[ P_{s_t,a_t}^c (V_{t-1}-V_{t+1}^\pi) - (V_{t-1}(s_{t+1}) - V_{t+1}^\pi(s_{t+1}) \right] \ge \varepsilon \right) \le \mathrm{e}^{-(\frac{2\varepsilon}{1-\gamma})^2 T}$$
Hence, with probability $1-\frac{\delta}{2}$,
$$(II) \le \frac{2}{1-\gamma}\sqrt{2T\ln{\frac{2}{\delta}}}.$$
For (III), by Lemma 5.2 in \cite{he2021nearly}, we have 
$$(III) \le \sum_{t=1}^T V_t(s_t) -V_t^\pi(s_t) + \frac{2|\calS|+2}{1-\gamma}.$$
We can further bound 
\begin{align*}
    \sum_{t=1}^{T} \text{UCB}_{t-1}(s_t,a_t)  \le \frac{3\gamma}{1-\gamma}\sqrt{\frac{|\calS|^2}{2}\log{\frac{4(n+1)|\calS|^2|\calA|T}{\delta}}} \min\{ |\calS|\sqrt{T+1},\sqrt{|\calS|(T+1)\ln{(T+1)}}\}.
\end{align*}
Putting things together we have 

\begin{align*}
     &\sum_{t=1}^T V_t(s_t) -V_t^\pi(s_t) \\
     \le&\gamma \left(  \sum_{t=1}^T V_t(s_t) -V_t^\pi(s_t)\right) \\
     &+\frac{6\gamma}{1-\gamma}\sqrt{\frac{|\calS|^2}{2}\log{\frac{4(n+1)|\calS|^2|\calA|T}{\delta}}} \min\{ |\calS|\sqrt{T+1},\sqrt{|\calS|(T+1)\ln{(T+1)}}\}\\
     & + \gamma\frac{2|\calS|+2}{1-\gamma} + \frac{2\gamma}{1-\gamma}\sqrt{2T\ln{\frac{2}{\delta}}}
\end{align*}
Hence,
\begin{align*}
     &\text{Regret(T)} \\
     \le &\sum_{t=1}^T V_t(s_t) -V_t^\pi(s_t) \\
     \le&\frac{6\gamma}{(1-\gamma)^2}\sqrt{\frac{|\calS|^2}{2}\log{\frac{4(n+1)|\calS|^2|\calA|T}{\delta}}} \min\{ |\calS|\sqrt{T+1},\sqrt{|\calS|(T+1)\ln{(T+1)}}\}\\
     & + \gamma\frac{2|\calS|+2}{(1-\gamma)^2} + \frac{2\gamma}{(1-\gamma)^2}\sqrt{2T\ln{\frac{2}{\delta}}} \\
      \le&\frac{6\gamma}{(1-\gamma)^2}\sqrt{\frac{|\calS|^2}{2}\log{\frac{4(3-2\alpha)|\calS|^2|\calA|T}{(1-\alpha)\delta}}} \min\{ |\calS|\sqrt{T+1},\sqrt{|\calS|(T+1)\ln{(T+1)}}\}\\
     & + \gamma\frac{2|\calS|+2}{(1-\gamma)^2} + \frac{2\gamma}{(1-\gamma)^2}\sqrt{2T\ln{\frac{2}{\delta}}} 
\end{align*}

For the expected regret, setting $\delta = \frac{1}{T}$ and notice $\text{Regret}(T) \le \frac{T}{1-\gamma}$, we obtain the desired result.
\hfill $\blacksquare$
% \fi

%% file: main.bbl
\begin{thebibliography}{37}
\providecommand{\natexlab}[1]{#1}
\providecommand{\url}[1]{\texttt{#1}}
\expandafter\ifx\csname urlstyle\endcsname\relax
  \providecommand{\doi}[1]{doi: #1}\else
  \providecommand{\doi}{doi: \begingroup \urlstyle{rm}\Url}\fi

\bibitem[Agrawal and Goyal(2014)]{agrawal2014thompson}
Shipra Agrawal and Navin Goyal.
\newblock Thompson sampling for contextual bandits with linear payoffs.
\newblock \emph{arXiv preprint arXiv:1209.3352}, 2014.

\bibitem[Badrinath and Kalathil(2021)]{badrinath2021robust}
Kishan~Panaganti Badrinath and Dileep Kalathil.
\newblock Robust reinforcement learning using least squares policy iteration with provable performance guarantees.
\newblock In \emph{International Conference on Machine Learning}, pages 511--520. PMLR, 2021.

\bibitem[Blanchet et~al.(2024)Blanchet, Lu, Zhang, and Zhong]{blanchet2024double}
Jose Blanchet, Miao Lu, Tong Zhang, and Han Zhong.
\newblock Double pessimism is provably efficient for distributionally robust offline reinforcement learning: Generic algorithm and robust partial coverage.
\newblock \emph{Advances in Neural Information Processing Systems}, 36, 2024.

\bibitem[Dong et~al.(2022)Dong, Li, Wang, and Zhang]{dong2022online}
Jing Dong, Jingwei Li, Baoxiang Wang, and Jingzhao Zhang.
\newblock Online policy optimization for robust {{MDP}}.
\newblock \emph{arXiv preprint arXiv:2209.13841}, 2022.

\bibitem[Dong et~al.(2019)Dong, Wang, Chen, and Wang]{dong2019q}
Kefan Dong, Yuanhao Wang, Xiaoyu Chen, and Liwei Wang.
\newblock Q-learning with {UCB} exploration is sample efficient for infinite-horizon {MDP}.
\newblock \emph{arXiv preprint arXiv:1901.09311}, 2019.

\bibitem[El~Ghaoui and Nilim(2005)]{el2005robust}
Laurent El~Ghaoui and Arnab Nilim.
\newblock Robust solutions to markov decision problems with uncertain transition matrices.
\newblock \emph{Operations Research}, 53\penalty0 (5):\penalty0 780--798, 2005.

\bibitem[Gonz{\'a}lez-Trejo et~al.(2002)Gonz{\'a}lez-Trejo, Hern{\'a}ndez-Lerma, and Hoyos-Reyes]{gonzalez2002minimax}
JI~Gonz{\'a}lez-Trejo, On{\'e}simo Hern{\'a}ndez-Lerma, and Luis~F Hoyos-Reyes.
\newblock Minimax control of discrete-time stochastic systems.
\newblock \emph{SIAM Journal on Control and Optimization}, 41\penalty0 (5):\penalty0 1626--1659, 2002.

\bibitem[He et~al.(2021)He, Zhou, and Gu]{he2021nearly}
Jiafan He, Dongruo Zhou, and Quanquan Gu.
\newblock Nearly minimax optimal reinforcement learning for discounted {MDP}s.
\newblock \emph{Advances in Neural Information Processing Systems}, 34:\penalty0 22288--22300, 2021.

\bibitem[Hong and Liu(2011)]{hong2011monte}
L~Jeff Hong and Guangwu Liu.
\newblock Monte carlo estimation of value-at-risk, conditional value-at-risk and their sensitivities.
\newblock In \emph{Proceedings of the 2011 Winter Simulation Conference (WSC)}, pages 95--107. IEEE, 2011.

\bibitem[Iyengar(2005)]{iyengar2005robust}
Garud~N Iyengar.
\newblock Robust dynamic programming.
\newblock \emph{Mathematics of Operations Research}, 30\penalty0 (2):\penalty0 257--280, 2005.

\bibitem[Kakade(2003)]{kakade2003sample}
Sham~Machandranath Kakade.
\newblock \emph{On the sample complexity of reinforcement learning}.
\newblock University of London, University College London (United Kingdom), 2003.

\bibitem[Kallus et~al.(2022)Kallus, Mao, Wang, and Zhou]{kallus2022doubly}
Nathan Kallus, Xiaojie Mao, Kaiwen Wang, and Zhengyuan Zhou.
\newblock Doubly robust distributionally robust off-policy evaluation and learning.
\newblock In \emph{International Conference on Machine Learning}, pages 10598--10632. PMLR, 2022.

\bibitem[Lattimore and Hutter(2012)]{lattimore2012pac}
Tor Lattimore and Marcus Hutter.
\newblock Pac bounds for discounted {MDP}s.
\newblock In \emph{Algorithmic Learning Theory: 23rd International Conference, ALT 2012, Lyon, France, October 29-31, 2012. Proceedings 23}, pages 320--334. Springer, 2012.

\bibitem[Levine et~al.(2020)Levine, Kumar, Tucker, and Fu]{levine2020offline}
Sergey Levine, Aviral Kumar, George Tucker, and Justin Fu.
\newblock Offline reinforcement learning: Tutorial, review, and perspectives on open problems.
\newblock \emph{arXiv preprint arXiv:2005.01643}, 2020.

\bibitem[Lin and Zhou(2023)]{lin2023approximate}
Yifan Lin and Enlu Zhou.
\newblock Approximate bilevel difference convex programming for bayesian risk markov decision processes.
\newblock \emph{arXiv preprint arXiv:2301.11415}, 2023.

\bibitem[Lin et~al.(2022{\natexlab{a}})Lin, Ren, and Zhou]{lin2022bayesian}
Yifan Lin, Yuxuan Ren, and Enlu Zhou.
\newblock Bayesian risk markov decision processes.
\newblock \emph{Advances in Neural Information Processing Systems}, 35:\penalty0 17430--17442, 2022{\natexlab{a}}.

\bibitem[Lin et~al.(2022{\natexlab{b}})Lin, Wang, and Zhou]{lin2022risk}
Yifan Lin, Yuhao Wang, and Enlu Zhou.
\newblock Risk-averse contextual multi-armed bandit problem with linear payoffs.
\newblock \emph{Journal of Systems Science and Systems Engineering}, pages 1--22, 2022{\natexlab{b}}.

\bibitem[Liu et~al.(2022)Liu, Bai, Blanchet, Dong, Xu, Zhou, and Zhou]{liu2022distributionally}
Zijian Liu, Qinxun Bai, Jose Blanchet, Perry Dong, Wei Xu, Zhengqing Zhou, and Zhengyuan Zhou.
\newblock Distributionally robust $ q $-learning.
\newblock In \emph{International Conference on Machine Learning}, pages 13623--13643. PMLR, 2022.

\bibitem[Mannor et~al.(2016)Mannor, Mebel, and Xu]{mannor2016robust}
Shie Mannor, Ofir Mebel, and Huan Xu.
\newblock Robust {MDP}s with k-rectangular uncertainty.
\newblock \emph{Mathematics of Operations Research}, 41\penalty0 (4):\penalty0 1484--1509, 2016.

\bibitem[Marchal and Arbel(2017)]{marchal2017sub}
Olivier Marchal and Julyan Arbel.
\newblock On the sub-gaussianity of the beta and dirichlet distributions.
\newblock 2017.

\bibitem[Mo et~al.(2021)Mo, Qi, and Liu]{mo2021learning}
Weibin Mo, Zhengling Qi, and Yufeng Liu.
\newblock Learning optimal distributionally robust individualized treatment rules.
\newblock \emph{Journal of the American Statistical Association}, 116\penalty0 (534):\penalty0 659--674, 2021.

\bibitem[Osband et~al.(2013)Osband, Russo, and Van~Roy]{osband2013more}
Ian Osband, Daniel Russo, and Benjamin Van~Roy.
\newblock (more) efficient reinforcement learning via posterior sampling.
\newblock \emph{Advances in Neural Information Processing Systems}, 26, 2013.

\bibitem[Panaganti and Kalathil(2022)]{panaganti2022sample}
Kishan Panaganti and Dileep Kalathil.
\newblock Sample complexity of robust reinforcement learning with a generative model.
\newblock In \emph{International Conference on Artificial Intelligence and Statistics}, pages 9582--9602. PMLR, 2022.

\bibitem[Panaganti et~al.(2022)Panaganti, Xu, Kalathil, and Ghavamzadeh]{panaganti2022robust}
Kishan Panaganti, Zaiyan Xu, Dileep Kalathil, and Mohammad Ghavamzadeh.
\newblock Robust reinforcement learning using offline data.
\newblock \emph{Advances in neural information processing systems}, 35:\penalty0 32211--32224, 2022.

\bibitem[Russo et~al.(2018)Russo, Van~Roy, Kazerouni, Osband, Wen, et~al.]{russo2018tutorial}
Daniel~J Russo, Benjamin Van~Roy, Abbas Kazerouni, Ian Osband, Zheng Wen, et~al.
\newblock A tutorial on thompson sampling.
\newblock \emph{Foundations and Trends{\textregistered} in Machine Learning}, 11\penalty0 (1):\penalty0 1--96, 2018.

\bibitem[Shen et~al.(2023)Shen, Xu, and Zavlanos]{shen2023wasserstein}
Yi~Shen, Pan Xu, and Michael Zavlanos.
\newblock Wasserstein distributionally robust policy evaluation and learning for contextual bandits.
\newblock \emph{Transactions on Machine Learning Research}, 2023.

\bibitem[Si et~al.(2023)Si, Zhang, Zhou, and Blanchet]{si2023distributionally}
Nian Si, Fan Zhang, Zhengyuan Zhou, and Jose Blanchet.
\newblock Distributionally robust batch contextual bandits.
\newblock \emph{Management Science}, 2023.

\bibitem[Wang et~al.(2023)Wang, Si, Blanchet, and Zhou]{wang2023finite}
Shengbo Wang, Nian Si, Jose Blanchet, and Zhengyuan Zhou.
\newblock A finite sample complexity bound for distributionally robust q-learning.
\newblock In \emph{International Conference on Artificial Intelligence and Statistics}, pages 3370--3398. PMLR, 2023.

\bibitem[Wang and Zou(2021)]{wang2021online}
Yue Wang and Shaofeng Zou.
\newblock Online robust reinforcement learning with model uncertainty.
\newblock \emph{Advances in Neural Information Processing Systems}, 34:\penalty0 7193--7206, 2021.

\bibitem[Wang and Zou(2022)]{wang2022policy}
Yue Wang and Shaofeng Zou.
\newblock Policy gradient method for robust reinforcement learning.
\newblock In \emph{International Conference on Machine Learning}, pages 23484--23526. PMLR, 2022.

\bibitem[Wang and Zhou(2023)]{wang2023bayesian}
Yuhao Wang and Enlu Zhou.
\newblock Bayesian risk-averse q-learning with streaming observations.
\newblock \emph{Advances in Neural Information Processing Systems}, 36, 2023.

\bibitem[Wiesemann et~al.(2013)Wiesemann, Kuhn, and Rustem]{wiesemann2013robust}
Wolfram Wiesemann, Daniel Kuhn, and Ber{\c{c}} Rustem.
\newblock Robust markov decision processes.
\newblock \emph{Mathematics of Operations Research}, 38\penalty0 (1):\penalty0 153--183, 2013.

\bibitem[Wu et~al.(2018)Wu, Zhu, and Zhou]{wu2018bayesian}
Di~Wu, Helin Zhu, and Enlu Zhou.
\newblock A bayesian risk approach to data-driven stochastic optimization: Formulations and asymptotics.
\newblock \emph{SIAM Journal on Optimization}, 28\penalty0 (2):\penalty0 1588--1612, 2018.

\bibitem[Xu and Mannor(2010)]{xu2010distributionally}
Huan Xu and Shie Mannor.
\newblock Distributionally robust markov decision processes.
\newblock \emph{Advances in Neural Information Processing Systems}, 23, 2010.

\bibitem[Zhou and Xie(2015)]{zhou2015simulation}
Enlu Zhou and Wei Xie.
\newblock Simulation optimization when facing input uncertainty.
\newblock In \emph{2015 Winter Simulation Conference (WSC)}, pages 3714--3724. IEEE, 2015.

\bibitem[Zhou et~al.(2024)Zhou, Liu, Cheng, Kalathil, Kumar, and Tian]{zhou2024natural}
Ruida Zhou, Tao Liu, Min Cheng, Dileep Kalathil, PR~Kumar, and Chao Tian.
\newblock Natural actor-critic for robust reinforcement learning with function approximation.
\newblock \emph{Advances in neural information processing systems}, 36, 2024.

\bibitem[Zhou et~al.(2021)Zhou, Zhou, Bai, Qiu, Blanchet, and Glynn]{zhou2021finite}
Zhengqing Zhou, Zhengyuan Zhou, Qinxun Bai, Linhai Qiu, Jose Blanchet, and Peter Glynn.
\newblock Finite-sample regret bound for distributionally robust offline tabular reinforcement learning.
\newblock In \emph{International Conference on Artificial Intelligence and Statistics}, pages 3331--3339. PMLR, 2021.

\end{thebibliography}
